\pdfoutput=1

\documentclass[11pt,dvipsnames]{article}

\usepackage{acl}

\usepackage{times}
\usepackage{latexsym}
\usepackage{graphicx}
\usepackage[noend]{algpseudocode}
\usepackage{algorithm}
\usepackage{xcolor}
\usepackage{multirow}
\usepackage{amsmath,amssymb}
\usepackage{caption}
\usepackage{subcaption}

\usepackage[T1]{fontenc}

\usepackage[utf8]{inputenc}

\usepackage{microtype}

\usepackage{inconsolata}

\newcommand{\btheta}{\boldsymbol{\theta}}
\newcommand{\bv}{\boldsymbol{v}}
\newcommand{\bg}{\boldsymbol{g}}
\newcommand{\bs}{\boldsymbol{s}}
\newcommand{\br}{\boldsymbol{r}}

\title{Should I try multiple optimizers when fine-tuning pre-trained Transformers for NLP tasks? Should I tune their hyperparameters?}

\author{
Nefeli Gkouti\textsuperscript{$1$, $2$, $3$}\space\space\space\space
  {\bf Prodromos Malakasiotis}\textsuperscript{$1, 4$}\\
  {\bf Stavros Toumpis}\textsuperscript{$1$}\space\space\space\space\space\space\space
  {\bf Ion Androutsopoulos}\textsuperscript{$1$, $3$}\\\\
  \textsuperscript{$1$} Department of Informatics, Athens University of Economics and Business, Greece \\
  \textsuperscript{$2$} Department of Informatics and Telecommunications,\\
  National and Kapodistrian University of Athens, Greece\\
  \textsuperscript{$3$} Archimedes/Athena RC, Greece \\
  \textsuperscript{$4$} Workable \\
  nefeli.gkouti@athenarc.gr,\space\space\{rulller, toumpis, ion\}@aueb.gr 
}

\begin{document}
\maketitle
\begin{abstract}  
NLP research has explored different neural model architectures and sizes, datasets, training objectives, and transfer learning techniques. However, the choice of optimizer during training has not been explored as extensively. Typically, some variant of Stochastic Gradient Descent (SGD) is employed, selected among numerous variants, using unclear criteria, often with minimal or no tuning of the optimizer's hyperparameters. Experimenting with five GLUE datasets, two models (DistilBERT and DistilRoBERTa), and seven popular optimizers (SGD, SGD with Momentum, Adam, AdaMax, Nadam, AdamW, and AdaBound), we find that when the hyperparameters of the optimizers are tuned, there is no substantial difference in test performance across the five more elaborate (adaptive) optimizers, despite differences in training loss. Furthermore, tuning just the learning rate is in most cases as good as tuning all the hyperparameters. Hence, we recommend picking any of the best-behaved adaptive optimizers (e.g., Adam) and tuning only its learning rate. When no hyperparameter can be tuned, SGD with Momentum is the best choice.
\end{abstract}

\section{Introduction}

NLP research has investigated how different neural model architectures (e.g., RNNs, CNNs, Transformers), model sizes, datasets, training (or pre-training) objectives, and transfer-learning techniques (e.g., pre-training and fine-tuning) affect performance and efficiency. However, the effects of using different optimizers to minimize the training loss have not been explored as extensively. Adam \cite{adam} is a popular choice, but there are dozens of alternatives, mostly variants of Stochastic Gradient Descent (SGD) \cite{sgd1,sgd2}.\footnote{\citet{valley} list more than a hundred  optimizers that have been used in deep learning.} Optimizer selection seems to be based on unclear criteria and anecdotal information. Furthermore, most optimizers have several hyperparameters, often minimally tuned (e.g., only the learning rate is tuned) or left to their default values. Hence, when models need to be trained (e.g., pre-trained or fine-tuned), it is unclear if the available computing resources should be used to try multiple optimizers, tune their hyperparameters, both, or none. 

Our work is inspired by \citet{valley}, who experimented with 15 optimizers and 8 tasks from DeepOBS \cite{DeepOBS}. Their most striking finding was that trying several optimizers with default hyperparameters was almost as beneficial as (and cheaper than) picking any single (competent) optimizer and tuning its hyperparameters.  Hence, practitioners would be advised to try multiple optimizers with defaults, rather than selecting a single optimizer (e.g., based on anecdotal evidence) and tuning its hyperparameters, when they cannot tune both the choice of optimizer and hyperparameters (which is expensive). In fact, tuning the hyperparameters of a single optimizer was only slightly better than using its defaults, which also advocates against hyperparameter tuning. However, \citet{valley} considered only one NLP task (character-level language modeling) with an RNN, acknowledging that their findings may not hold with more complicated models, such as Transformers. They also found indications that the best optimizer may depend on the model and task.

We complement the work of \citet{valley} from an NLP perspective by investigating empirically if it is worth (a) trying multiple optimizers and/or (b) tuning their hyperparameters (and which ones), when fine-tuning pre-trained Transformer encoders. We experiment with five tasks from GLUE \cite{glue}, using seven popular optimizers, namely SGD, SGD with Momentum (SGDM) \cite{mom}, Adam and AdaMax \cite{adam}, Nadam \cite{nadam}, AdamW \cite{adamw}, and AdaBound \cite{adabound}. For each task and optimizer, we fine-tune DistilBERT \cite{distil} and DistilRoBERTa,\footnote{\url{https://huggingface.co/distilroberta-base}} two efficient pre-trained Transformers that allow us to complete the many experiments of this work with our limited budget. We consider three cases: using the default hyperparameters of the optimizers, tuning all their hyperparameters, or tuning only their learning rates. 

With the exception of the two non-adaptive optimizers considered, i.e., plain SGD and SGDM, which are largely unaffected by hyperparameter tuning, the test performance of the other five (adaptive) optimizers improves substantially when their hyperparameters are tuned, unlike smaller overall improvements reported by \citet{valley}. Interestingly, in most cases tuning only the learning rate is as good as (and much cheaper than) tuning all the hyperparameters. Furthermore, when hyperparameters (or just the learning rate) are tuned, the adaptive optimizers have very similar test scores in most cases, unlike plain SGD and SGDM, which are clearly the worst and second worst, respectively. This parity of test performance of the adaptive optimizers is obtained despite occasional differences in the training loss they reach. When no hyperparameter can be tuned (e.g., due to limited budget), SGDM is the best choice and AdaBound occasionally works relatively well, but the other optimizers considered are much worse. Trying multiple optimizers with defaults \cite{valley} is reasonably good too, because of the good untuned performance of SGDM and AdaBound. However, our experiments suggest that picking just one of the best-behaved adaptive optimizers i.e., with consistent top-performance across datasets, e.g., Adam, and tuning only its learning rate is the best strategy.

\section{Optimizers Considered}\label{sec:optimizers}
\label{overview}

All the optimizers we consider aim at tuning the weights vector $\btheta$ of a model, so that a loss function $f(\btheta)= \frac{1}{K}\sum_{k=1}^K f_k(\btheta)$ is minimized, where $f_k(\btheta)$ is the loss of the $k$-th training example, and $K$ the size of the training set. The gradient of $f(\btheta)$,
\begin{equation}
\bg(\btheta) \triangleq \nabla f(\btheta) = \frac{1}{K}\sum_{k=1}^K \nabla f_k(\btheta),
\label{eq:exact_gradient}
\end{equation}
is of special interest as it points to the direction along which $f(\btheta)$ increases the fastest. The optimizers we consider are iterative, i.e., they create a sequence of points $\{\btheta_t,~t=0,1,\dots\}$, such that each $\btheta_{t+1}$ is selected by taking a step away from the previous point $\btheta_t$ in an attempt to decrease $f(\btheta)$. The step could be towards the opposite direction of $\bg(\btheta)$. However, exactly computing $\bg(\btheta)$ at each step is too costly for large training set sizes $K$.

\subsection{Non-adaptive Optimizers}

The simplest optimizer is pure \textbf{Gradient Descent (GD)} (Algorithm~\ref{pgd}). At each iteration the gradient $\bg(\btheta)$ is computed exactly (line 4, as in Eq.~\ref{eq:exact_gradient}), then the next point $\btheta_{t+1}$ is selected to be towards the opposite direction. The \textbf{learning rate} $\epsilon>0$ is a hyperparameter that affects the size of the steps. Computing the exact gradient at each step, however, is too costly for large training sets (i.e., large $K$).

\begin{algorithm}[t]
{\small
\begin{algorithmic}[1]
\State \textbf{Input:} 
\begin{itemize}
\item initial time step $t\leftarrow 0$; initial weight vector $\btheta_0$ 
\item learning rate $\epsilon>0$
\item \colorbox{Aquamarine}{momentum parameter $\alpha \in [0,1)$}
\item \colorbox{Aquamarine}{initial velocity $\bv_0\leftarrow 0$}
\end{itemize}
\While{stopping criterion not met}
    \State 
    \colorbox{Apricot}{select all examples ($m \leftarrow K$)}
    
    \hskip-0.5em\colorbox{GreenYellow}{sample mini-batch of $m \ll K$ examples}
    
    \hskip-0.5em\colorbox{Aquamarine}{sample mini-batch of $m \ll K$ examples}
 
    \State $\bg_t\leftarrow\frac{1}{m}\sum_{k=1}^m \nabla f_k(\btheta_t)$

    \State  $\btheta_{t+1}\leftarrow\btheta_t-\epsilon \bg_t$ \colorbox{Aquamarine}{$ + \alpha \bv_t$}
 
    \State \colorbox{Aquamarine}{$\bv_{t+1} \leftarrow \alpha \bv_t - \epsilon \bg_t$};    
    $\;t \leftarrow t+1$
\EndWhile
\end{algorithmic}
} 
\caption{\small\colorbox{Apricot}{Gradient Descent (GD)} \\\colorbox{GreenYellow}{Stochastic GD (SGD)} \colorbox{Aquamarine}{SGD with Momentum (SGDM)}}
\label{pgd}
\end{algorithm}

\vspace{2mm}\noindent\textbf{Stochastic Gradient Descent (SGD)} (Algorithm~\ref{pgd})  \emph{estimates} the gradient at each step, by using a sample (mini-batch) of $m \ll K$ examples (line 4); the next point is selected as in GD (line 5).

\vspace{2mm}\noindent\textbf{Stochastic Gradient Descent with Momentum (SGDM)} \cite{polyak1964} (Algorithm~\ref{pgd}) aims to accelerate learning by suppressing oscillations in the created sequence of points. It maintains an exponentially weighted moving average of past gradients, termed the \textbf{velocity}, $\bv_t$ (line 6). The direction and size of the next step (line 5) are now determined by a linear combination of the latest gradient estimate $\bg_t$ and the velocity ($\bv_t$). Intuitively, the sequence of points $\{\btheta_t,~t=0,1,\dots\}$ resembles the movement of a ball traveling (in the space of weight vectors) towards points of lower altitude (loss), but also subject to its own momentum.

\vspace{2mm}\noindent The optimizers above are called non-adaptive, because their learning rate $\epsilon$ is fixed. The optimizers we discuss next modify $\epsilon$ while creating the sequence of points and are, hence, called adaptive.

\subsection{Adaptive Optimizers} 

The loss function $f(\btheta)$ of large neural models corresponds to a complicated hyper-surface. In some directions, the loss may change rapidly, in other directions slowly. This makes choosing the learning rate $\epsilon$ difficult: if $\epsilon$ is too large, the minimum along the chosen direction of a step may be overshot; if, however, $\epsilon$ is too small, progress towards smaller values of the loss function will be slow. This difficulty of choosing a good, uniform, $\epsilon$ a priori is one of the reasons adaptive optimizers are so useful.

\vspace{2mm}\noindent\textbf{Adaptive Moments (Adam)} \cite{adam} follows Algorithm~\ref{adam}. Like SGDM, it uses the concept of momentum by maintaining a velocity-type vector $\bs_t$ (line 6), termed the \textbf{first moment}. It also adapts the learning rate by scaling it (line 10) roughly inversely proportionally to the square root of an exponentially weighed average $\br_t$ (line 7) of the component-wise squared gradient estimate, $\bg^2_t$, termed the \textbf{second moment}. Lines 8 and 9 normalize the two moments to take into account biases due to their initial values $\mathbf{0}$ \cite{adam}.

\begin{algorithm}[t]
{\small 
\begin{algorithmic}[1]
\State \textbf{Input:}
\begin{itemize}
\item initial time step $t \leftarrow 0$; initial weight vector $\btheta_0$ 
\item learning  rate $\epsilon>0$; decay rates 
  $\rho_1,~\rho_2 \in [0, 1)$
\item small constant $\delta>0$
\item first moment $\bs_0 \leftarrow \mathbf{0}$; second moment $\br_0 \leftarrow \mathbf{0}$
\item \colorbox{Aquamarine}{regularization factor $\lambda \in (0,1)$}
\end{itemize}
\While{stopping criterion not met}
    \State $t \leftarrow t+1$
    \State sample mini-batch of $m$ examples
    \State $\bg_t \leftarrow \frac{1}{m}\sum_{i=1}^{m}\nabla f_{i}(\boldsymbol{\btheta}_{t-1}) $
    \State $\bs_t \leftarrow \rho_{1}\bs_{t-1} + (1-\rho_1)\bg_t$
    \State $\br_t \leftarrow \rho_{2}\br_{t-1} + (1-\rho_2)\bg^2_t $
    \State 
    $\hat{\bs}_t \leftarrow$ \colorbox{Apricot}{$\frac{\bs_t}{1-{{\rho_1}^t}}$} \colorbox{Aquamarine}{$\frac{\bs_t}{1-{{\rho_1}^t}}$} \colorbox{GreenYellow}{$\frac{\rho_1 \bs_t}{1-{{\rho_1}^{t+1}}}+\frac{(1-\rho_1) \bg_t}{1-{{\rho_1}^t}}$}

    \State $\hat{\br}_t \leftarrow$  \colorbox{GreenYellow}{$\rho_2 \cdot$}  $\frac{\br_t}{1-{\rho_2}^t}$ 

    \State $\boldsymbol{\btheta}_{t} \leftarrow \boldsymbol{\btheta}_{t-1} -\epsilon\frac{\hat{\bs}_t}{\delta+\sqrt{\hat{\br}_t}}$ \colorbox{Aquamarine}{$-\lambda \btheta_{t-1}$} 
\EndWhile
\end{algorithmic}
} 
\caption{\small\colorbox{Apricot}{Adaptive Moment Optimization (Adam)}\\\colorbox{GreenYellow}{Nesterov-Accelerated Adam (Nadam)}\\\colorbox{Aquamarine}{Adam with decoupled weight decay (AdamW)}\\ (all vector operations are elementwise)}\label{adam}
\end{algorithm}

\begin{algorithm}[t]
{\small
\begin{algorithmic}[1]
\State \textbf{Input:}
\begin{itemize}
\item initial time step $t \leftarrow0$; initial weight vector $\btheta_0$ 
\item 
learning  rate $\epsilon>0$; decay rates 
  $\rho_1,~\rho_2 \in [0, 1)$
\item small constant $\delta>0$
\item first moment $\bs_0 \leftarrow \mathbf{0}$; second moment $\br_0 \leftarrow \mathbf{0}$
\item \colorbox{GreenYellow}{lower bound function $\eta^l_t$}
\item \colorbox{GreenYellow}{upper bound function $\eta^u_t$}
\end{itemize}
\While{stopping criterion not met}
    \State $t \leftarrow t + 1$
    \State sample mini-batch of $m$ examples
    \State $\bg_t \leftarrow\frac{1}{m}\sum_{i=1}^{m}\nabla f_{i}(\boldsymbol{\btheta}_{t-1}) $
    \State $\bs_t \leftarrow \rho_{1}\bs_{t-1} + (1-\rho_1)\bg_t$
    
    \State     $\br_t \leftarrow$ \colorbox{Apricot}{$\max\{\rho_2 \br_{t-1}, |\bg_t|\}$} \colorbox{GreenYellow}{$\rho_{2}\br_{t-1} + (1-\rho_2)\bg^2_t$}
    
    \State \colorbox{GreenYellow}{$\eta_t \leftarrow \left[ \frac{\epsilon}{\sqrt{\br}}\right]^{\eta^u_t}_{\eta^l_t}$}\State
$\boldsymbol{\btheta}_t \leftarrow$ \colorbox{Apricot} {$\boldsymbol{\btheta}_{t-1} -\frac{\epsilon}{1-\rho_1^t} \cdot \frac{\bs_t}{\br_t}$}  \colorbox{GreenYellow}{$\boldsymbol{\btheta}_{t-1} -\eta_t \cdot \bs_t$} 
    
\EndWhile
\end{algorithmic}
} 
\caption{\small\colorbox{Apricot}{AdaMax} \hspace*{2mm} \colorbox{GreenYellow}{AdaBound} \\(all vector operations are elementwise)} \label{adamax_adabound}
\end{algorithm}

\vspace{2mm}\noindent\textbf{Nadam} \cite{nadam} is identical to Adam, except that it uses Nesterov momentum~\cite{nag}, which has been shown to be somewhat superior to plain momentum~\cite{nesvsmom}. The only difference is in lines 8, 9 in Algorithm~\ref{adam}, which ensure that the momentum step incorporates the estimated gradient at the location where it is used.

\vspace{2mm}\noindent\textbf{AdamW} \cite{adamw} (Algorithm~\ref{adam}) is based on the empirical observation that smaller weights tend to overfit less. Hence, it adds (in line 10) a term $-\lambda \btheta_{t-1}$, where $\lambda \in (0,1)$, to decay the weight vector towards the origin. 

\vspace{2mm}\noindent\textbf{AdaMax} \cite{adam}  is identical to Adam, except that the two moments $\bs_t$ and $\br_t$ are not normalized, and the second moment ($\br_t$) is computed using line 7 in  Algorithm~\ref{adamax_adabound}, thus it is no longer an exponentially weighed average. This variation of Adam was proposed in the same paper that introduced Adam, as a more stable variant.

\vspace{2mm}\noindent\textbf{AdaBound} \cite{adabound}  (Algorithm~\ref{adamax_adabound}) ensures that extreme values for the learning rate are avoided, by bounding it by both a dynamic upper bound $\eta^u_t$ and a dynamic lower bound $\eta^l_t$  that start from infinity and zero, respectively, and then converge to a finite common value $\epsilon$ as the time step $t$ increases. The operation $[x]_l^u$ in line 8 clips the vector $x$ elementwise so that the output lies in the interval $[l,u]$.
AdaBound initially behaves like Adam, and gradually transforms to SGD.

\vspace{2mm}\noindent The optimizers we consider include simple non-adaptive baselines (SGD, SGDM), the most commonly used adaptive optimizer (Adam), its sibling AdaMax, as well as adaptive optimizers that incorporate influential ideas, in particular Nesterov momentum (Nadam), weight decay (AdamW), and dynamic bounds (AdaBound). AdamW and AdaBound are also two of the most recent optimizers.

\section{Experiments}

\subsection{Datasets and Evaluation Measures}\label{subsec:data}

We experiment with five GLUE tasks \cite{glue}.\footnote{To speed up our experiments, we do not use the other four GLUE tasks (QQP, QNLI, RTE, WNLI), which are all textual inference/paraphrasing tasks, like MRPC and MNLI.} The datasets of all tasks are in English. Each experiment is repeated with five random training/development/test splits, and we report average scores and standard deviations over the repetitions.\vspace{1mm}

\noindent\textbf{SST-2} \cite{sst2} is a binary sentiment classification dataset with 68.8k sentences from movie reviews (one label per sentence). To speed up our experiments, we sample 18k (from the 68.8k) sentences anew in each of the five repetitions and split them into training (15k), development (1.5k) and test (1.5k) subsets. The class distribution of the 68.8k sentences (55\% positive sentiment) is preserved in all subsets of every split.\vspace{2mm}

\noindent\textbf{MRPC} \cite{mrpc} contains 5.8k sentence pairs from online news. Each pair is classified as containing paraphrases (sentences with the same meaning) or not. We use 80\% of the 5.8k pairs for training, 10\% for development, 10\% for testing, preserving the class distribution (67\% paraphrases).\vspace{2mm}

\noindent\textbf{CoLA} \cite{cola} contains 9.6k word sequences labeled to indicate if each sequence is a grammatically correct sentence or not. We use 80\% of the sequences for training, 10\% for development and 10\% for testing, preserving the class distribution (70\% acceptable).\vspace{2mm}

\noindent\textbf{STS-B} \cite{stsb} contains 7.2k sentence pairs from news headlines, video/image captions, and natural language inference data. Each pair is annotated with a similarity score from 1 to 5. In each repetition, we sample 80\% of the 7.2k pairs for training, 10\% for development, 10\% for testing.\vspace{2mm}

\noindent\textbf{MNLI} \cite{mnli} contains 393k premise-hypothesis pairs for training, and 19.6k pairs for development. The task is to predict if the premise entails, contradicts, or is neutral to the hypothesis. To speed up the experiments, in each repetition we sample (anew) 50k from the 393k pairs for training, 9.8k from the 19.6k for development, and the remaining 9.8k for testing, preserving the original class distribution (balanced).\footnote{We also ensure that development and test sets contain 50\% `in-domain' (matched) pairs and 50\% `out-of-domain'.}
\vspace{2mm}

\noindent\textbf{Evaluation measures:} We use the measures adopted by GLUE \cite{glue}, i.e., Accuracy for SST-2 and MNLI, Macro-F1 for MRPC, Matthews correlation \cite{matthews1975comparison} for CoLA, and Pearson correlation \cite{pearson} for STS-B.

\subsection{Experimental Setup} \label{subsec:setup}

\noindent\textbf{Transformer models:} Given the volume of the experiments and our limited resources, we fine-tune: (i) DistilBERT \cite{distil}, a distilled BERT-base \cite{bert} with 40\% fewer parameters that runs 60\% faster, but retains 95\% of BERT-base’s performance on GLUE, according to its creators; and (ii) DistilRoBERTa, a similarly distilled version of RoBERTa-base \cite{roberta}.\vspace{1mm}

\noindent\textbf{Hyperparameter tuning:}
For each optimizer, model, task, and data split (Section~\ref{subsec:data}), we try 30 different combinations (trials) of hyperparameter values, as selected by Optuna \cite{optuna}, seeking to maximize the evaluation measure of the task on the development subset.\footnote{In Optuna, we use Tree-Structured Parzen Estimation \cite{bergstra2011tpe,bergstra2013tpe} and median-based pruning.} In each trial, we retain the weights from the epoch with the best development score. The hyperparameter search space of each optimizer includes the default values proposed by its creators, with the exception of the learning rate $\epsilon$ of adaptive optimizers, since it is standard practice when fine-tuning Transformers with adaptive optimizers to use much smaller $\epsilon$.\footnote{More details on the hyperparameter search space and the selected values are provided in Appendix~\ref{sec:hyperparameter}.} We repeat the experiments, tuning only $\epsilon$. We also report results with default hyperparameter values.\vspace{2mm}

\noindent\textbf{Loss functions:} We minimize cross-entropy for the classification tasks (SST-2, MRPC, CoLA, and MNLI), and mean squared error for STS-B.

\begin{figure*}
\centering

\includegraphics[width=0.7\textwidth]{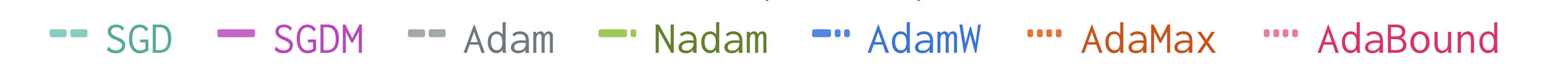}

    \begin{subfigure}[b]{0.49\textwidth}
        \centering
        \includegraphics[width=\textwidth]{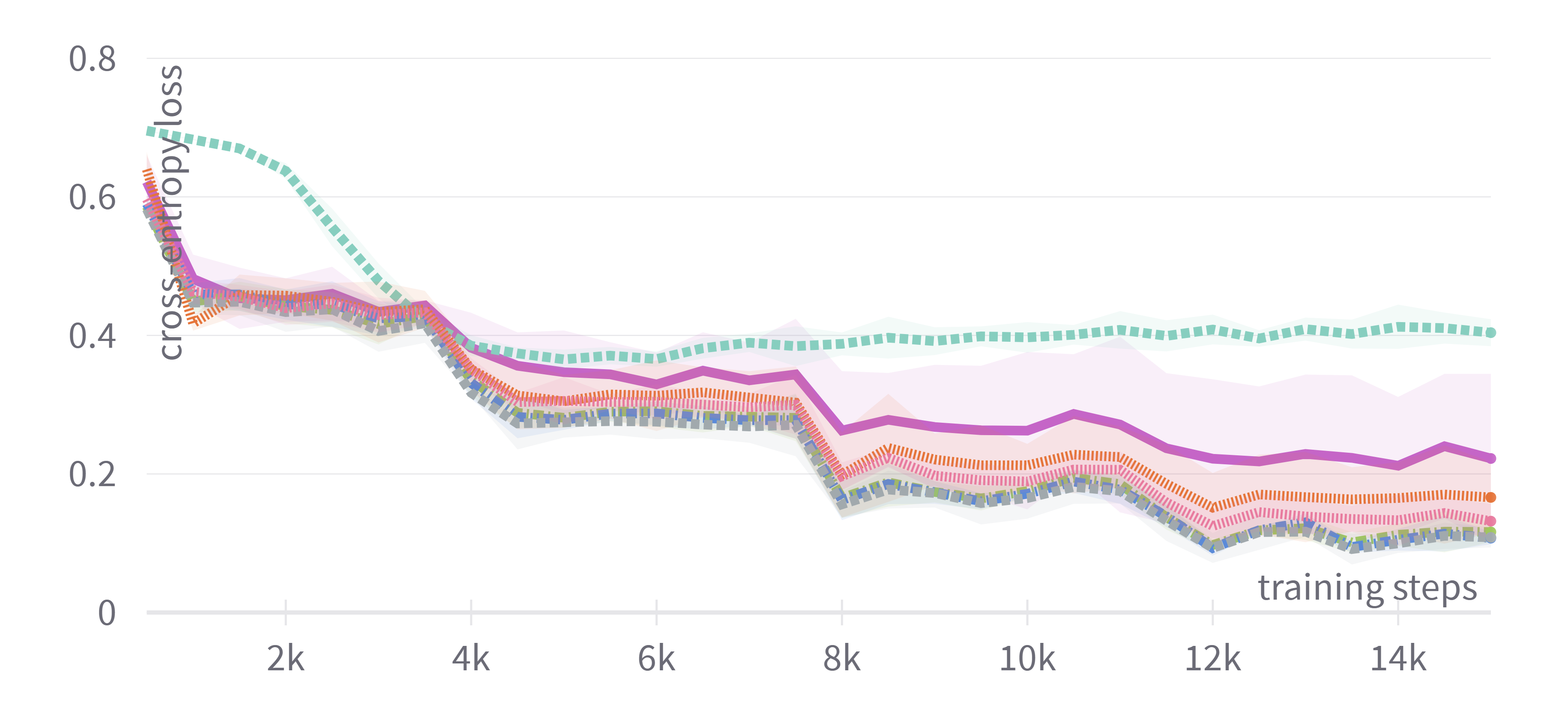}
        \caption{SST-2}
        \label{subfig:sst2losstuned}
        \vspace{0.5mm}
    \end{subfigure}
    \hfill\addtocounter{subfigure}{-1}
    \begin{subfigure}[b]{0.49\textwidth}
        \centering
        \includegraphics[width=\textwidth]{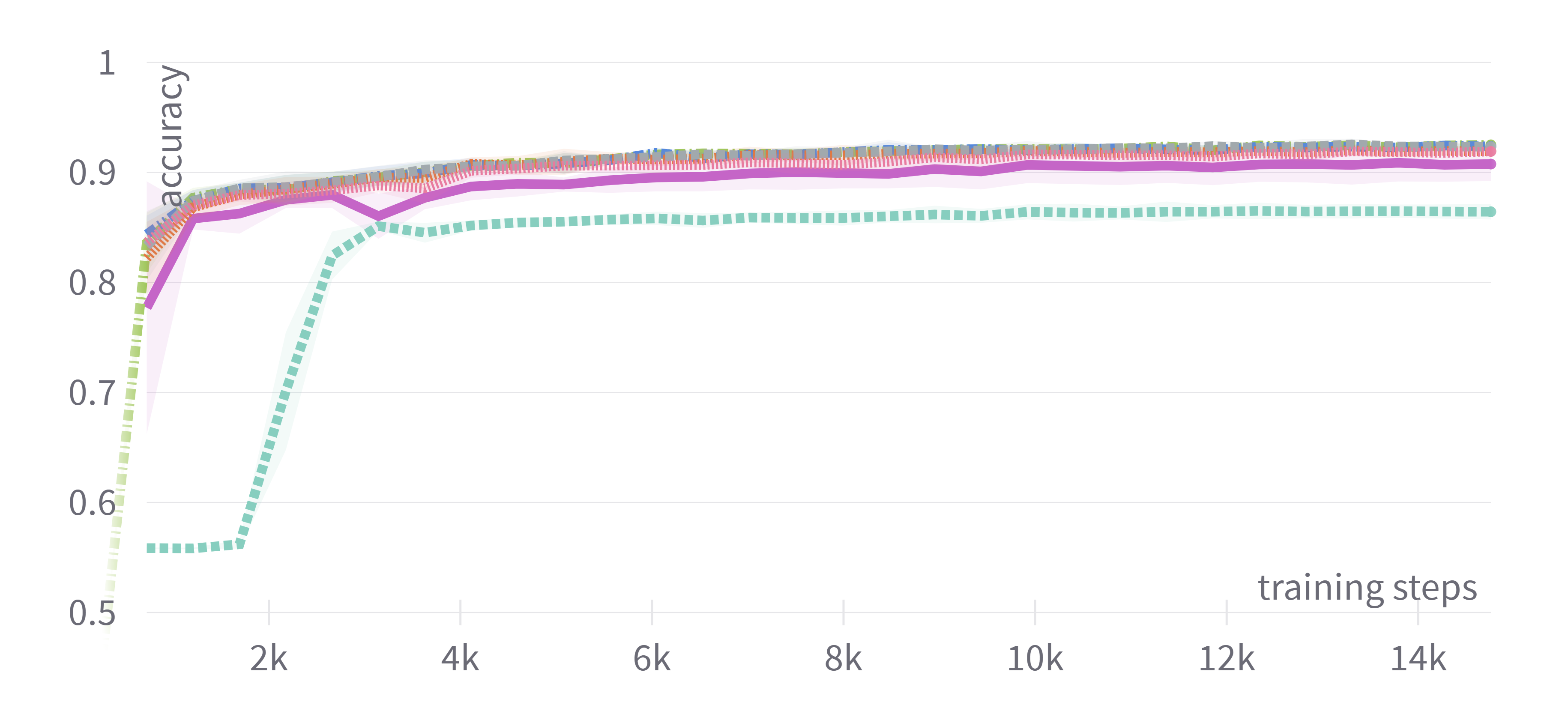}
        \caption{SST-2}
        \label{subfig:sst2acctuned}
        \vspace{0.5mm}
    \end{subfigure}
    \hfill
    \begin{subfigure}[b]{0.49\textwidth}
        \centering
        \includegraphics[width=\textwidth]{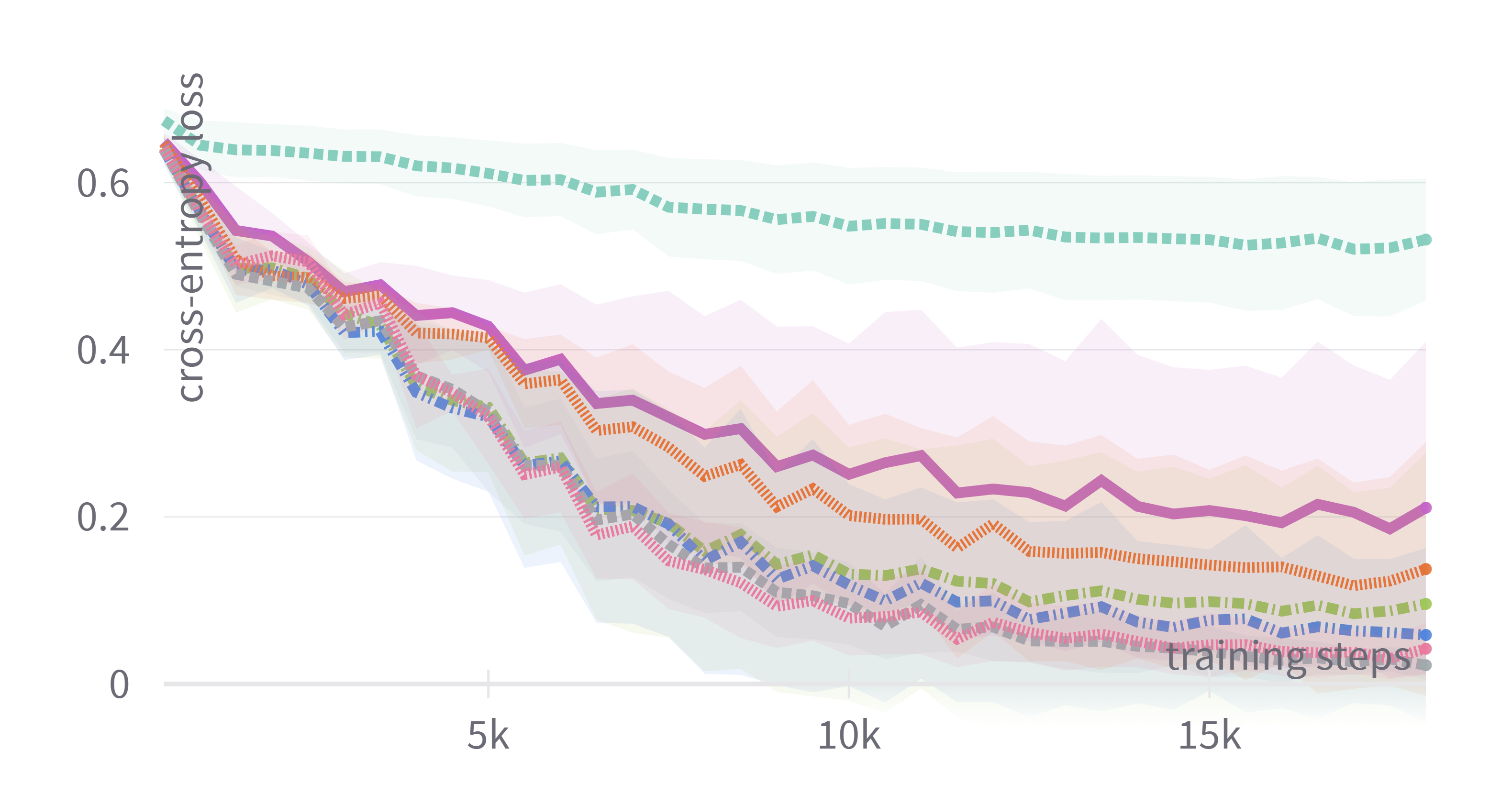}
        \caption{MRPC}
        \label{subfig:mrpclosstuned}
        \vspace{0.5mm}
    \end{subfigure}
    \hfill\addtocounter{subfigure}{-1}
    \begin{subfigure}[b]{0.49\textwidth}
        \centering
        \includegraphics[width=\textwidth]{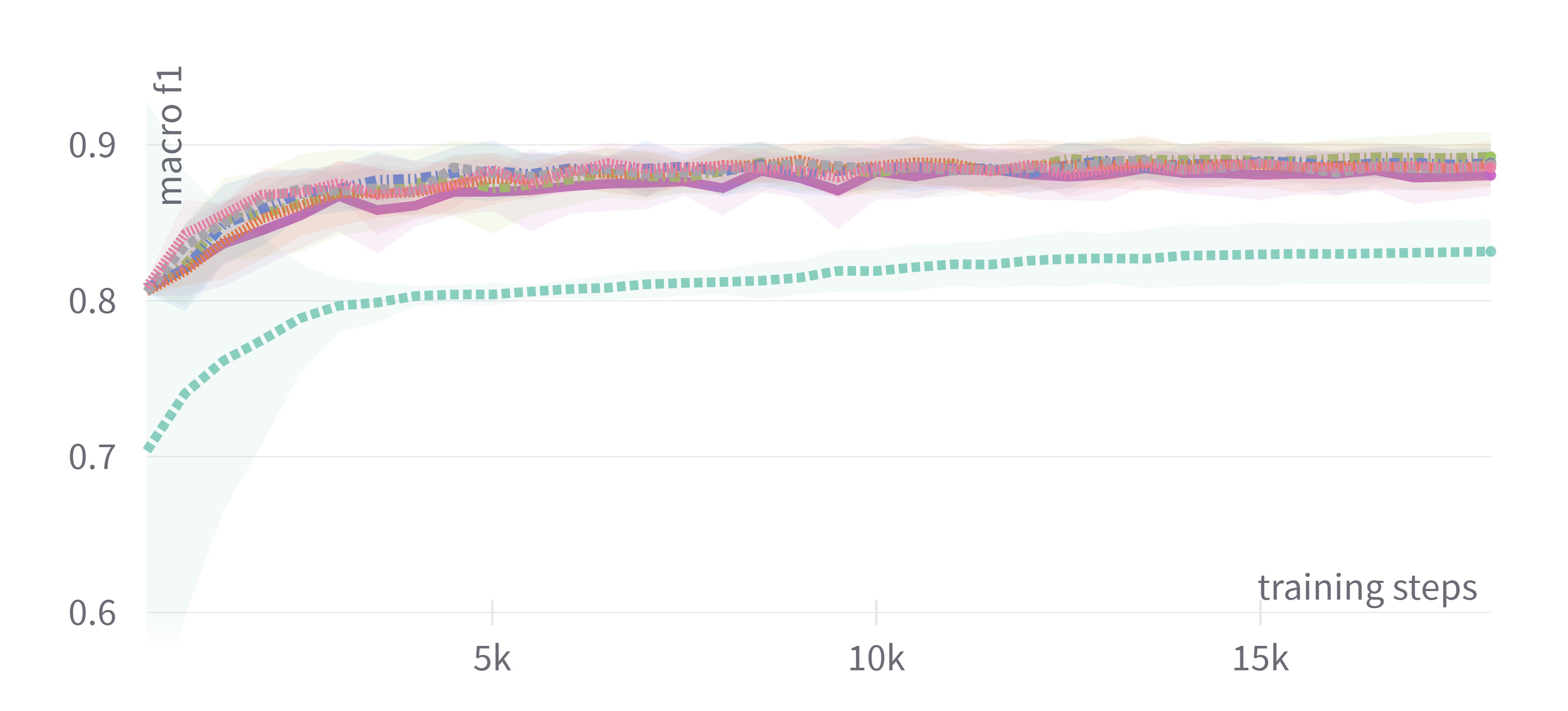}
        \caption{MRPC}
        \label{subfig:mrpcf1tuned}
        \vspace{0.5mm}
    \end{subfigure}
    \hfill
    \begin{subfigure}[b]{0.49\textwidth}
        \centering
        \includegraphics[width=\textwidth]{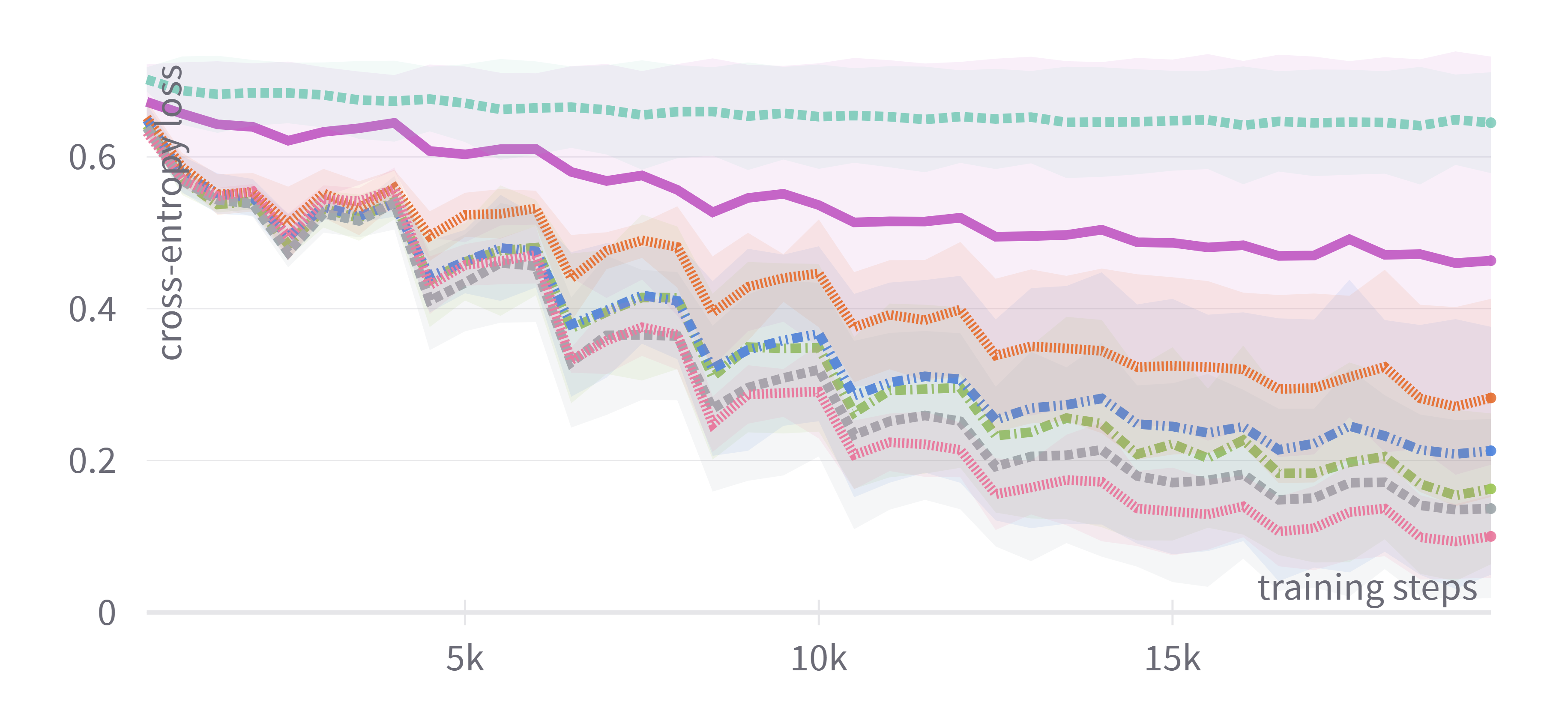}
        \caption{CoLA}
        \label{subfig:colalosstuned}
        \vspace{0.5mm}
    \end{subfigure}
    \hfill\addtocounter{subfigure}{-1}
    \begin{subfigure}[b]{0.49\textwidth}
        \centering
        \includegraphics[width=\textwidth]{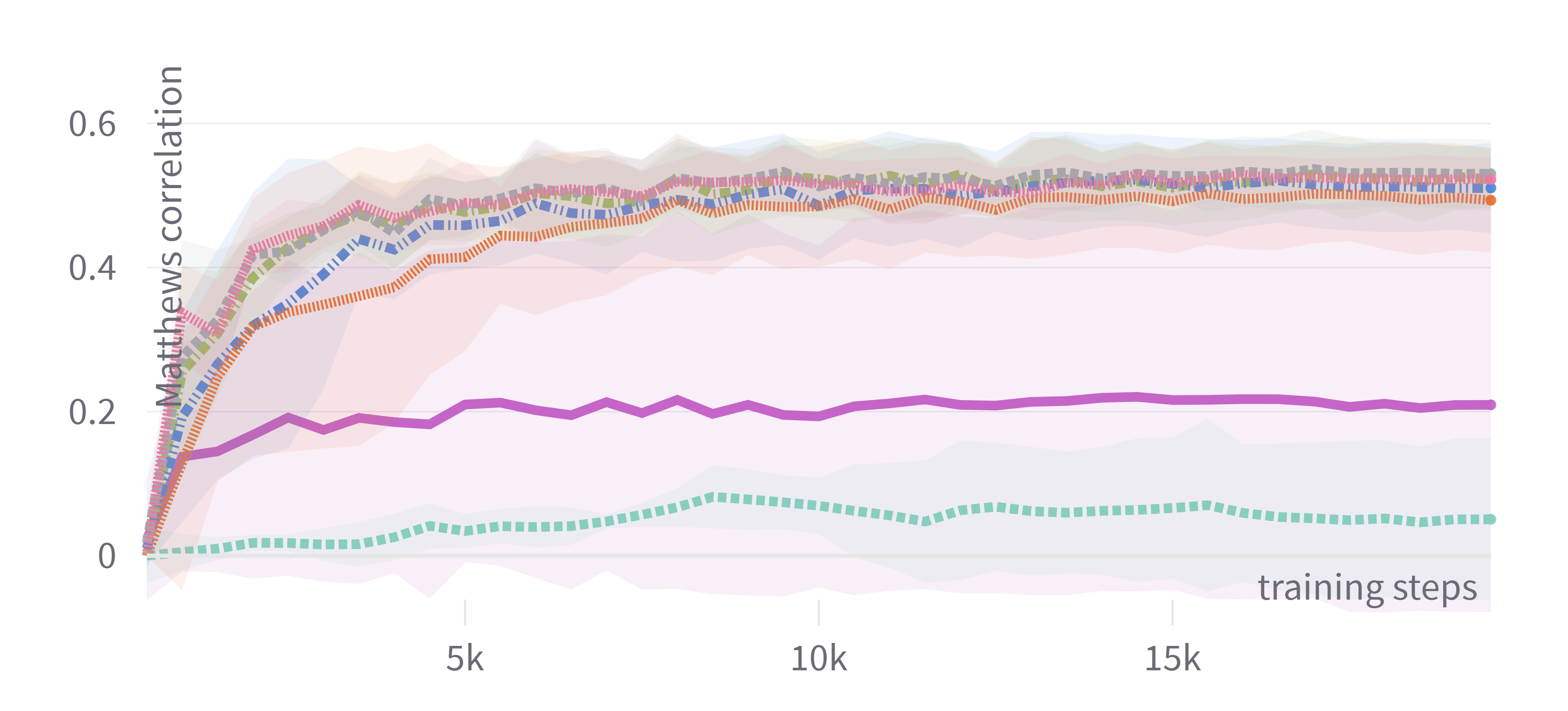}
        \caption{CoLA}
        \label{subfig:colamathtuned}
        \vspace{0.5mm}
    \end{subfigure}
    \hfill
    \begin{subfigure}[b]{0.49\textwidth}
        \centering
        \includegraphics[width=\textwidth]{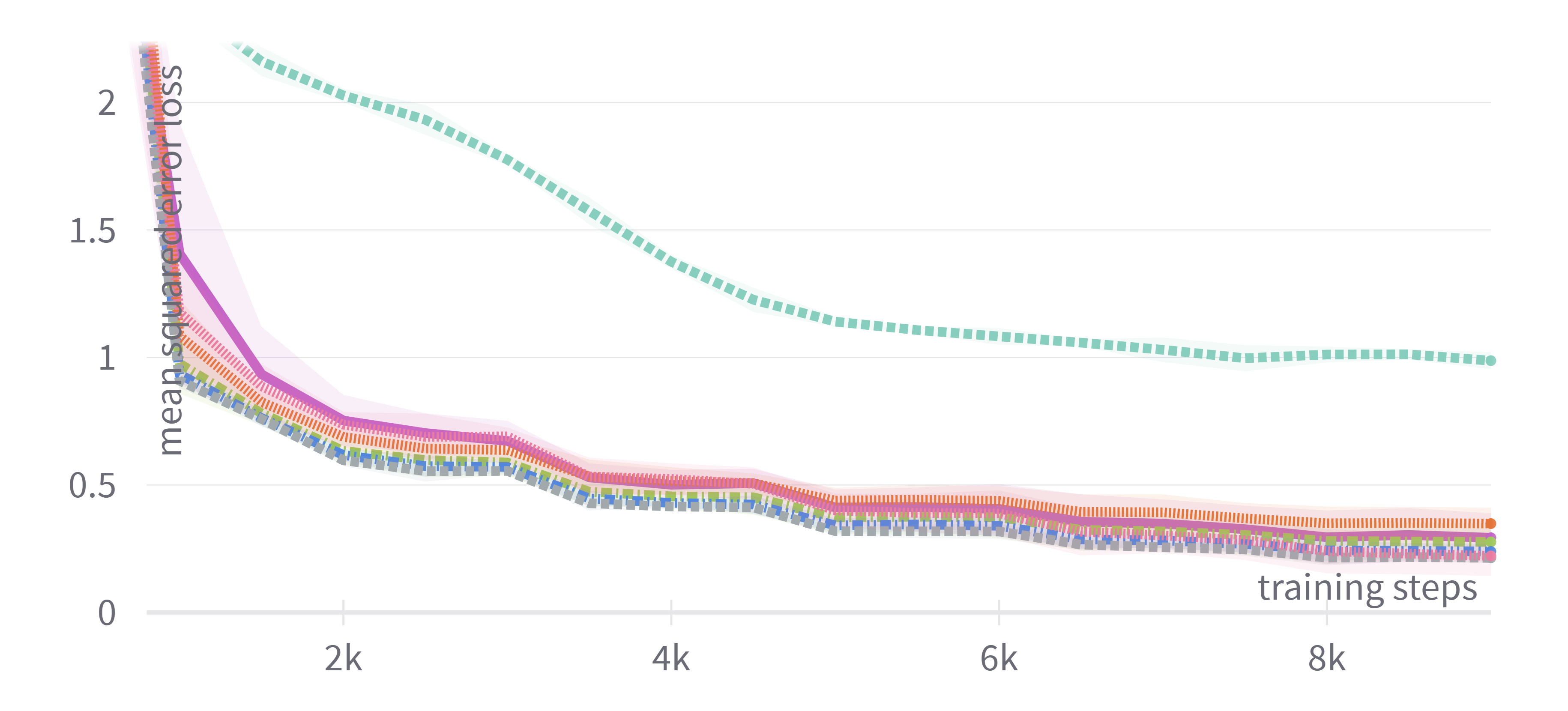}
        \caption{STS-B}
        \label{subfig:stsblosstuned}
        \vspace{0.5mm}
    \end{subfigure}
    \hfill\addtocounter{subfigure}{-1}
    \begin{subfigure}[b]{0.49\textwidth}
        \centering
        \includegraphics[width=\textwidth]{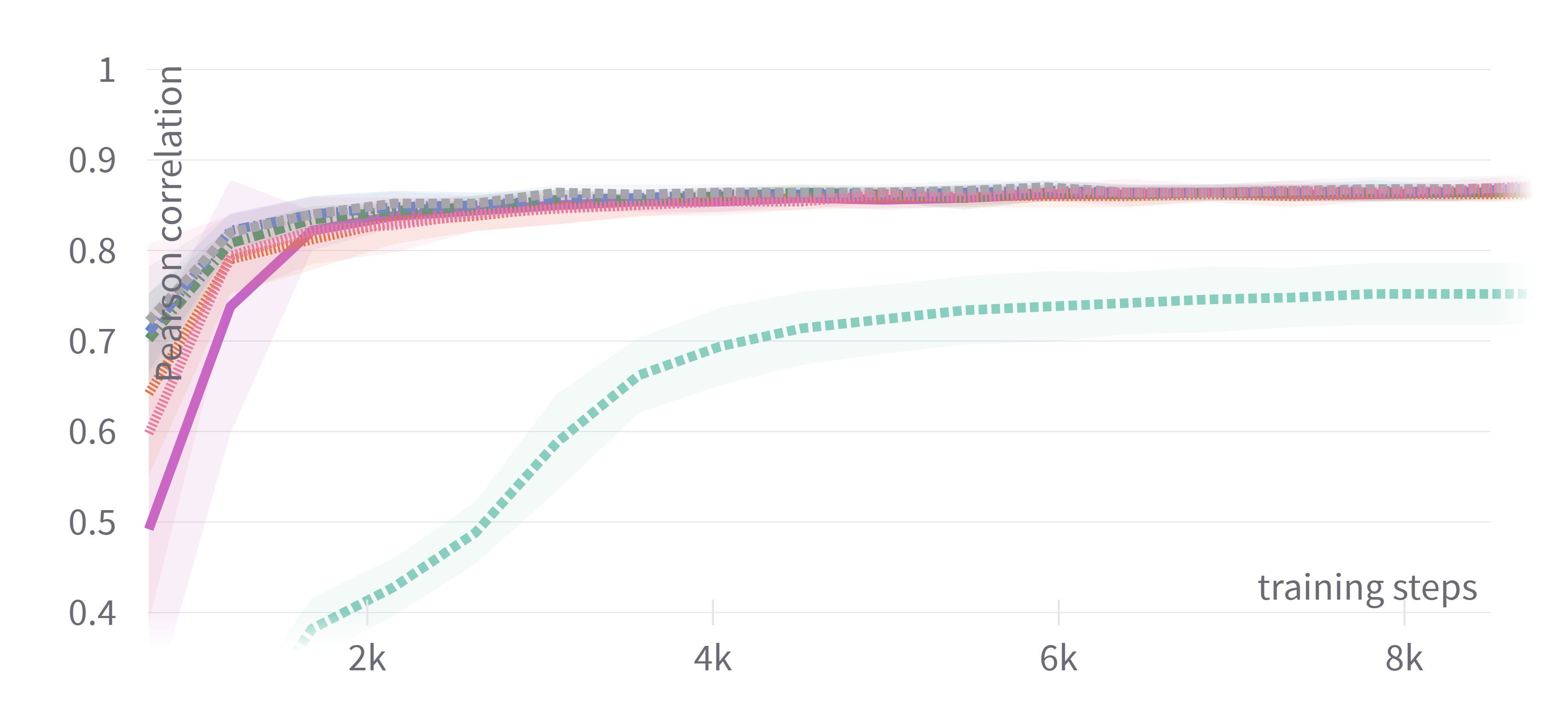}
        \caption{STS-B}
        \label{subfig:stsbpeartuned}
        \vspace{0.5mm}
    \end{subfigure}
    \hfill
    \begin{subfigure}[b]{0.49\textwidth}
        \centering
        \includegraphics[width=\textwidth]{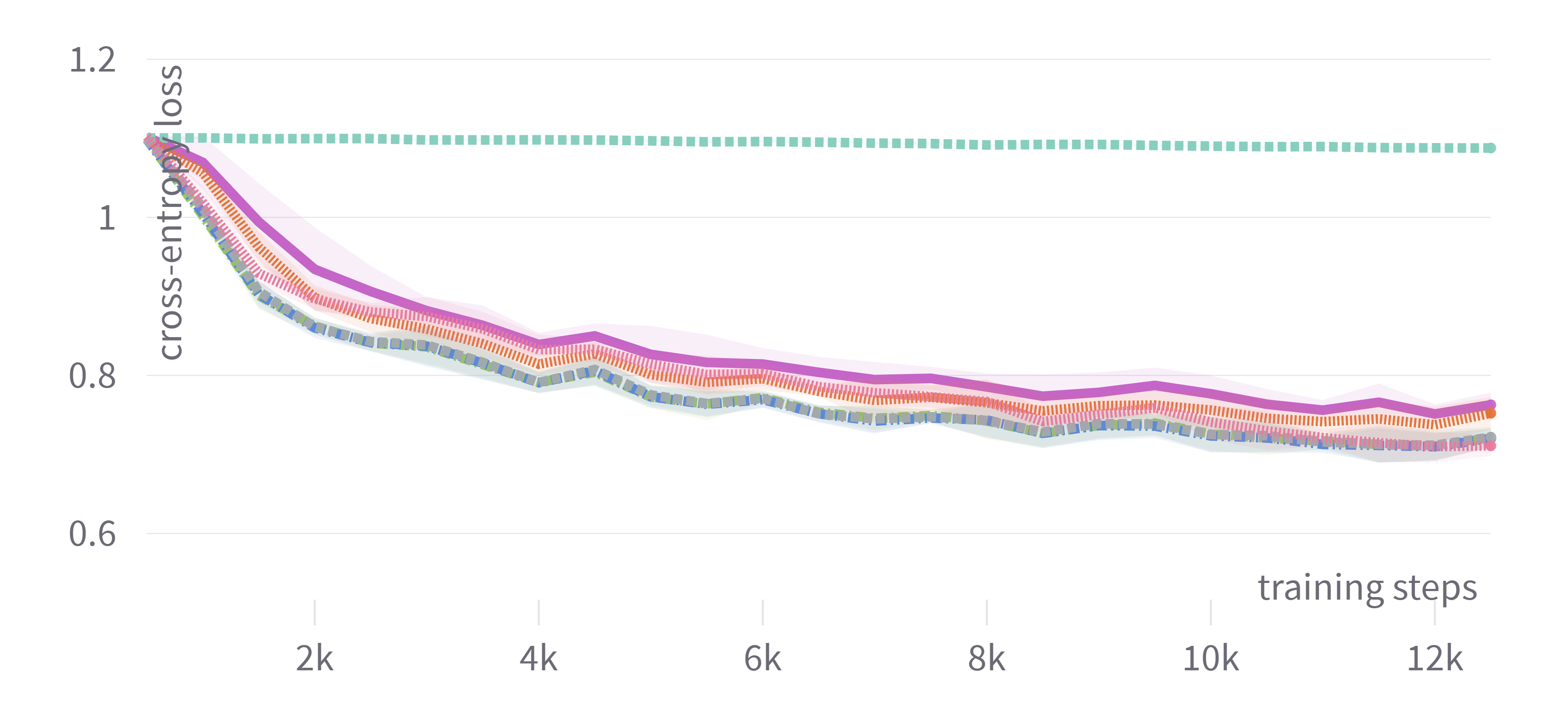}
        \caption{MNLI}
        \label{subfig:mnlilosstuned}
    \end{subfigure}
    \hfill\addtocounter{subfigure}{-1}
    \begin{subfigure}[b]{0.49\textwidth}
        \centering
        \includegraphics[width=\textwidth]{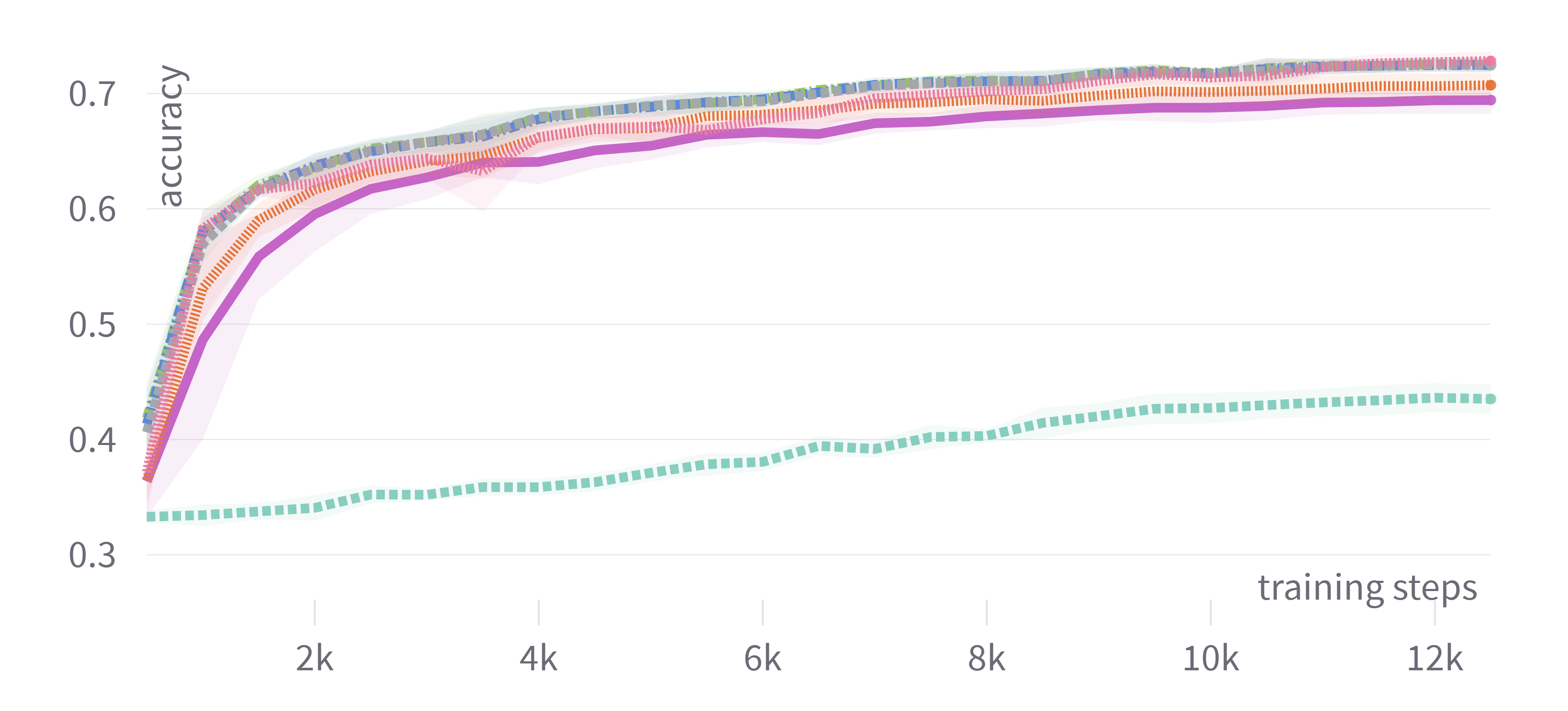}
        \caption{MNLI}
        \label{subfig:mnliacctuned}
    \end{subfigure}

\vspace*{-2mm}
\caption{\textbf{Training loss} (left) and \textbf{evaluation score on development data} (right) with \textbf{all hyperparameters of the optimizers tuned}, as a function of training steps, using \textbf{DistilBERT}. For each dataset, we use \textbf{five random data splits}, and plot the \textbf{average} and \textbf{standard deviation} (shadow) over the five splits. \textbf{Plain SGD} is clearly the \textbf{worst}, but adding \textbf{Momentum (SGDM)} turns it to a \textbf{competent} optimizer in terms of development scores, except for CoLA. The five \textbf{adaptive optimizers} (Adam, Nadam, AdamW, AdaMax, AdaBound) have \textbf{almost identical development score curves} across the tasks, \textbf{despite occasional differences in the training losses} they reach.
}
\label{fig:tuned_curves}

\end{figure*}

\begin{table*}[htbp]
\centering
{\small
\begin{tabular}
{|l|c|c|c|c|c|}
 \hline
   & SST-2 & MRPC & CoLA & STS-B & MNLI\\
   Optimizer & Accuracy & Macro-F1 & Matthews & Pearson & Accuracy \\
    \hline 
  AdaBound  &$91.61 \; (1.05)$  &$ 81.34 \; (1.17)$ &$\textbf{0.53} \; (0.03) $ &$\textbf{0.87} \; (0.01)  $&$\textbf{72.33}   \; (0.35)$  \\
 AdamW&$ \textbf{91.93}  \; (0.73) $  &$81.01  \; (1.09)  $  &$0.51 \; (0.05)  $  &$ \textbf{0.87} \; (0.01) $  &$ 72.17 \; (0.38)$  \\
 AdaMax &$ 91.73  \; (1.10) $  &$80.88  \; (0.68)   $  &$0.51 \; (0.07)   $  &$0.86 \; (0.01)  $  &$70.44 \; (0.70)$  \\
 Nadam   &$ 91.76  \; (0.97) $  &$ \textbf{81.85}  \; (3.75) $  &$\textbf{0.53} \; (0.03)   $  &$ 0.86 \; (0.01) $  &$72.04 \; (0.36)$  \\
  Adam  &$ 91.63  \; (0.75) $  &$80.81  \; (1.83)  $  &$ 0.52 \; (0.03)  $  &$ \textbf{0.87} \; (0.01) $  &$72.08 \; (0.37)$  \\
SGDM  &$90.53  \; (1.78)   $  &$79.51  \; (1.58) $  &$0.22 \; (0.26) $  &$\textbf{0.87} \; (0.01)  $  &$ 68.84 \; (1.62)$ \\
SGD  &$86.17  \; (1.16)  $  &$65.45  \; (9.94)  $  &$ 0.09 \; (0.14)  $  &$  0.74 \; (0.03)$  &$43.64 \; (0.65)$  \\
  \hline
\end{tabular}
} 
\centering
\caption{Evaluation scores on \textbf{test data} with \textbf{all hyperparameters}  of the optimizers \textbf{tuned}, using \textbf{DistilBERT}. For each dataset, we use \textbf{five random splits} of training/development/test data (the same for all optimizers) and report the \textbf{average} test score and the \textbf{standard deviation} over the five splits. \textbf{Plain SGD} is clearly the \textbf{worst}, but \textbf{SGDM} is \textbf{competitive}, except for CoLA. The \textbf{five adaptive optimizers} (top five) \textbf{all perform very similarly}. 
}
\label{table:tunedtest}
\end{table*}

\begin{table*}[htb]
\centering
{\small
\begin{tabular}
{|l|c|c|c|c|c|}
 \hline
   & SST-2 & MRPC & CoLA & STS-B & MNLI\\
   Optimizer & Accuracy & Macro-F1 & Matthews & Pearson & Accuracy \\
    \hline 
  AdaBound  & $  91.40 \; (1.16) $ &$  \textbf{81.59} \; (1.98) $& $  0.49 \; (0.06) $ &$  \textbf{0.87} \; (0.01) $ &$  \textbf{72.74} \; (0.85) $ \\
 AdamW& $  91.87 \; (0.88) $ &$  81.32 \; (1.75) $ &$  0.54 \; (0.03) $ &$  \textbf{0.87} \; (0.01) $ &$  72.20 \; (0.28) $ \\
 AdaMax& $  89.52 \; (1.16) $ &$  81.12 \; (0.59) $&$  0.48 \; (0.07) $ &$  0.84 \; (0.01) $ & $  66.80 \; (0.28) $ \\
 Nadam &$  \textbf{92.79} \; (0.37) $ &$  80.91 \; (1.69) $ &$  \textbf{0.55} \; (0.04) $ &$  \textbf{0.87} \; (0.01) $ &$  72.36 \; (0.37) $  \\
  Adam &$  91.75 \; (0.91) $ &$  80.81 \; (1.83) $ &$  0.52 \; (0.02) $ & $  0.86 \; (0.01) $ &$  72.19 \; (0.27) $\\
SGDM & $  89.79 \; (1.23) $ &$  80.79 \; (0.79) $&$  0.46 \; (0.10) $ &$  0.85 \; (0.01) $ &$  67.31 \; (0.61) $  \\
SGD  &$86.17  \; (1.16)  $  &$65.45  \; (9.94)  $  &$ 0.09 \; (0.14)  $  &$  0.74 \; (0.03)$  &$43.64 \; (0.65)$  \\
  \hline
\end{tabular}
} 
\centering
\caption{Evaluation scores on \textbf{test data}, having \textbf{tuned only the learning rate}, with all other hyperparameters of the optimizers set to their defaults, using \textbf{DistilBERT}. Again, we use five random splits and report the average and standard deviation. \textbf{In most cases, tuning only the learning rate leads to very similar results as tuning all the hyperparameters} (cf.\ Table~\ref{table:tunedtest}). 
Exceptions include AdaMax, which now lags behind on  CoLA and (more noticeably) on MNLI, as well as  AdaBound, whose performance deteriorates on CoLA (comparing to Table~\ref{table:tunedtest}). Interestingly, 
SGDM is now competitive to the five adaptive optimizers on CoLA (where it lagged behind in Table~\ref{table:tunedtest}). 
}
\label{table:tunedlrtest}
\end{table*}

\subsection{Experimental Results}
We include in the main paper only results using DistilBERT. DistilRoBERTa results are reported in Appendix~\ref{app:additional_results}, and lead to the same conclusions. 

Figure~\ref{fig:tuned_curves} shows the training loss for each task and optimizer (left column) and the corresponding evaluation score on development data (right column), as a function of training steps, using DistilBERT, when all the hyperparameters of the optimizers are tuned. For each curve, we plot the average and standard deviation (shadow) over the five data splits (Section~\ref{subsec:data}). SGD clearly struggles to learn the training data in all five tasks (left), which is also reflected in its development scores (right). However, adding Momentum turns it to a competent optimizer (SGDM) in terms of development scores. An exception is CoLA, where SGDM is clearly worse in development score than the five more elaborate (adaptive) optimizers (Adam, Nadam, AdamW, AdaMax, and AdaBound), in accordance with its poor training loss, though it still outpeforms SGD. The adaptive optimizers have almost identical development score curves across the tasks, with some minor differences in CoLA where AdaMax and AdamW are slightly worse, again reflecting their inferior training losses. Otherwise, differences in the training losses reached by the five adaptive optimizers (when there are any) do not lead to visible differences in development scores.\footnote{Curves for the cases where we tune only the learning rate or use the defaults can be found in Appendix~\ref{app:additional_results}.}

Table~\ref{table:tunedtest} tells a similar story, now evaluating on test data, again using DistilBERT. Again, SGD is clearly the worst, but SGDM is competitive, except for CoLA. The five adaptive optimizers all perform very similarly. Interestingly, in most cases tuning only the learning rate (Table~\ref{table:tunedlrtest}) leads to very similar results as (and is much cheaper than) tuning all the hyperparameters (Table~\ref{table:tunedtest}). Some exceptions are reported in the caption of Table~\ref{table:tunedlrtest}.

\begin{table*}[htb]
\centering
{\small 
\begin{tabular}
{|l|c|c|c|c|c|}
 \hline
   & SST-2 & MRPC & CoLA & STS-B & MNLI\\
   Optimizer & Accuracy & Macro-F1 & Matthews & Pearson & Accuracy \\
    \hline 
  AdaBound  &$ \textbf{90.88} \; (1.41) $  &$ 76.14 \; (2.63) $  &$0.19 \; (0.26)  $  &$ \textbf{0.86} \; (0.01) $  &$65.58 \; (8.48) $  \\
 AdamW   &$ 55.83 \; (0.06) $  &$ 62.57 \; (0.00) $  &$0.00  \; (0.00) $  &$ 0.23 \; (0.22) $  &$ 35.34 \; (0.01)$  \\
 AdaMax   &$ 59.49 \; (7.35) $  &$62.57 \; (0.00)  $  &$ 0.00  \; (0.00) $  &$ 0.50 \; (0.15) $  &$35.34 \; (0.01) $  \\
 Nadam    &$ 55.80 \; (0.00) $  &$ 62.57  \; (0.00) $  &$ 0.00  \; (0.00)  $  &$0.14 \; (0.11)$  &$ 35.34 \; (0.01)$  \\
  Adam    &$ 55.83 \; (0.06) $  &$62.57 \; (0.00)  $  &$ 0.00  \; (0.00)  $  &$  0.27 \; (0.17)$  &$35.34 \; (0.01)$ \\
SGDM     &$ 89.89 \; (1.13) $  &$\textbf{80.00}  \; (0.51) $  &$\textbf{0.49} \; (0.03)  $  &$ \textbf{0.86} \; (0.01) $  &$ \textbf{67.56} \; (0.24) $ \\
SGD    &$ 86.35 \; (0.96) $  &$ 70.90 \; (1.59) $  &$ 0.33 \; (0.04) $  &$ 0.74 \; (0.03) $  &$ 44.65 \; (0.20)$  \\
  \hline
\end{tabular}
} 
\centering
\caption{Evaluation scores on \textbf{test data}, with \textbf{all hyperparameters} of the optimizers set to \textbf{defaults}, using \textbf{DistilBERT}. Again, we use five random splits and report the average and standard deviation. \textbf{SGDM is not affected} by the lack of hyperparameter tuning (cf.\ Table~\ref{table:tunedlrtest}) and is \textbf{now the best} overall. Plain SGD is also not particularly affected, and actually performs much better on MRPC and CoLA untuned (cf.\ Table~\ref{table:tunedlrtest}). Although negatively affected by the use of defaults, \textbf{AdaBound} is now the best optimizer on SST-2 and STS-B, and much better than the other four adaptive optimizers (but worse than SGDM) on the other datasets.
}
\label{table:defaulttest}
\end{table*}

Table~\ref{table:defaulttest} shows test results with default hyperparameter values, again using DistilBERT. SGDM is not affected by the lack of hyperparameter tuning (cf.\ Table~\ref{table:tunedlrtest}) and is now the best overall. Plain SGD is also not particularly affected, and actually performs much better on MRPC and CoLA untuned (cf.\ Table~\ref{table:tunedlrtest}). Overall, it seems that the defaults of the two non-adaptive optimizers are good global choices; tuning their hyperparameters occasionally overfits the development data. By contrast, the adaptive optimizers are negatively affected by the lack of tuning. AdaBound is the least affected and is now (Table~\ref{table:defaulttest}) overall the second best and clearly better than the other adaptive optimizers. This may be related to the fact that AdaBound behaves similarly to SGD at the end of training (Section~\ref{sec:optimizers}). All seven optimizers have very similar default learning rates (Appendix~\ref{sec:hyperparameter}), hence the superior out-of-the-box 
performance of SGDM and AdaBound is not due to different default learning rates.

Trying multiple optimizers with defaults \cite{valley} is competitive too. Based on development scores (Fig.~\ref{fig:default_curves}), one would select the same optimizers (per task) whose (test) scores are shown in bold in Table~\ref{table:defaulttest}. The resulting test scores are competitive, but worse than the best scores of Tables~\ref{table:tunedtest}--\ref{table:tunedlrtest}. Also, the competitive scores of trying multiple optimizers with defaults are due only to the good out-of-the-box performance of SGDM and (to a lesser extent) AdaBound; the other optimizers perform much worse with defaults, hence trying them would have been a waste of resources. 

Therefore, based on our experiments, we recommend picking just one adaptive optimizer and tuning only its learning rate.  Among the adaptive optimizers we considered, we recommend picking AdamW, Nadam, or Adam, since AdaBound and AdaMax were not top performers across all datasets when tuning only the learning rate (Tables~\ref{table:tunedlrtest}, \ref{table:distilrobertatunedlrtest}). 

\section{Related Work}

DeepOBS \cite{DeepOBS} is an optimizer benchmarking suite that includes several classical datasets, models (e.g., CNNs, RNNs), optimizer implementations (currently SGD, SGDM, Adam), and facilities to compare optimizers. However, it includes only one NLP task (character-level language modeling with an RNN) and no  Transformer models. A similar observation can be made for the more recent AlgoPerf benchmark \cite{DahlBenchmark}, which includes Transformers, but only one NLP task (machine translation). MultiTask \cite{MultiTask} also considers only three NLP tasks (language modelling with characters or words/sub-words, text classification), all with RNNs.

As already noted, we were inspired by \citet{valley}, who experimented with 15 optimizers and 8 tasks (from DeepOBS), but only one NLP task (the only one of DeepOBS), without considering Transformers. They found that Adam remained a strong contender, with more recent variants failing to consistently outperform it. Tuning the hyperparameters of a single optimizer was overall only slightly better than using its default hyperparameter values (median improvement 3.4\% for a tuning budget of 50 trials and diminishing returns for larger budgets). Trying several optimizers with defaults was almost as beneficial as (and cheaper than) picking any single (competent) optimizer and tuning it. \citeauthor{valley} acknowledged, however, that their findings may not hold with more complicated models, such as Transformers. They also found indications that the best optimizer may depend on the model and task. They employed random search for hyperparameter tuning, whereas we used Optuna (Section~\ref{subsec:setup}). They also experimented with four update schedules for the learning rate (constant, cosine decay, cosine with warm restarts, and trapezoidal) on top of the tuned rate, with results indicating that non-constant schedules add small gains; we considered only a constant schedule.

\citet{marg} reported that adaptive optimizers may lead to worse development or test performance than SGD, even in cases where the adaptive optimizers reach lower training losses. They considered artificial datasets, an image classification task (using a CNN), character-level language modelling, and two parsing tasks, the latter three tasks using LSTM-based models. However, they only tuned the learning rate and the learning rate decay scheme, as pointed out by \citet{onemp}.

\citet{onemp} introduced the notion of \emph{inclusion} between optimizers. For example, SGD is a special case of (is included by) SGDM for $\alpha = 0$ (Algorithm~\ref{pgd}). With an exhaustive hyperparameter search, an optimizer should never perform worse than an optimizer it includes. Indeed, \citeauthor{onemp} show that with extensive hyperparameter tuning, inclusion relationships reflect end-task performance, and they criticize previous work by \citet{marg} and \citet{DeepOBS} for not having tuned all hyperparameters. We tuned all hyperparameters, but found that tuning only the learning rate was equally good. We also note that extensive tuning of the kind recommended by \citeauthor{onemp} (e.g., with coarser to finer swaps to explore and define the search space anew per task) is computationally very expensive. Finally, we note that \citet{onemp} considered only image classification and language modelling, the latter with an LSTM and a (not pre-trained) Transformer.

\citet{bench} pointed out that when comparing optimizers, it is important to consider how easy it is to reach reasonable  performance with a limited budget (number of trials), rather than comparing performance scores obtained with very extensive (and costly) hyperparameter tuning, unlike the setting of \citet{onemp}. They experimented with SGDM, Adagrad \cite{adagrad}, Adam, and AdamW, with a range of budgets, in nine tasks, of which only two were NLP tasks (sentiment analysis, news classification), without considering Transformers. They recommended using Adam and tuning only its learning rate, especially with low budgets, which agrees with our conclusions. Although we experimented with a single, relatively small budget (30 trials), we used the same budget for all optimizers, like \citet{bench}~ Their evaluation protocol, which efficiently simulates multiple budgets, could be used in future extensions of our work, though it is incompatible with our use of Optuna, as it requires random search. 

A particularly interesting research direction is to semi-automatically discover new optimization algorithms via evolutionary program searchs and manual intervention. \citet{Lion} recently used this approach to produce Lion, an optimizer that, among other experiments, was reported to be overall slightly better than AdamW on GLUE, when using the T5 model \cite{T5}, but tuning only the learning rates and decoupled weight decay hyperparameters of the two optimizers.

The optimizers we considered are first-order, i.e., they compute only the gradient of the loss function and not its Hessian (second-order partial derivatives), which would be prohibitively costly ($M^2$ second-order derivatives at each step, for a model with $M$ weights). Interestingly, \citet{liu2023sophia} investigate an \emph{approximate} second-order optimizer for use in language model pretraining. 

Furthermore, the optimizers we considered use a single value of the learning rate at each step. \emph{Backtracking} methods consider multiple learning rate values at each step, computing the loss for each one. Although backtracking is very common in traditional optimization, apparently it has not received sufficient attention in machine learning; an exception is the work of \citet{truong2021backtracking}.

\section{Conclusions}

We investigated if it is worth (a) trying multiple optimizers and/or (b) tuning their hyperparameters (and which ones), when fine-tuning a pre-trained Transformer. We experimented with five GLUE datasets, two efficient pre-trained Transformer encoders (DistilBERT, DistilRoBERTa), and seven popular optimizers (SGD, SGDM, Adam, AdaMax, Nadam, AdamW, and AdaBound). With the exception of the two non-adaptive optimizers (SGD, SGDM), which were largely unaffected by hyperparameter tuning, the test performance of the other five (adaptive) optimizers improved substantially when they were tuned, unlike previously reported smaller overall 
gains \cite{valley}. In most cases, tuning only the learning rate was as good as (and cheaper than) tuning all the hyperparameters. Furthermore, when hyperparameters (or just the learning rate) were tuned, all the adaptive optimizers had very similar test scores, unlike SGD and SGDM, which were clearly the worst and second worst, respectively. 
This parity of test performance of the adaptive optimizers was obtained despite occasional differences in the training loss they reached. When no hyperparameter was tuned (which might be the case with a low budget), SGDM was the best choice and AdaBound the second best; the other optimizers were much worse. Trying multiple optimizers with defaults \cite{valley} worked reasonably well too, but only because of the good untuned performance of SGDM and (to a lesser extent) AdaBound; trying the other optimizers untuned would have been a waste of resources. Hence, we suggest picking just one adaptive optimizer and tuning only its learning rate; we recommend AdamW, Nadam, or Adam, which were consistently top performers when tuning only the learning rate. 

Our work can help save substantial effort, computational resources, and energy, by reducing the number of hyperparameter tuning experiments practitioners perform. However, our findings need to be complemented by future additional experiments with more models, more NLP tasks (including pre-training tasks), and different tuning budgets (i.e., different maximum number of trials); see also Section~\ref{sec:limitations}. This is a computationally very expensive exploration that might be best handled by multiple groups performing and reporting similar studies. Towards this direction, we make all the code, data, and results of our experiments publicly available.\footnote{\url{https://github.com/nlpaueb/nlp-optimizers}}

\section{Limitations} \label{sec:limitations}
We examined different optimizers when \emph{fine-tuning} pre-trained Transformers for NLP downstream tasks. Given our limited resources, we did not consider \emph{pre-training} and we experimented only with two lightweight encoder-only models, i.e., DistilBERT and DistilRoBERTa. Thus, we limited ourselves to five \emph{classification} datasets for single word sequences or pairs of sequences (Section~\ref{subsec:data}), and did not consider sequence-to-sequence or sequence generation tasks, which would require encoder-decoder or decoder-only models. Also, given our limited resources, we considered seven popular optimizers (Section~\ref{sec:optimizers}), among many more available.

During hyperparameter tuning, we restricted ourselves to the direct hyperparameters of the optimizers. Thus, we did not consider tuning the batch size, although it often affects the choice of learning rate.\footnote{We used a fixed batch size of 4 in all experiments.} Also, given that the effect of a non-constant learning rate may vary substantially with respect to the optimizer and task \cite{valley}, we experimented with a constant (but tuned) learning rate only, leaving the investigation of other update schedules (e.g., cosine decay) for future work.

Finally, we did not measure how the training speed is affected by the choice of optimizer and hyperparameter tuning \cite{DeepOBS}. In our work, the training speed can only be indirectly inferred from the learning curves (Fig.~\ref{fig:tuned_curves}--\ref{fig:distilroberta_default_curves}) and depends on the size of each dataset. We also did not vary the tuning budget; we used a fixed budget of 30 trials in all experiments, which is close to the `small' budget (25 trials) of \citet{valley}.

\section{Ethical considerations}
Selecting the best optimization scheme, which includes the choice of optimizer and tuning its hyperparameters, is much more expensive than training the final model with tuned hyperparameters \cite{MultiTask}. Thus, selecting the best optimization scheme has a large economic and environmental impact, which could be greatly reduced with the help of proper guidelines to narrow the search space without compromising performance. Of course, the effort to develop such guidelines (which includes our work) also requires substantial resources. However, by making code, data, and results publicly available (as we do), we believe that this effort will help save significant resources in the long run.

\section*{Acknowledgements}
This work was partially supported by project MIS 5154714 of the National Recovery and Resilience Plan Greece 2.0 funded by the European Union under the NextGenerationEU Program. It was also supported by the TPU Research Cloud (TRC) program of Google.\footnote{\url{https://sites.research.google/trc/about/}}

\bibliography{custom}

\section*{Appendix}
\appendix

\section{Hyperparameter Tuning}\label{sec:hyperparameter}

Tables~\ref{table:hypersearch}--\ref{table:only_lr} show the \emph{search space}, \emph{default values}, and \emph{tuned values} (averaged over the five random splits) of the hyperparameters across the optimizers and datasets, when using DistilBERT.\footnote{The tuned hyperparameter values for DistilRoBERTa will be available in our code repo and lead to similar conclusions.} In most cases, the tuned hyperparameter values are different than the defaults. Regarding the learning rate ($\epsilon$), as already noted in Section~\ref{subsec:setup}, the search space we use for adaptive optimizers does not include the default values (see Tables~\ref{table:hypersearch}--\ref{table:lr}), because it is standard practice when fine-tuning Transformers with adaptive optimizers to use much smaller learning rates. Nevertheless, we observe (Tables~\ref{table:lr}, \ref{table:only_lr}) that the tuned learning rates of SGDM and SGD are closer to the default, compared to the other optimizers, which may explain why SGDM is the best optimizer overall when using the defaults.

\section{Additional Results}\label{app:additional_results}

\begin{figure*}
\centering

\includegraphics[width=0.7\textwidth]{optimizers.png}

    \begin{subfigure}[b]{0.49\textwidth}
        \centering
        \includegraphics[width=\textwidth]{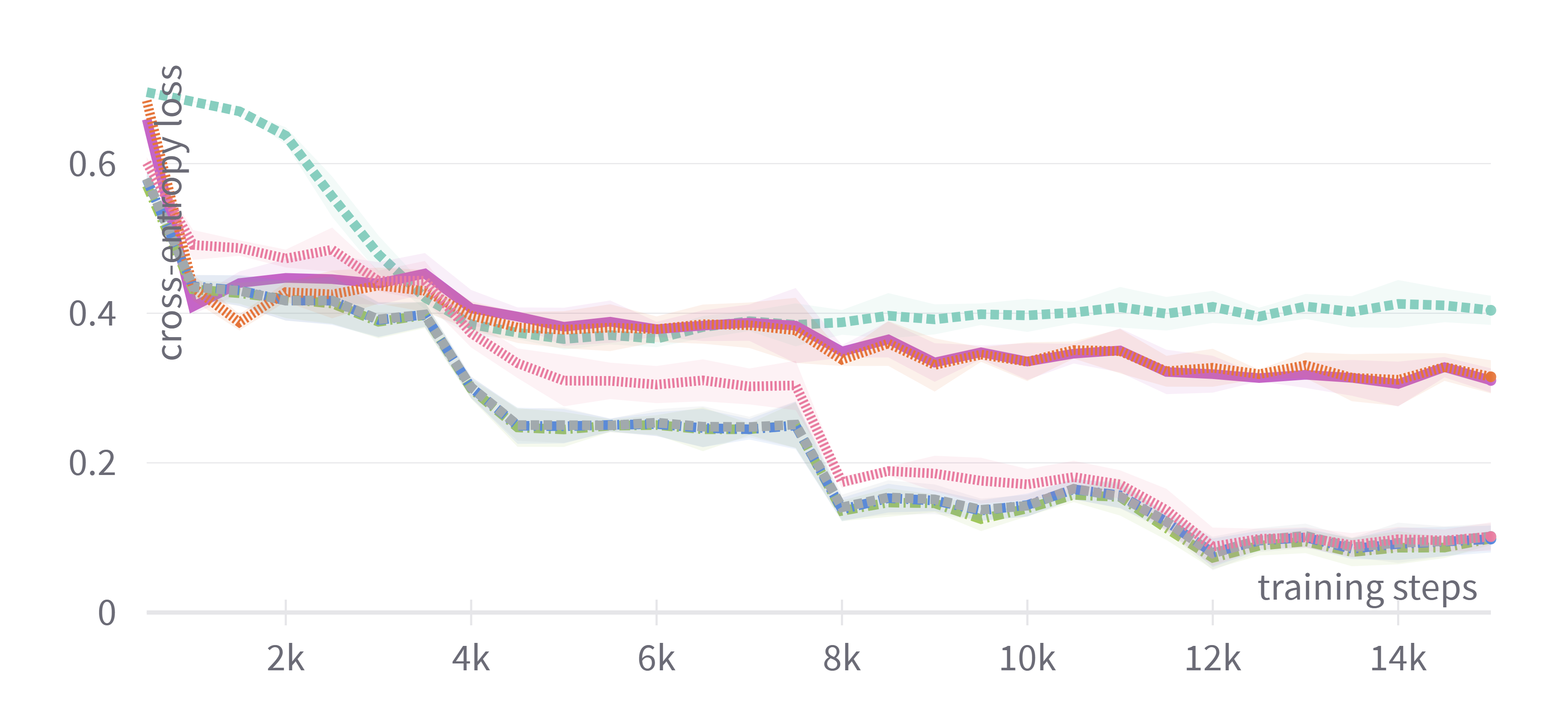}
        \caption{SST-2}
        \label{subfig:sst2losslrtuned}
        \vspace{0.5mm}
    \end{subfigure}
    \hfill\addtocounter{subfigure}{-1}
    \begin{subfigure}[b]{0.49\textwidth}
        \centering
        \includegraphics[width=\textwidth]{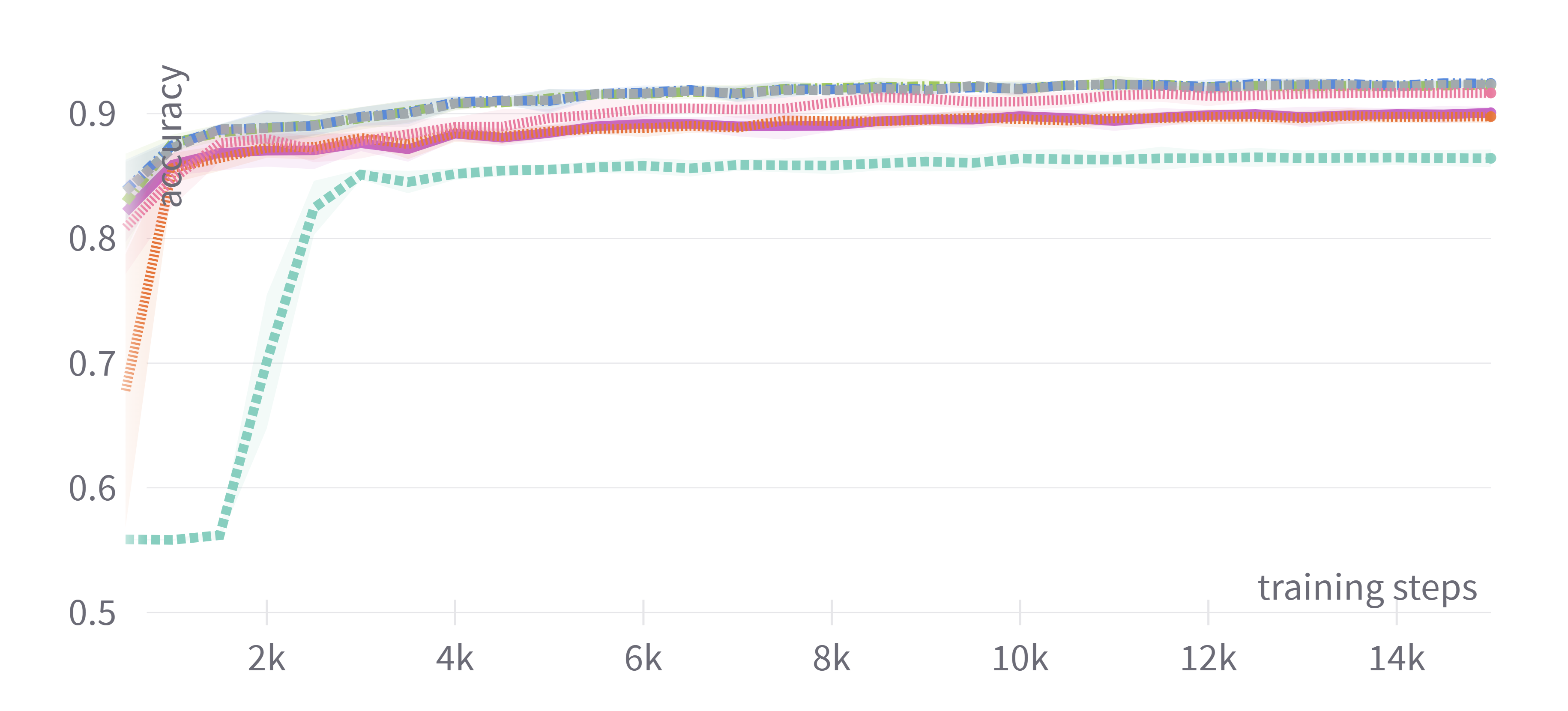}
        \caption{SST-2}
        \label{subfig:sst2lacclrtuned}
        \vspace{0.5mm}
    \end{subfigure}
    \hfill
    \begin{subfigure}[b]{0.49\textwidth}
        \centering
        \includegraphics[width=\columnwidth]{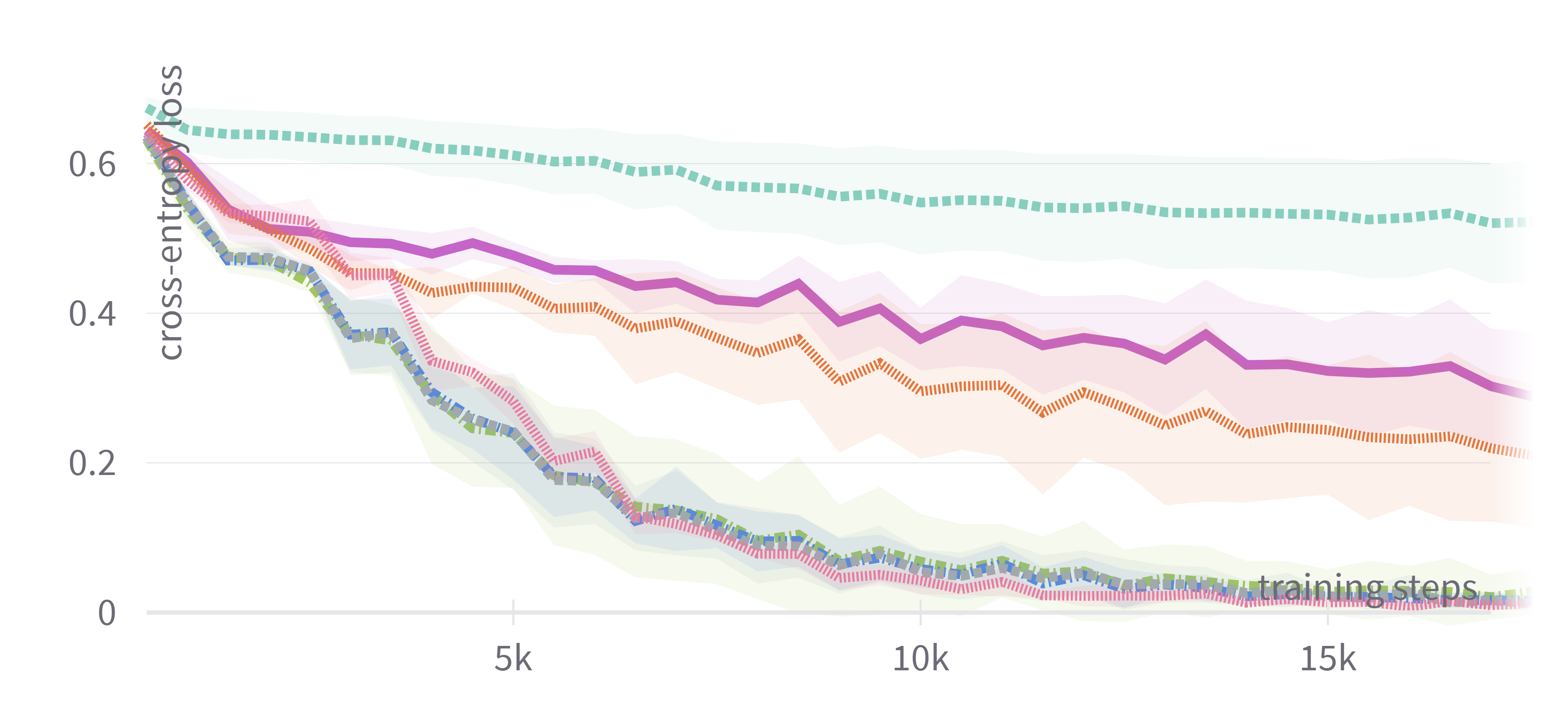}
        \caption{MRPC}
        \label{subfig:mrpclosslrtuned}
        \vspace{0.5mm}
    \end{subfigure}
    \hfill\addtocounter{subfigure}{-1}
    \begin{subfigure}[b]{0.49\textwidth}
        \centering
        \includegraphics[width=\textwidth]{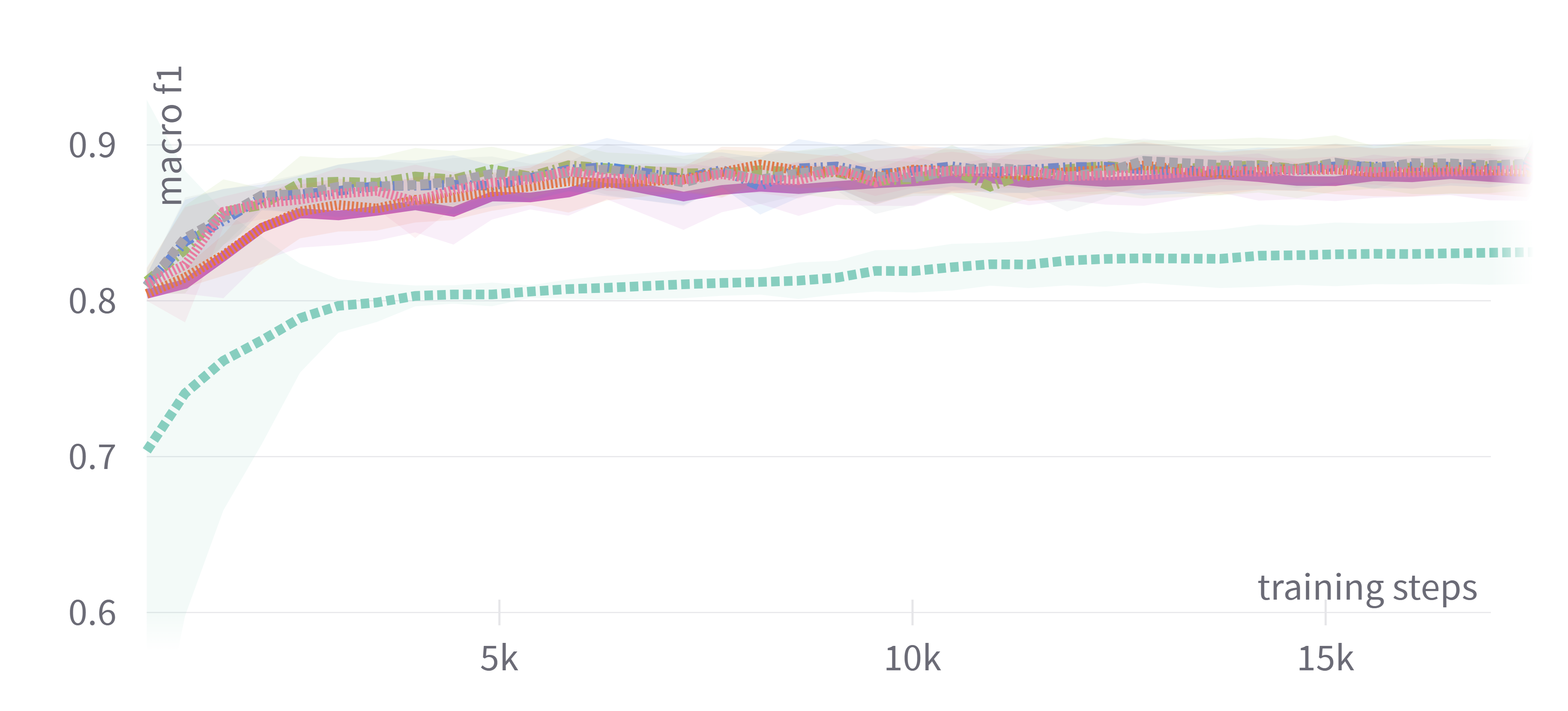}
        \caption{MRPC}
        \label{subfig:mrpcf1lrtuned}
        \vspace{0.5mm}
    \end{subfigure}
    \hfill
    \begin{subfigure}[b]{0.49\textwidth}
        \centering
        \includegraphics[width=\textwidth]{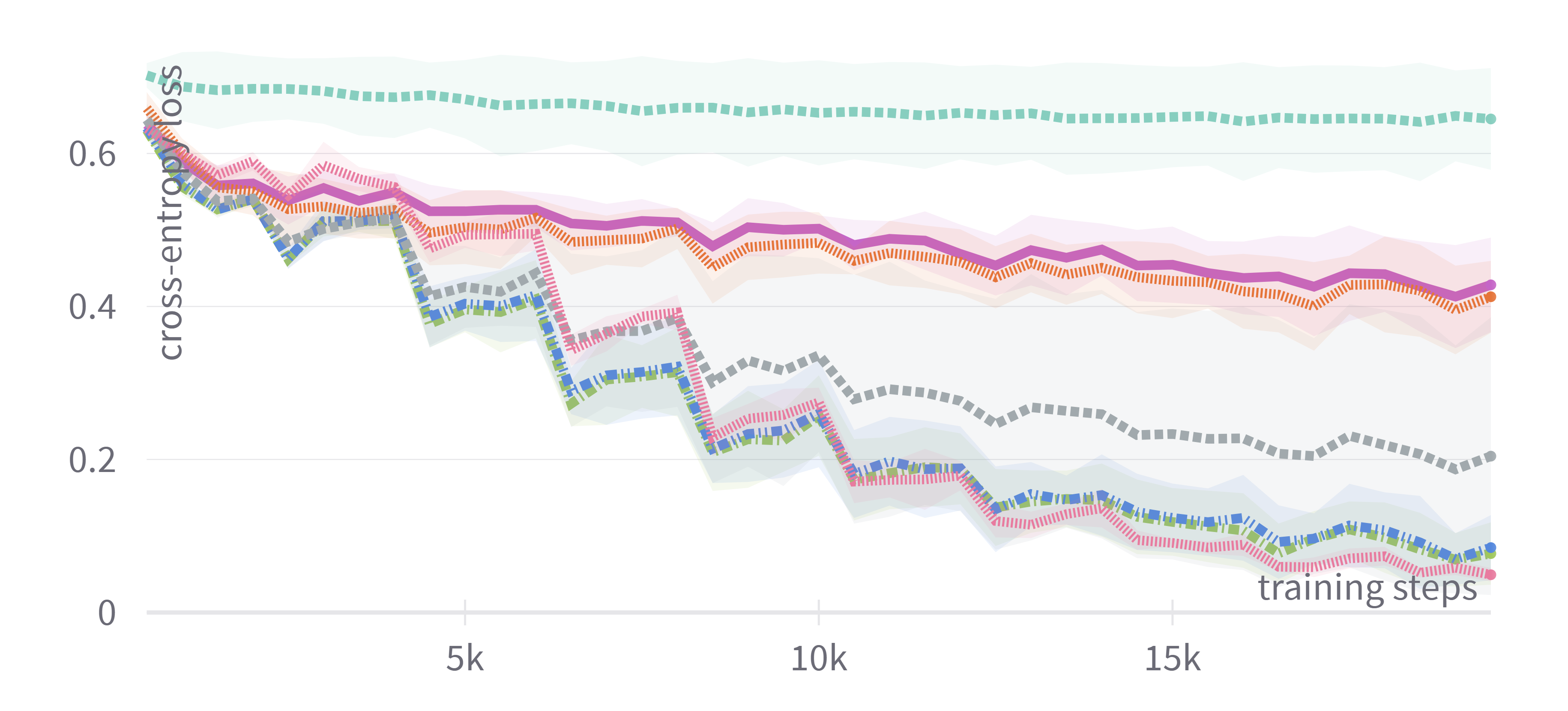}
        \caption{CoLA}
        \label{subfig:colalosslrtuned}
        \vspace{0.5mm}
    \end{subfigure}
    \hfill\addtocounter{subfigure}{-1}
    \begin{subfigure}[b]{0.49\textwidth}
        \centering
        \includegraphics[width=\textwidth]{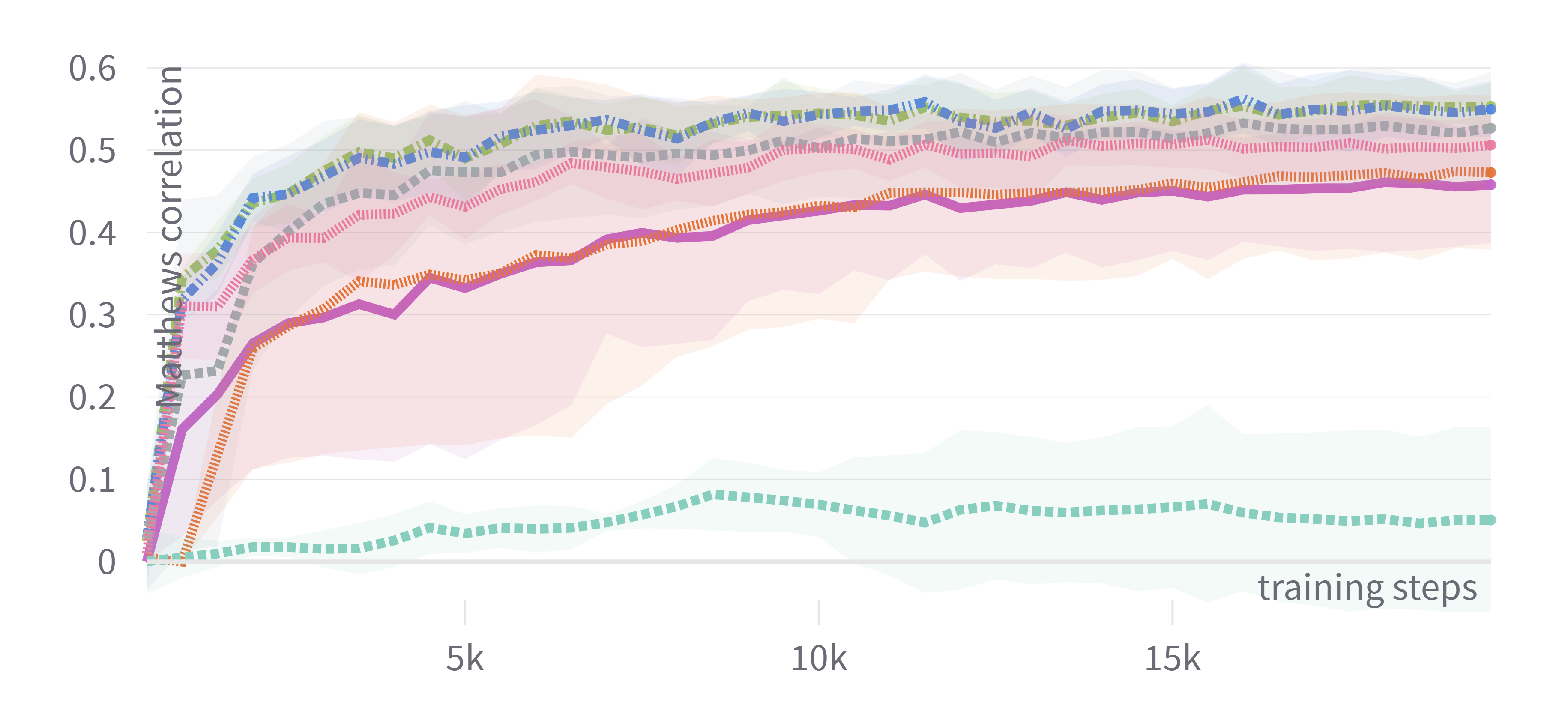}
        \caption{CoLA}
        \label{subfig:colamathlrtuned}
        \vspace{0.5mm}
    \end{subfigure}
    \hfill
    \begin{subfigure}[b]{0.49\textwidth}
        \centering
        \includegraphics[width=\textwidth]{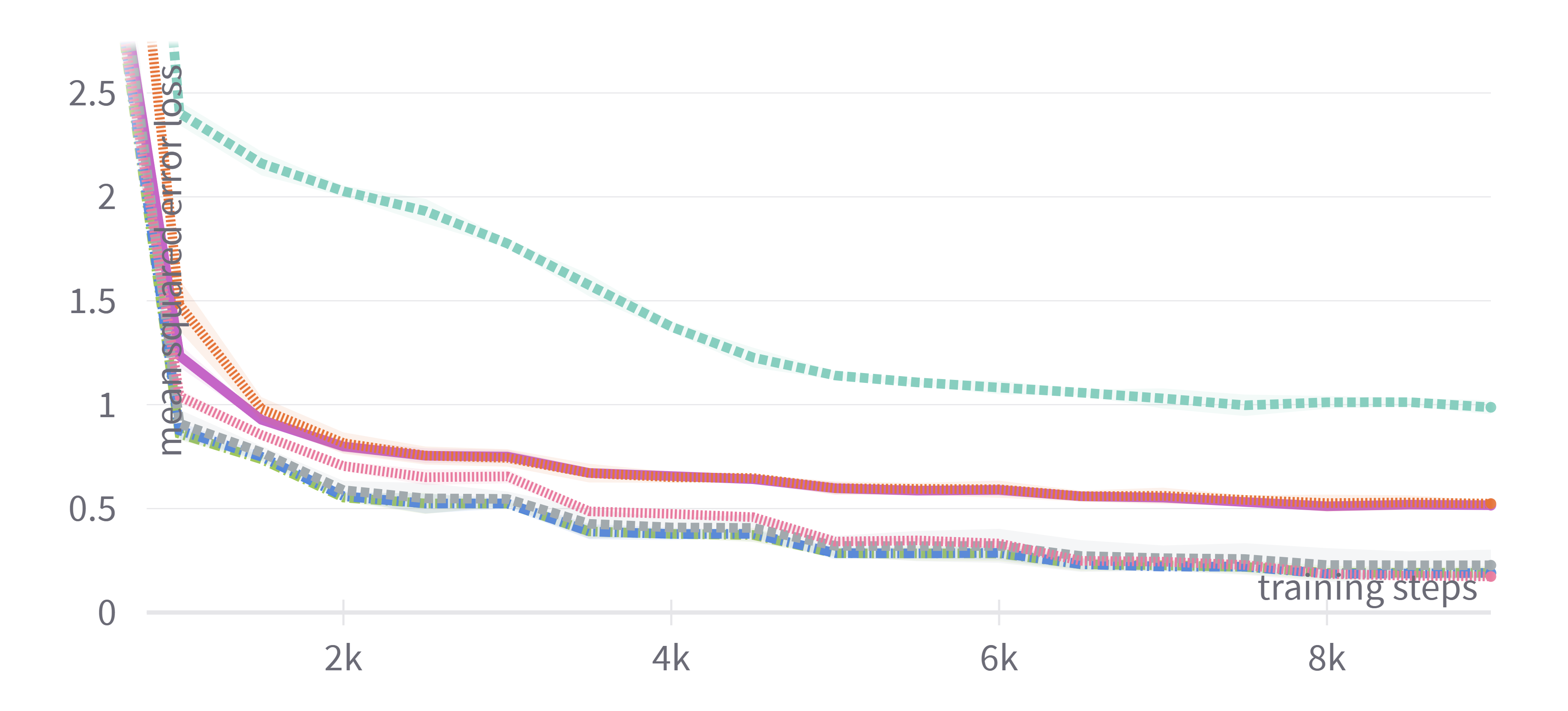}
        \caption{STS-B}
        \label{subfig:stsblosslrtuned}
        \vspace{0.5mm}
    \end{subfigure}
    \hfill\addtocounter{subfigure}{-1}
    \begin{subfigure}[b]{0.49\textwidth}
        \centering
        \includegraphics[width=\textwidth]{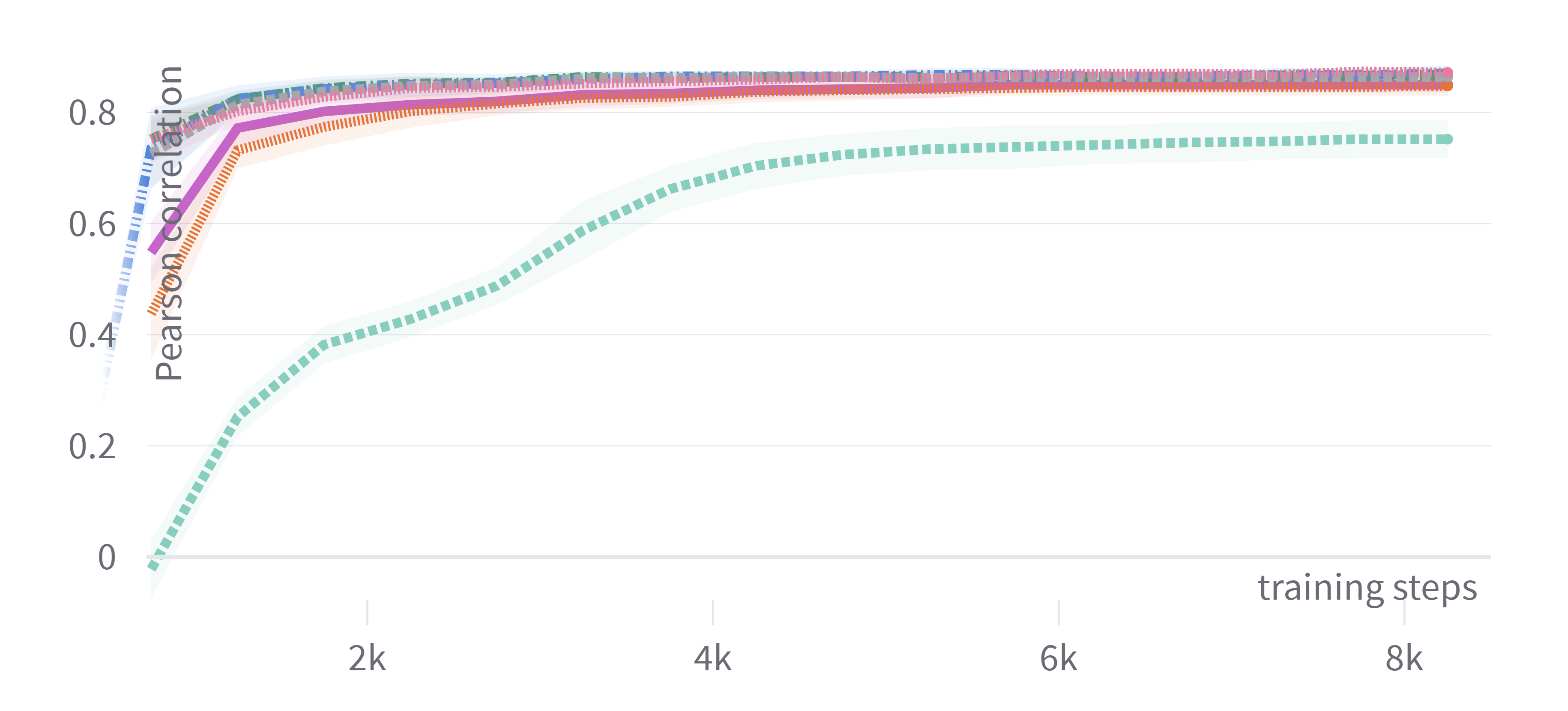}
        \caption{STS-B}
        \label{subfig:stsbpearlrtuned}
        \vspace{0.5mm}
    \end{subfigure}
    \hfill
    \begin{subfigure}[b]{0.49\textwidth}
        \centering
        \includegraphics[width=\textwidth]{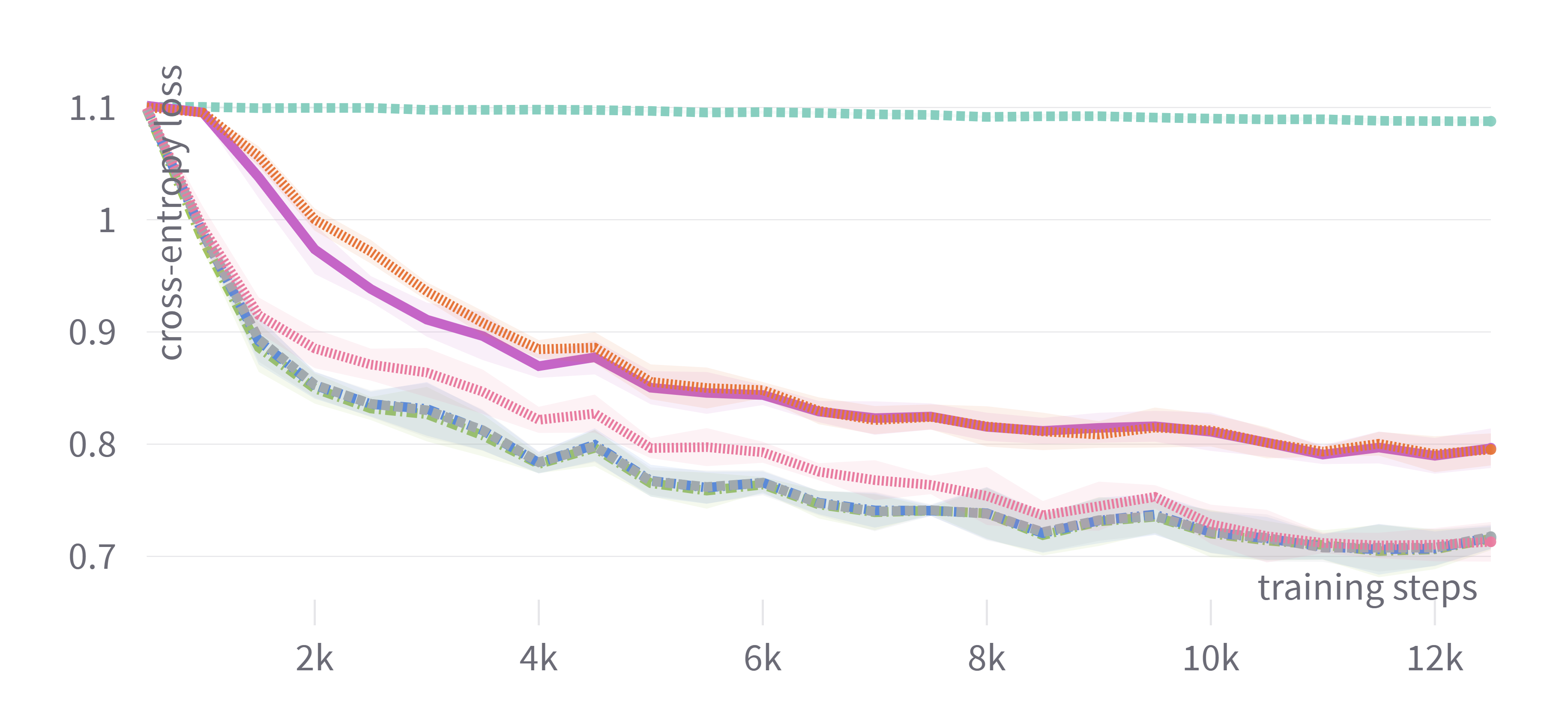}
        \caption{MNLI}
        \label{subfig:mnlilosslrtuned}
    \end{subfigure}
    \hfill\addtocounter{subfigure}{-1}
    \begin{subfigure}[b]{0.49\textwidth}
        \centering
        \includegraphics[width=\textwidth]{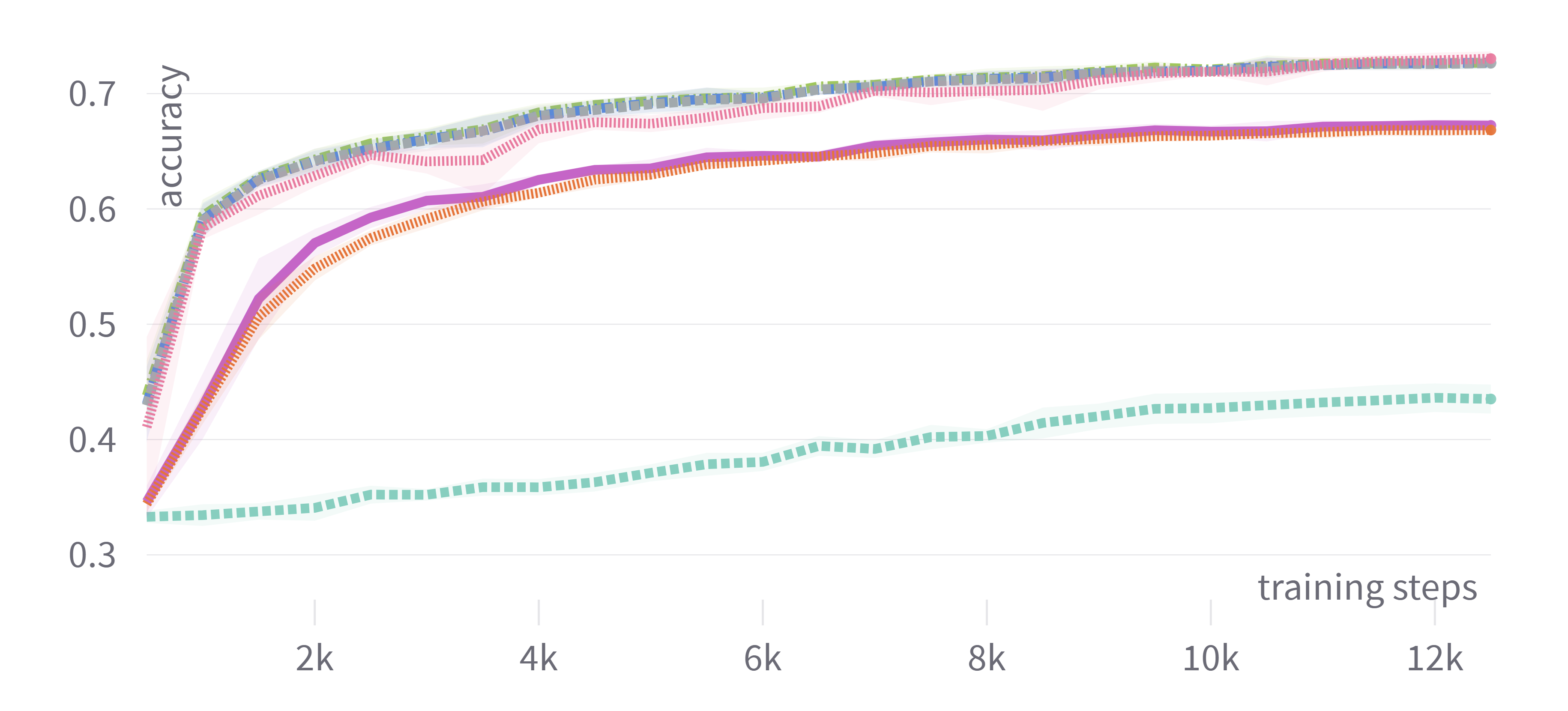}
        \caption{MNLI}
        \label{subfig:mnliacclrtuned}
    \end{subfigure}

\vspace*{-2mm}
\caption{\textbf{Training loss} (left) and \textbf{evaluation score on development data} (right) having \textbf{tuned only the learning rate of the optimizers}, as a function of training steps, using \textbf{DistilBERT}. For each dataset, we use \textbf{five random data splits}, and plot the \textbf{average} and \textbf{standard deviation} (shadow). As when tuning all the hyperparameters (Fig.~\ref{fig:tuned_curves}), SGD is clearly the \textbf{worst}, but adding \textbf{Momentum (SGDM)} turns it to a \textbf{competent} optimizer in terms of development scores. The five \textbf{adaptive optimizers} (Adam, Nadam, AdamW, AdaMax, AdaBound) \textbf{perform similarly overall} in terms of \textbf{development scores}, except for AdaMax which lags behind on CoLA and (more) on MNLI . Differences in training loss do not necessarily give rise to substantial differences in development scores.
}
\label{fig:tunedlr_curves}

\end{figure*}

Figure~\ref{fig:tunedlr_curves} shows the training loss per task and optimizer (left) and the corresponding evaluation score on development data (right), as a function of training steps, when \emph{only the learning rate} of each optimizer is tuned, using DistilBERT. For each curve, we plot the average and standard deviation (shadow) over the five data splits (Section~\ref{subsec:data}). As when tuning all hyperparameters (Fig.~\ref{fig:tuned_curves}), SGD struggles to learn the training data in all five tasks (left), which is also reflected in its development scores (right). Again, adding Momentum to SGD turns it to a competent optimizer (SGDM) in terms of development scores, now even on CoLA (cf.\ Fig.~\ref{fig:tuned_curves}). The five adaptive optimizers perform similarly overall in terms of development scores, except for AdaMax, which is visibly worse 
(along with SGDM) on CoLA and (to a larger extent) MNLI. Differences (when there are any) in the training loss of different optimizers do not always lead to substantial differences in development scores.

\begin{figure*}
\centering

\includegraphics[width=0.7\textwidth]{optimizers.png}

    \begin{subfigure}[b]{0.49\textwidth}
        \centering
        \includegraphics[width=\textwidth]{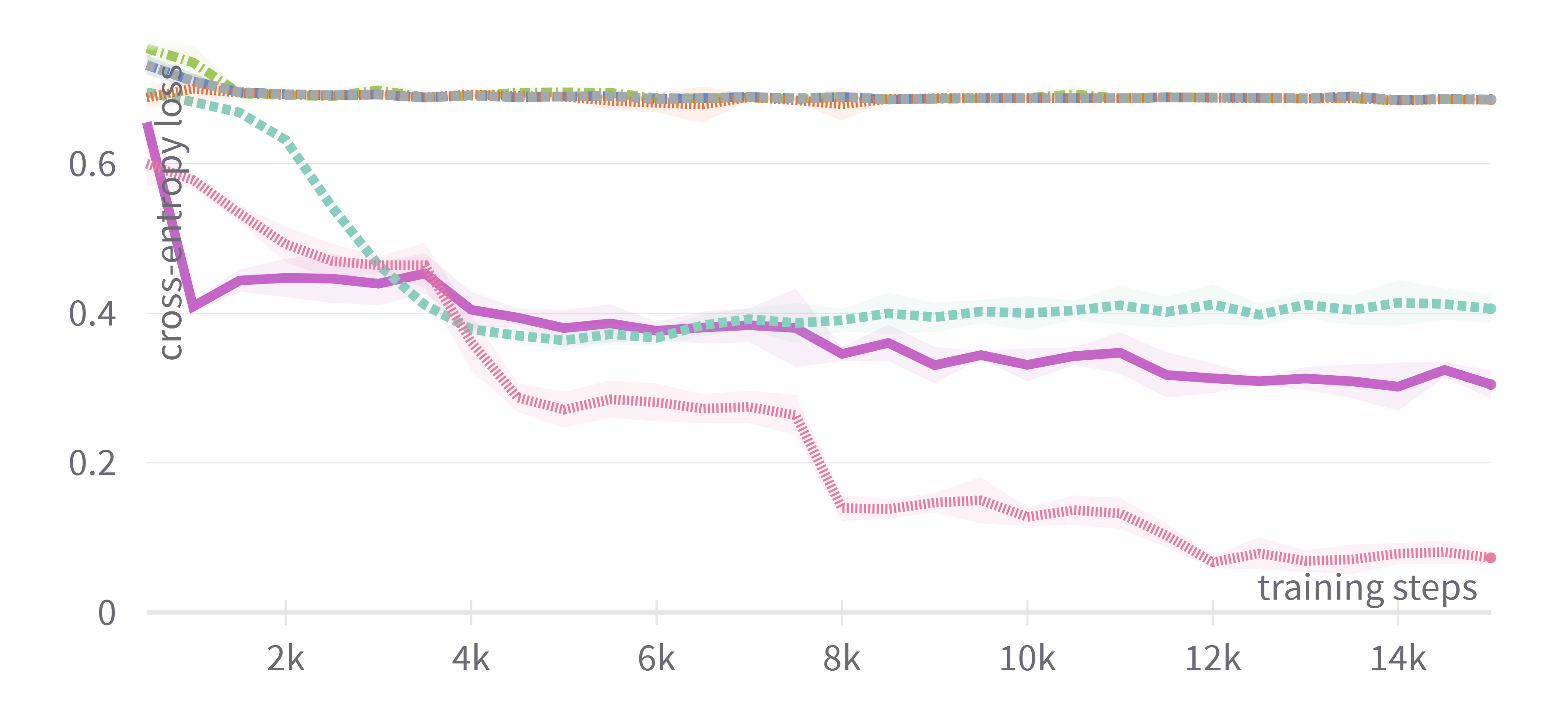}
        \caption{SST-2}
        \label{subfig:sst2lossdef}
        \vspace{0.5mm}
    \end{subfigure}
    \hfill\addtocounter{subfigure}{-1}
    \begin{subfigure}[b]{0.49\textwidth}
        \centering
        \includegraphics[width=\textwidth]{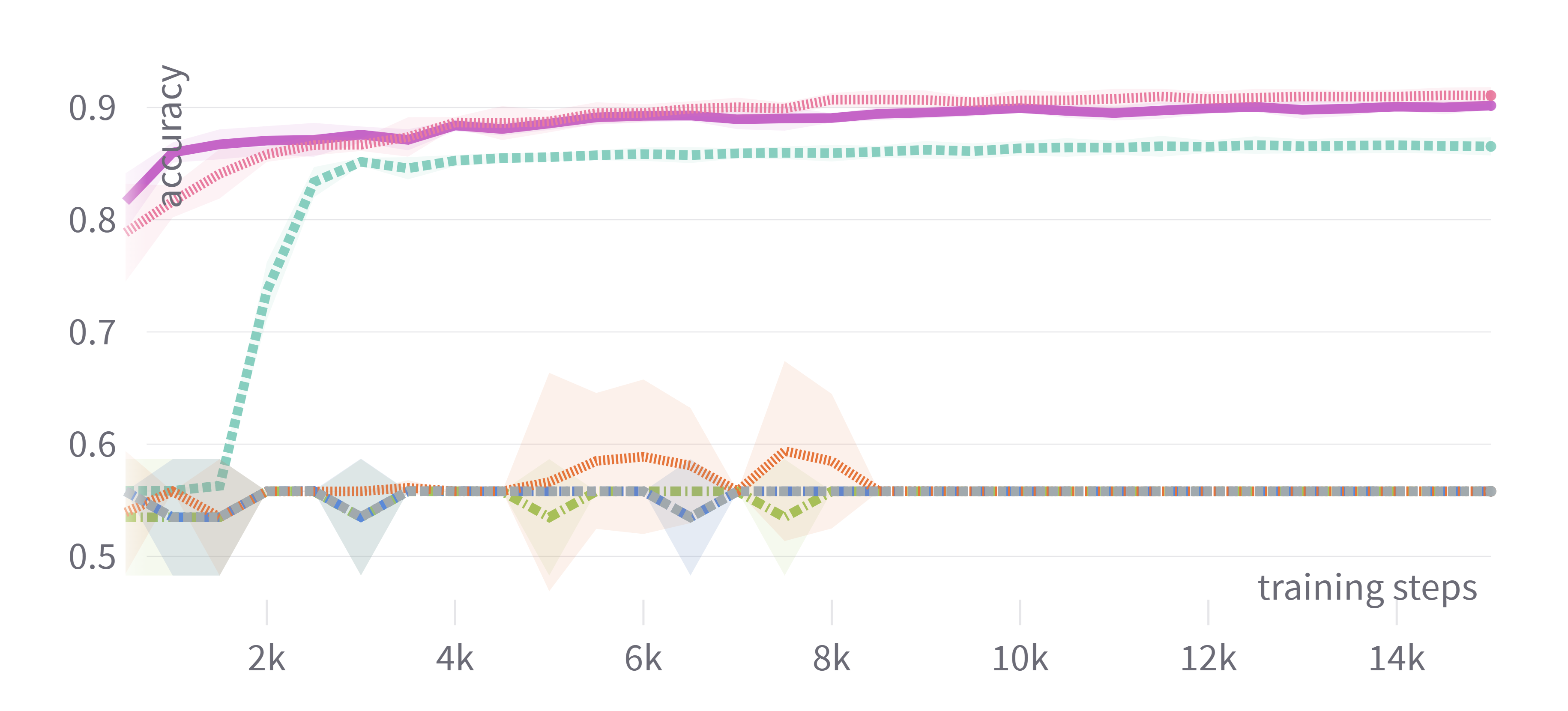}
        \caption{SST-2}
        \label{subfig:sst2accdef}
        \vspace{0.5mm}
    \end{subfigure}
    \hfill
    \begin{subfigure}[b]{0.49\textwidth}
        \centering
        \includegraphics[width=\columnwidth]{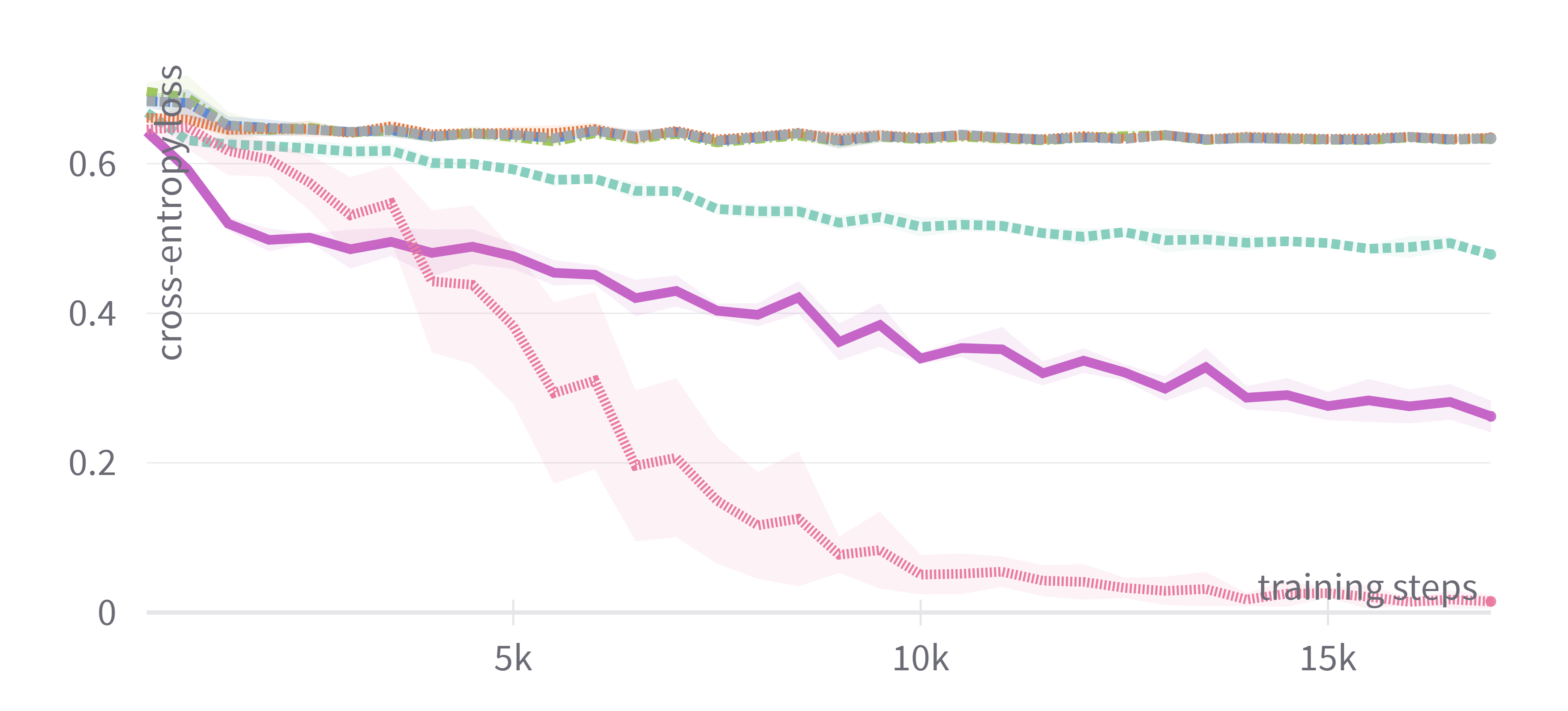}
        \caption{MRPC}
        \label{subfig:mrpclossdef}
        \vspace{0.5mm}
    \end{subfigure}
    \hfill\addtocounter{subfigure}{-1}
    \begin{subfigure}[b]{0.49\textwidth}
        \centering
        \includegraphics[width=\textwidth]{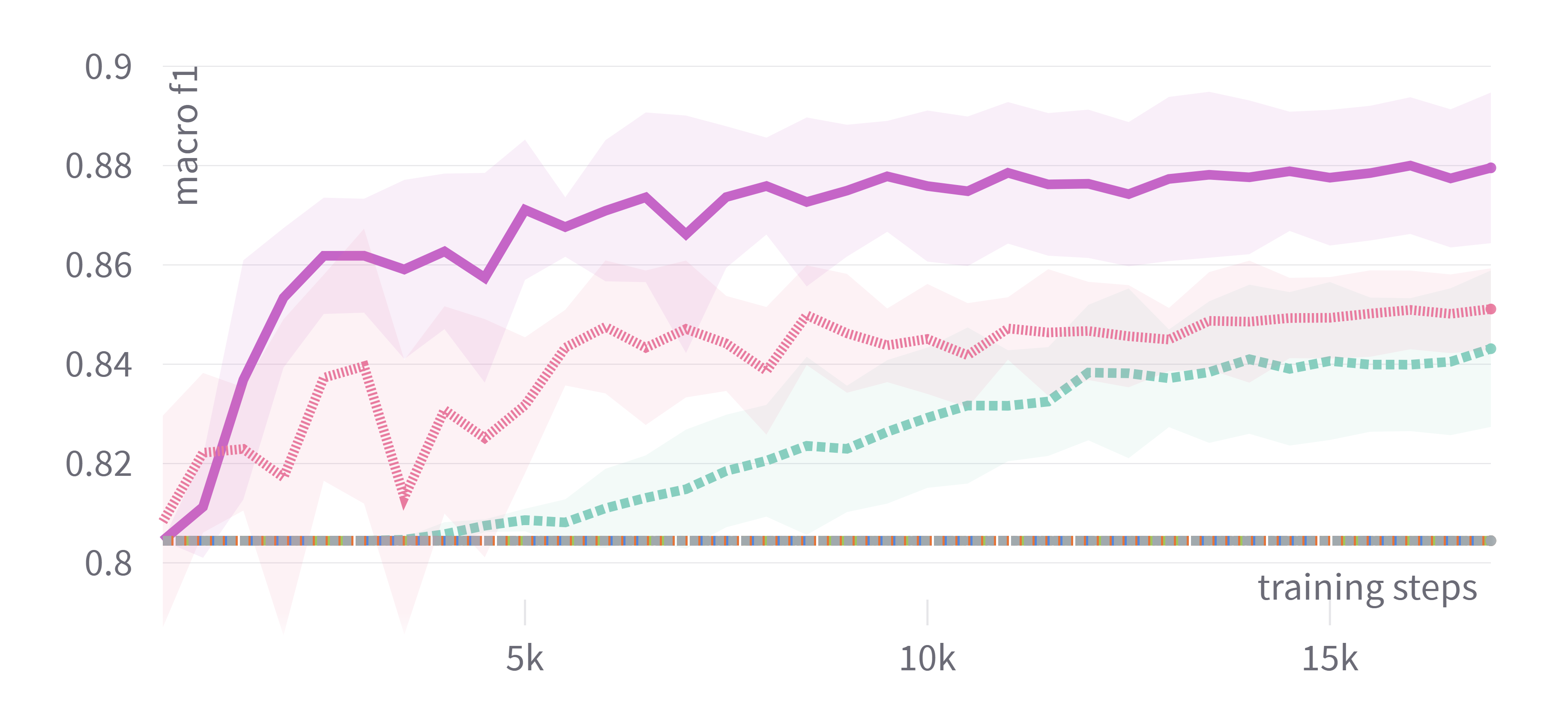}
        \caption{MRPC}
        \label{subfig:mrpcf1def}
        \vspace{0.5mm}
    \end{subfigure}
    \hfill
    \begin{subfigure}[b]{0.49\textwidth}
        \centering
        \includegraphics[width=\textwidth]{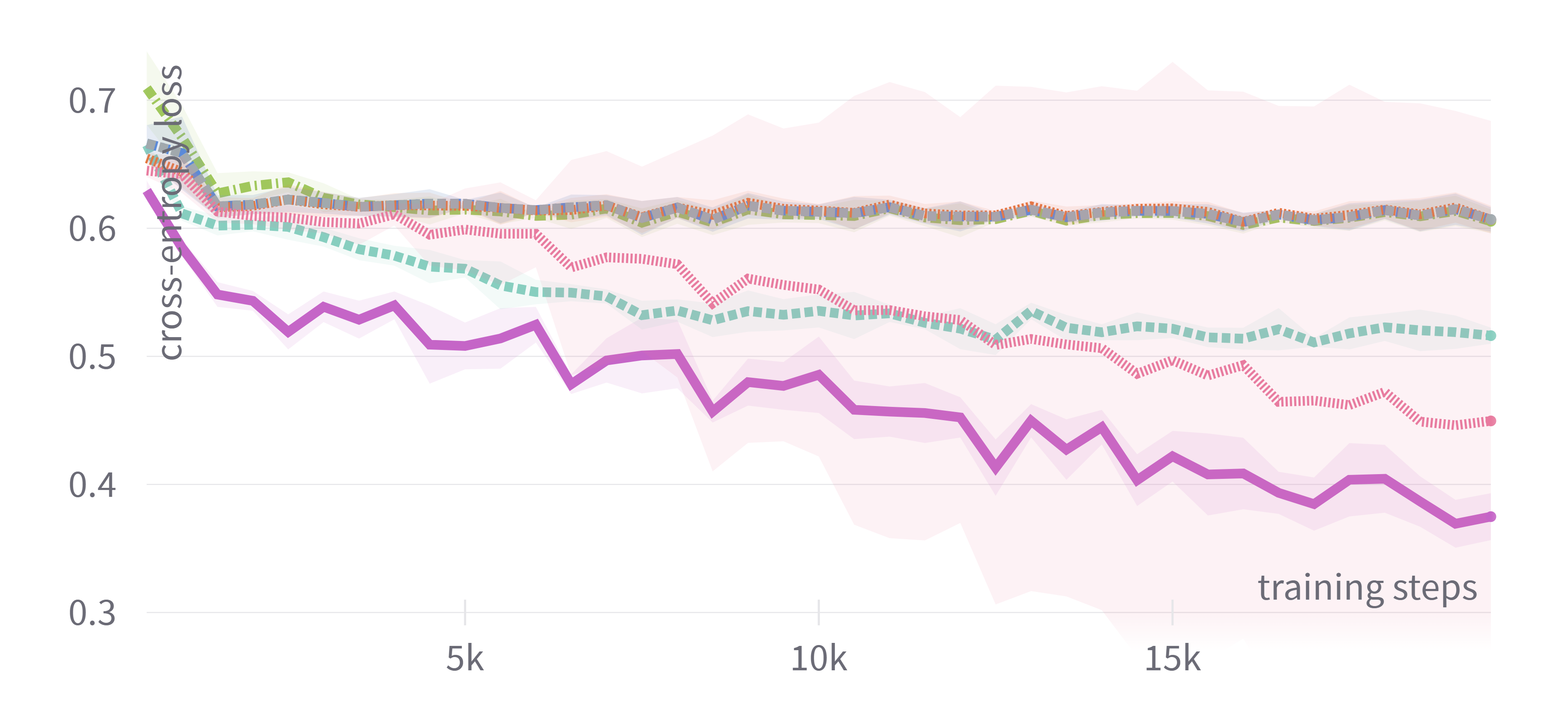}
        \caption{CoLA}
        \label{subfig:colalossdef}
        \vspace{0.5mm}
    \end{subfigure}
    \hfill\addtocounter{subfigure}{-1}
    \begin{subfigure}[b]{0.49\textwidth}
        \centering
        \includegraphics[width=\textwidth]{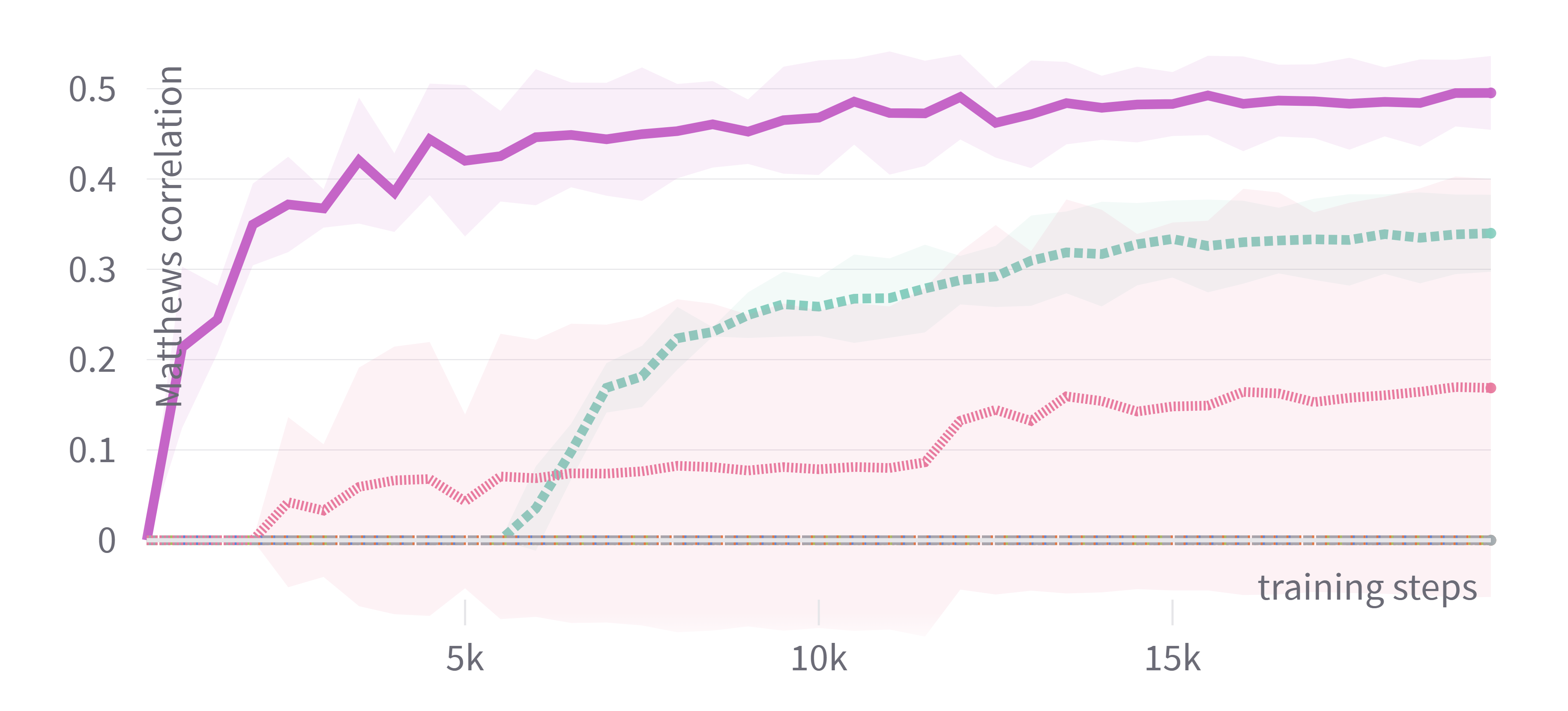}
        \caption{CoLA}
        \label{subfig:colamathdef}
        \vspace{0.5mm}
    \end{subfigure}
    \hfill
    \begin{subfigure}[b]{0.49\textwidth}
        \centering
        \includegraphics[width=\textwidth]{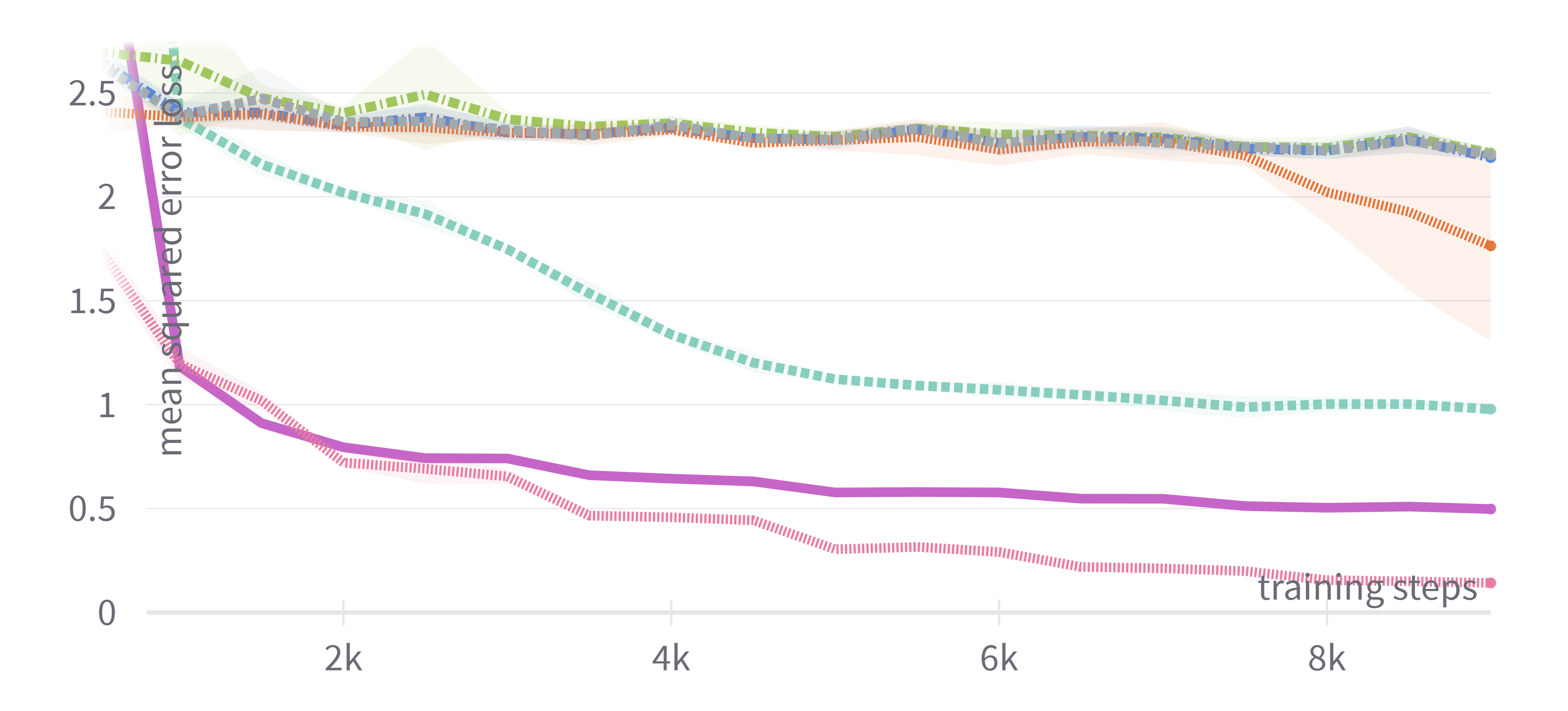}
        \caption{STS-B}
        \label{subfig:stsblossdef}
        \vspace{0.5mm}
    \end{subfigure}
    \hfill\addtocounter{subfigure}{-1}
    \begin{subfigure}[b]{0.49\textwidth}
        \centering
        \includegraphics[width=\textwidth]{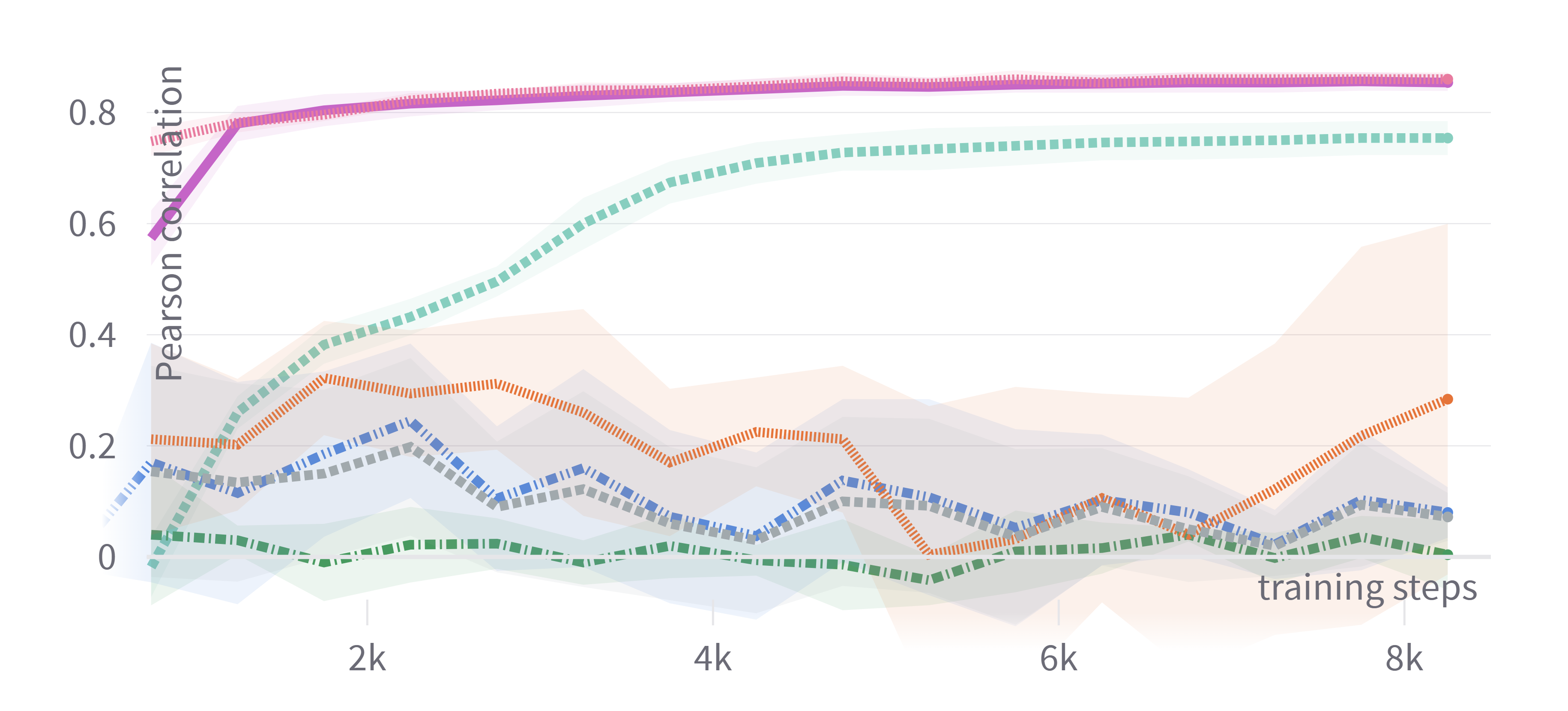}
        \caption{STS-B}
        \label{subfig:stsbpearlossdef}
        \vspace{0.5mm}
    \end{subfigure}
    \hfill
    \begin{subfigure}[b]{0.49\textwidth}
        \centering
        \includegraphics[width=\textwidth]{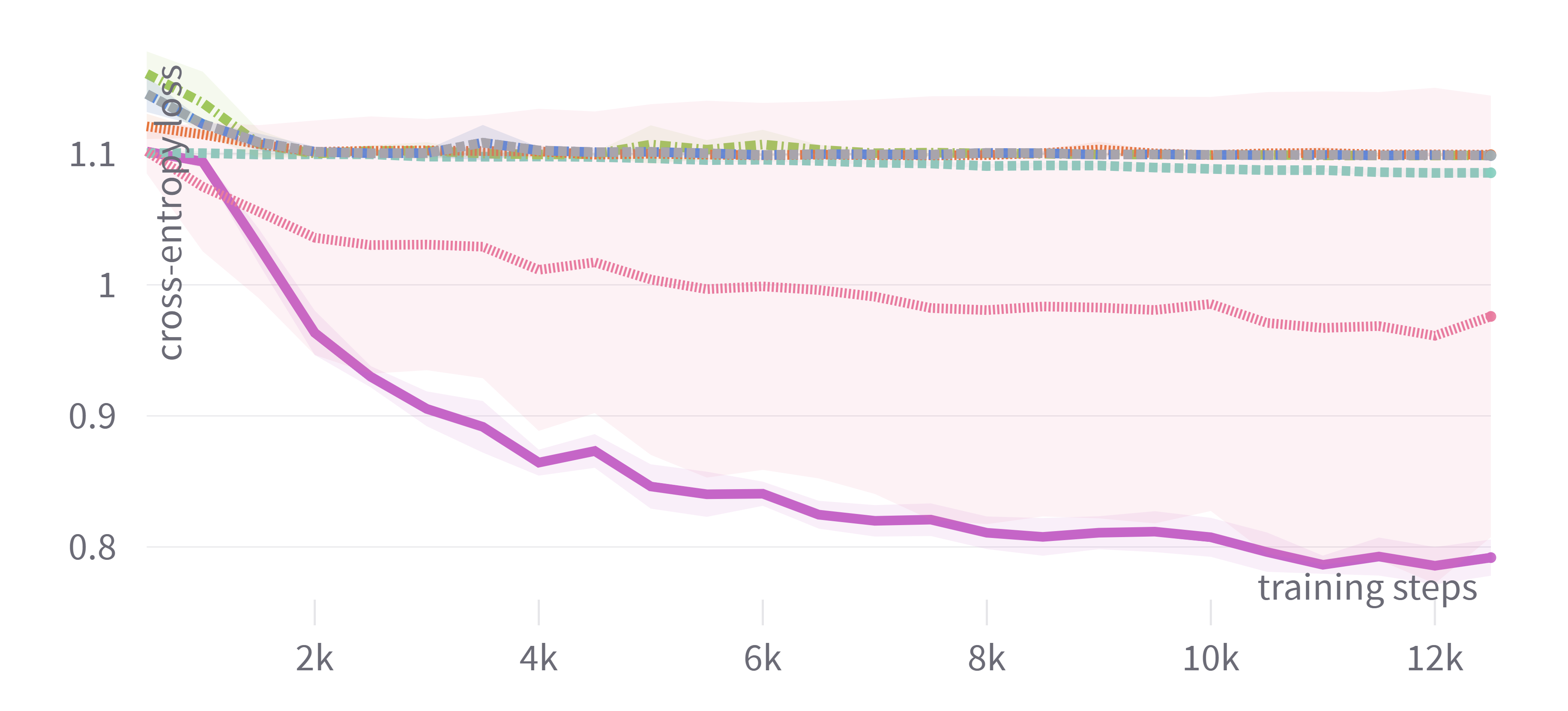}
        \caption{MNLI}
        \label{subfig:mnlilossdef}
    \end{subfigure}
    \hfill\addtocounter{subfigure}{-1}
    \begin{subfigure}[b]{0.49\textwidth}
        \centering
        \includegraphics[width=\textwidth]{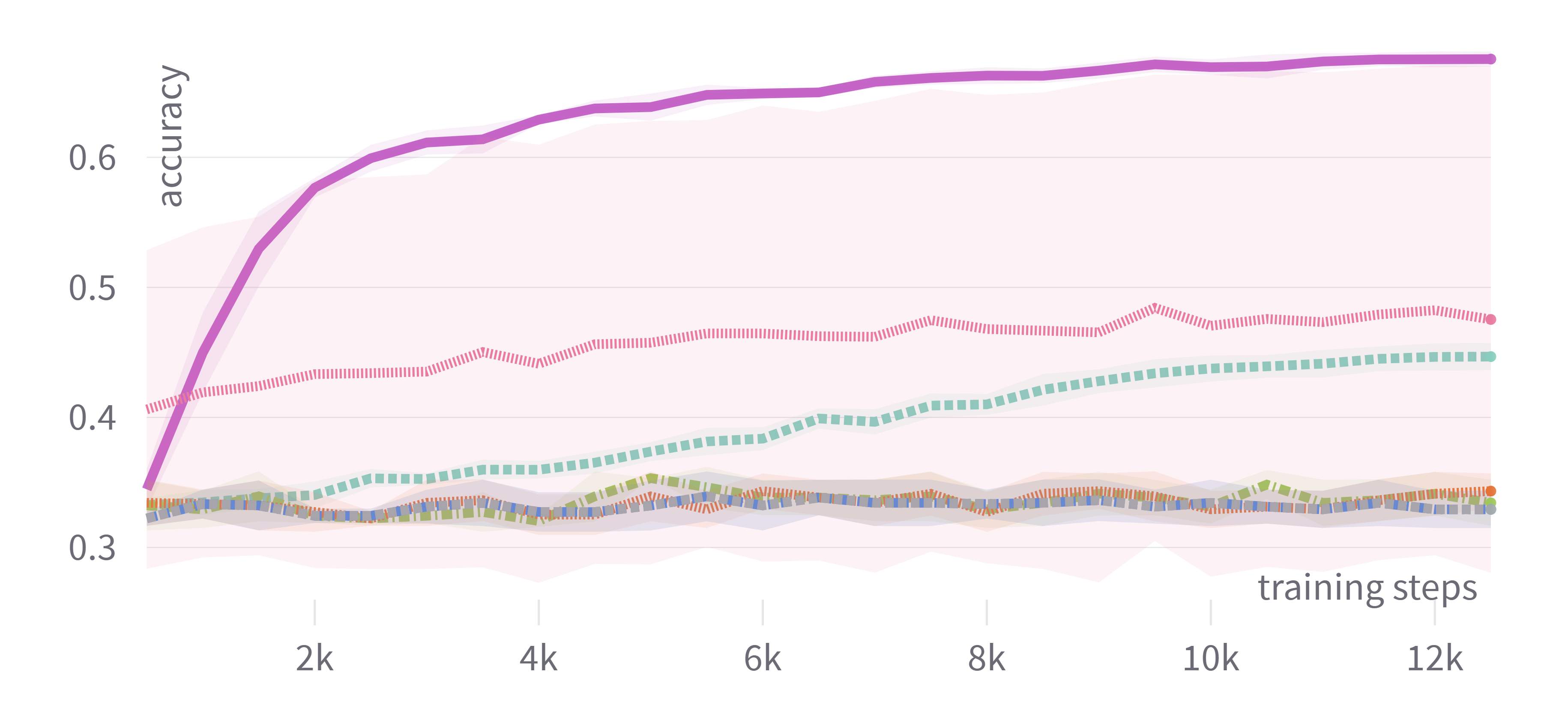}
        \caption{MNLI}
        \label{subfig:mnliaccdef}
    \end{subfigure}

\vspace*{-2mm}
\caption{\textbf{Training loss} (left) and \textbf{evaluation score on development data} (right) with \textbf{all hyperparameters of the optimizers set to their defaults}, as a function of training steps, using \textbf{DistilBERT}. Again, we use \textbf{five random data splits}, and plot the \textbf{average} and \textbf{standard deviation} (shadow). \textbf{SGDM} is not affected by the lack of hyperparameter tuning and is now the \textbf{best overall}, improving upon plain SGD. \textbf{AdaBound} matches SGDM in performance on SST-2 and STS-B, but does not perform as well on the other datasets, although it \textbf{is still better than the other adaptive optimizers}.
Differences in training loss are not always reflected to differences in development scores.}
\label{fig:default_curves}

\end{figure*}

Figure~\ref{fig:default_curves} shows the corresponding curves when all hyperparameters are set to their \emph{defaults}, again using DistilBERT. SGDM is not affected by the lack of tuning and is now the best overall, improving upon plain SGD, which is now overall a strong contender in terms of development scores (in agreement with the test scores of Table~\ref{table:defaulttest}). Although negatively affected by the use of defaults, AdaBound is now overall the second best in terms of development scores (again, as in Table~\ref{table:defaulttest}); it matches SGDM's development scores on SST-2 and STS-B, but does not perform as well on the other datasets, although it is still better than the other adaptive optimizers. Again, differences in training loss are not always reflected to differences in development scores. In SST-2 and MRPC, for example, AdaBound reaches a much lower training loss than SGDM, but the development curve of AdaBound is almost identical (in SST-2) or worse (in MRPC) than the corresponding curve of SGDM. 

\begin{figure*}
\centering

\includegraphics[width=0.7\textwidth]{optimizers.png}

    \begin{subfigure}[b]{0.49\textwidth}
        \centering
        \includegraphics[width=\textwidth]{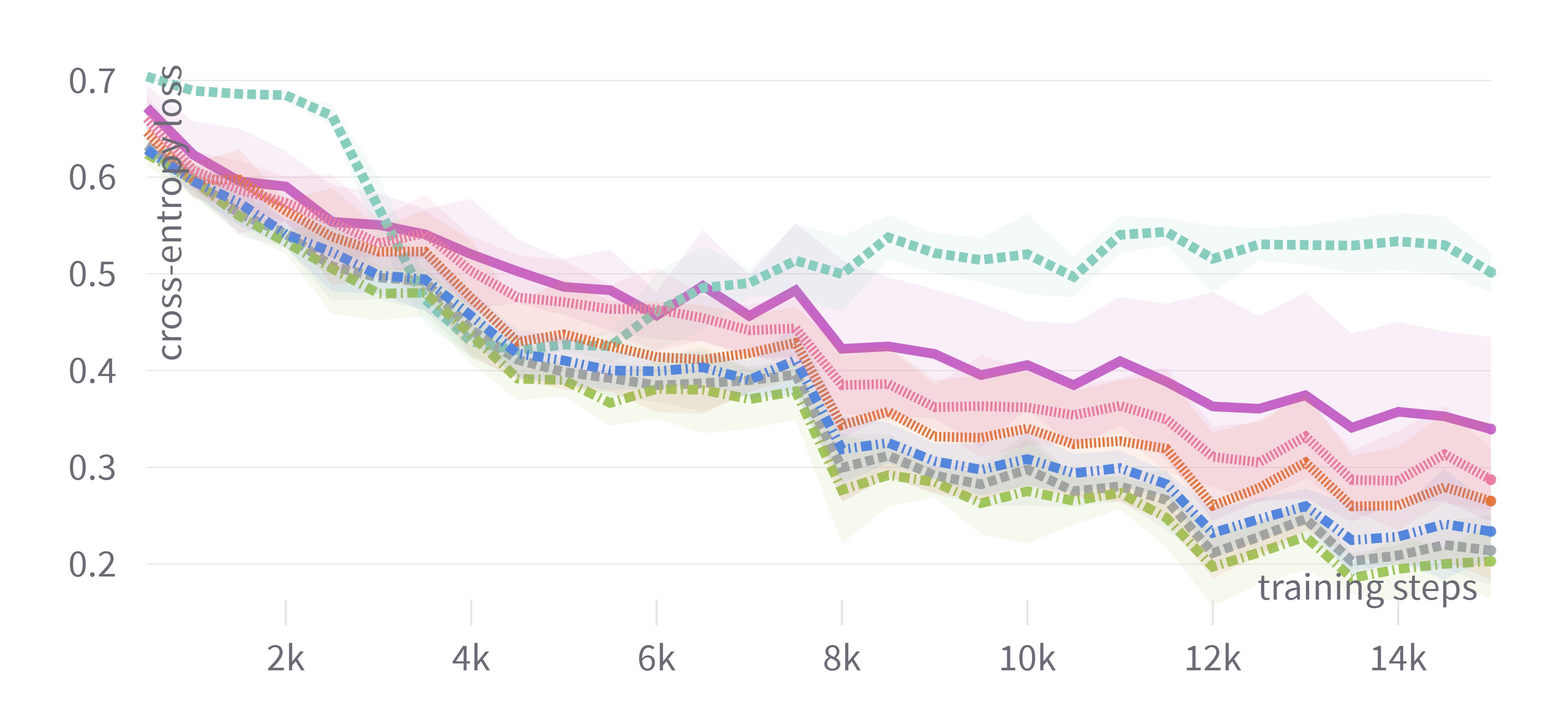}
        \caption{SST-2}
        \label{subfig:distilrobertasst2losstuned}
        \vspace{0.5mm}
    \end{subfigure}
    \hfill\addtocounter{subfigure}{-1}
    \begin{subfigure}[b]{0.49\textwidth}
        \centering
        \includegraphics[width=\textwidth]{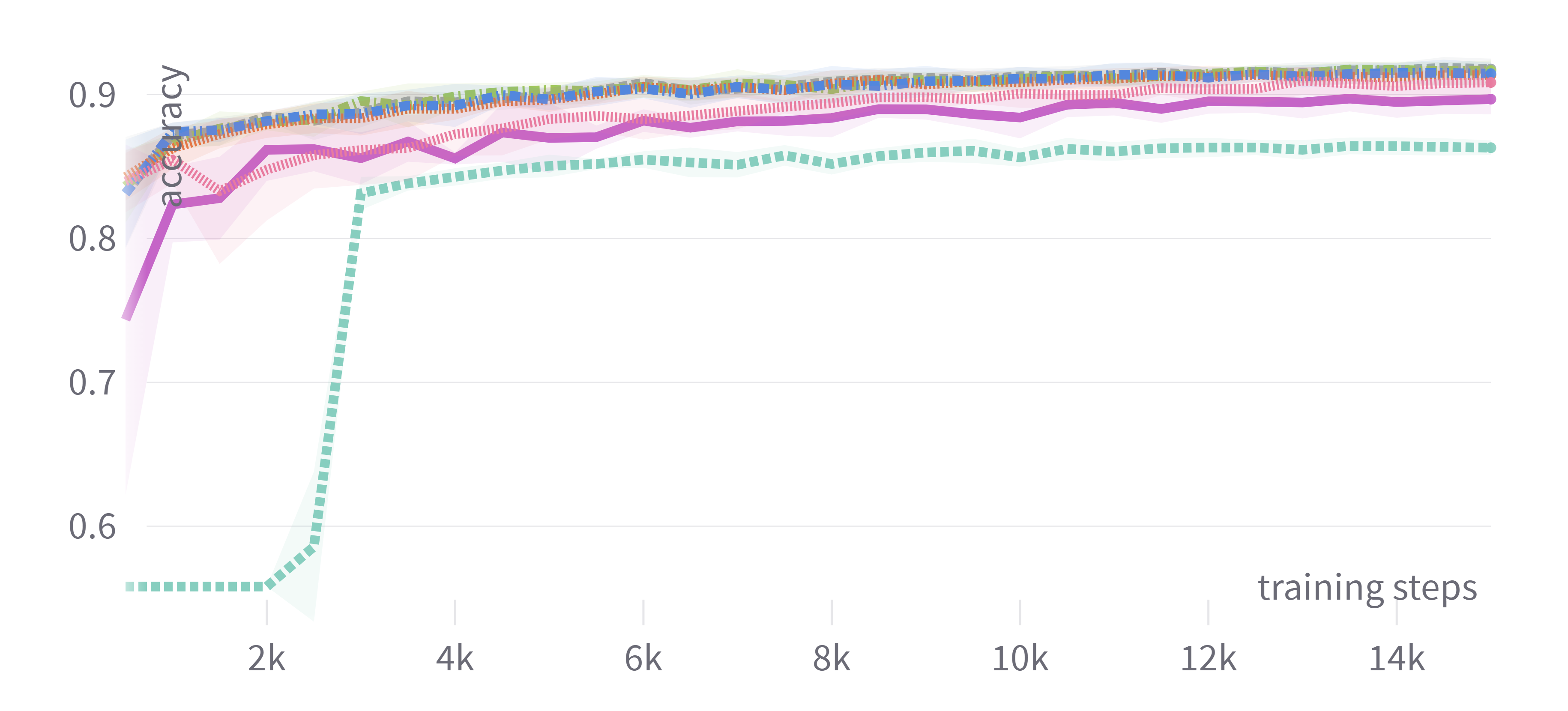}
        \caption{SST-2}
        \label{subfig:distilrobertasst2acctuned}
        \vspace{0.5mm}
    \end{subfigure}
    \hfill
    \begin{subfigure}[b]{0.49\textwidth}
        \centering
        \includegraphics[width=\textwidth]{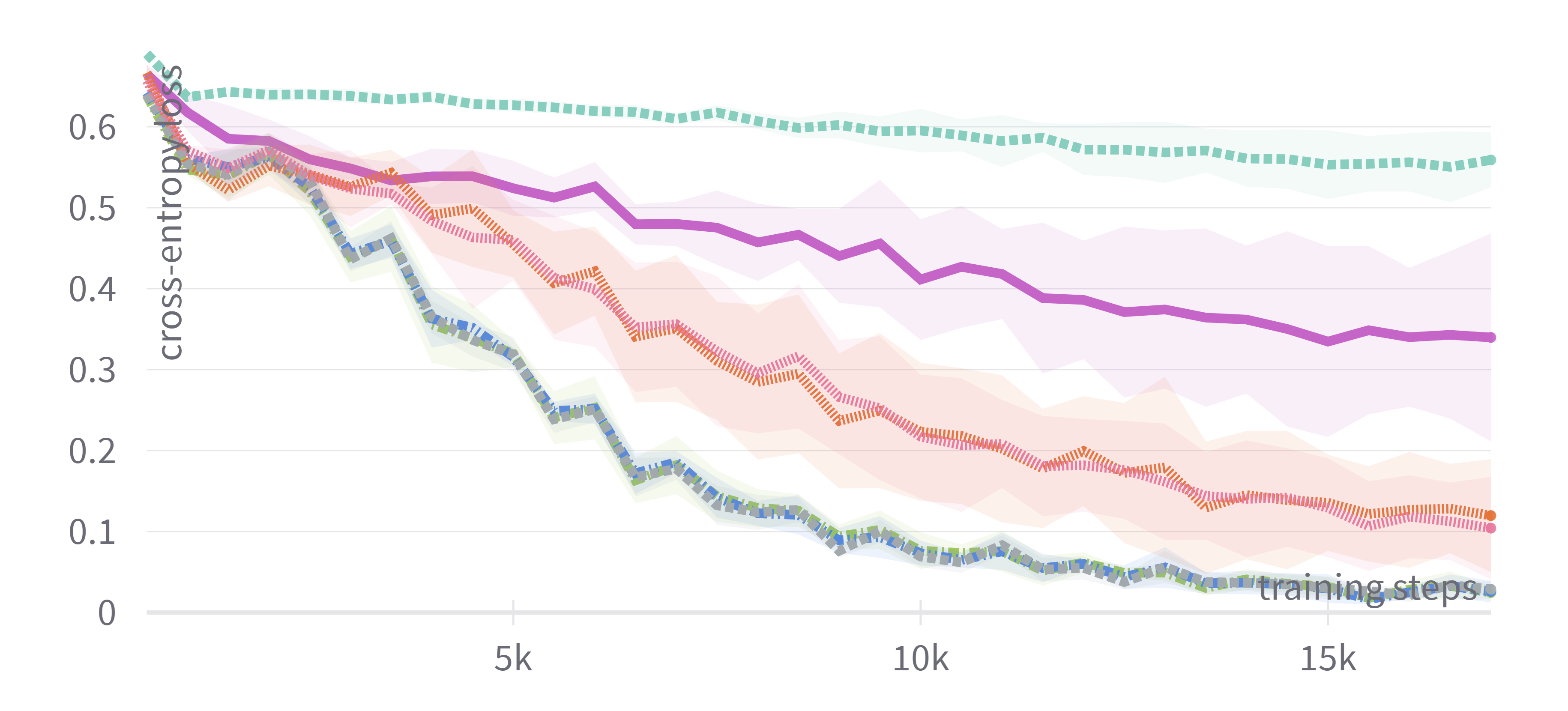}
        \caption{MRPC}
        \label{subfig:distilrobertamrpclosstuned}
        \vspace{0.5mm}
    \end{subfigure}
    \hfill\addtocounter{subfigure}{-1}
    \begin{subfigure}[b]{0.49\textwidth}
        \centering
        \includegraphics[width=\textwidth]{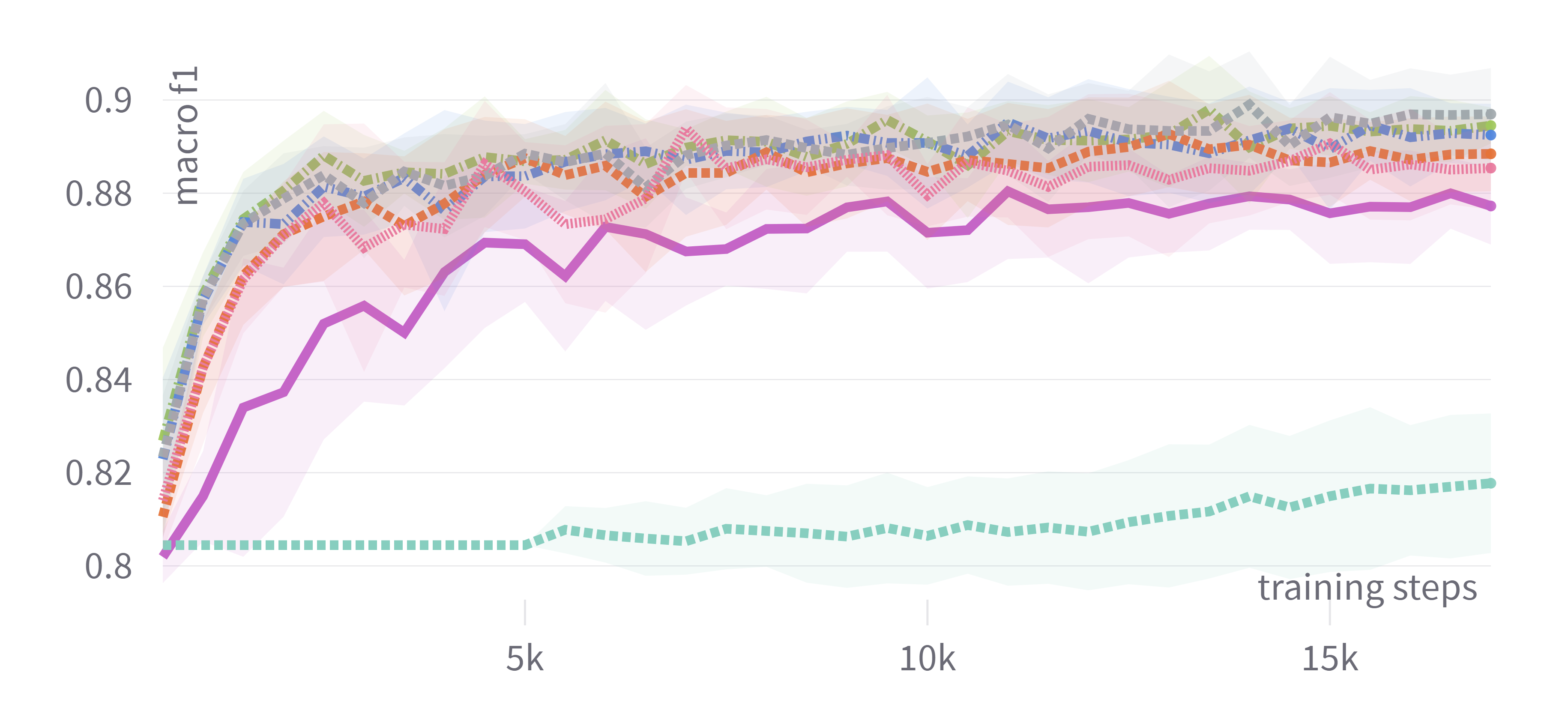}
        \caption{MRPC}
        \label{subfig:distilrobertamrpcf1tuned}
        \vspace{0.5mm}
    \end{subfigure}
    \hfill
    \begin{subfigure}[b]{0.49\textwidth}
        \centering
        \includegraphics[width=\textwidth]{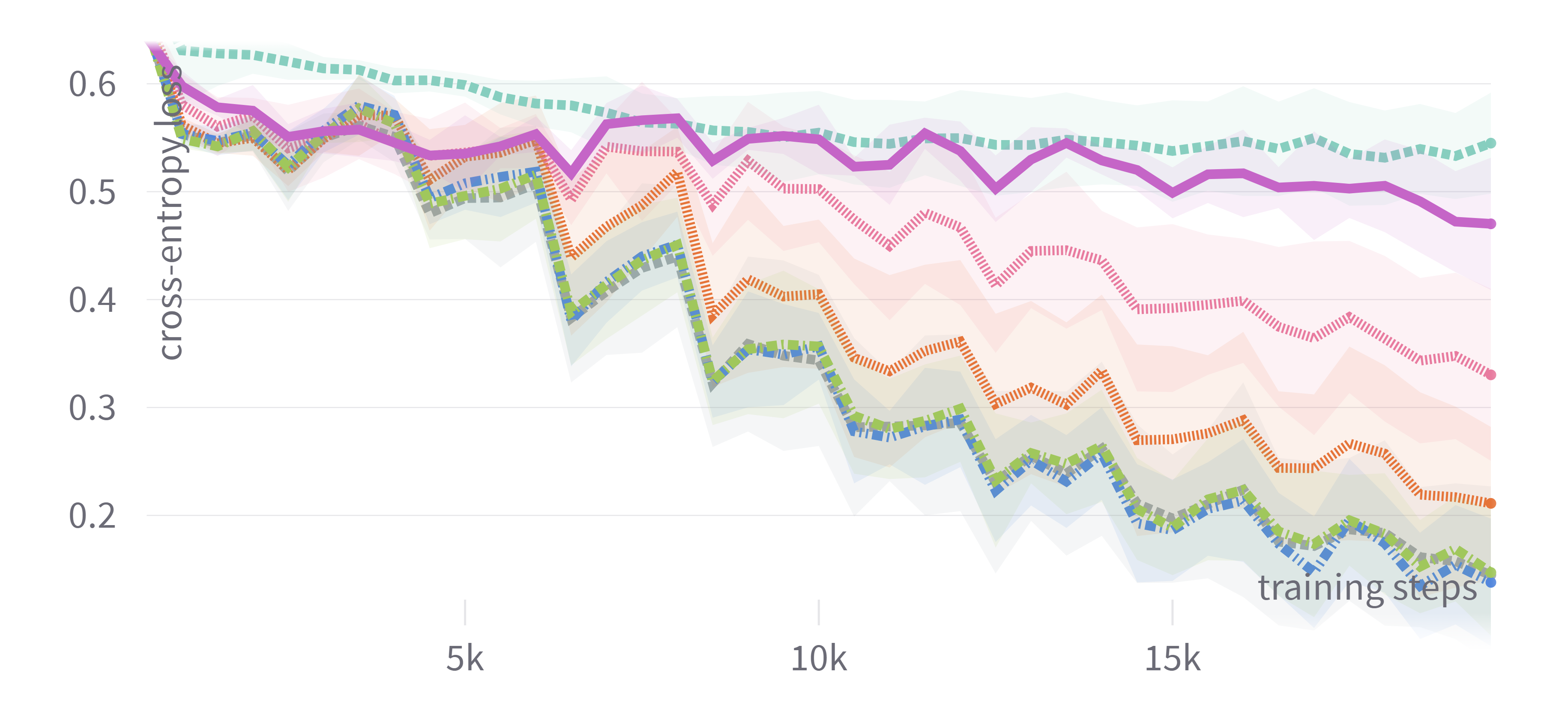}
        \caption{CoLA}
        \label{subfig:distilrobertacolalosstuned}
        \vspace{0.5mm}
    \end{subfigure}
    \hfill\addtocounter{subfigure}{-1}
    \begin{subfigure}[b]{0.49\textwidth}
        \centering
        \includegraphics[width=\textwidth]{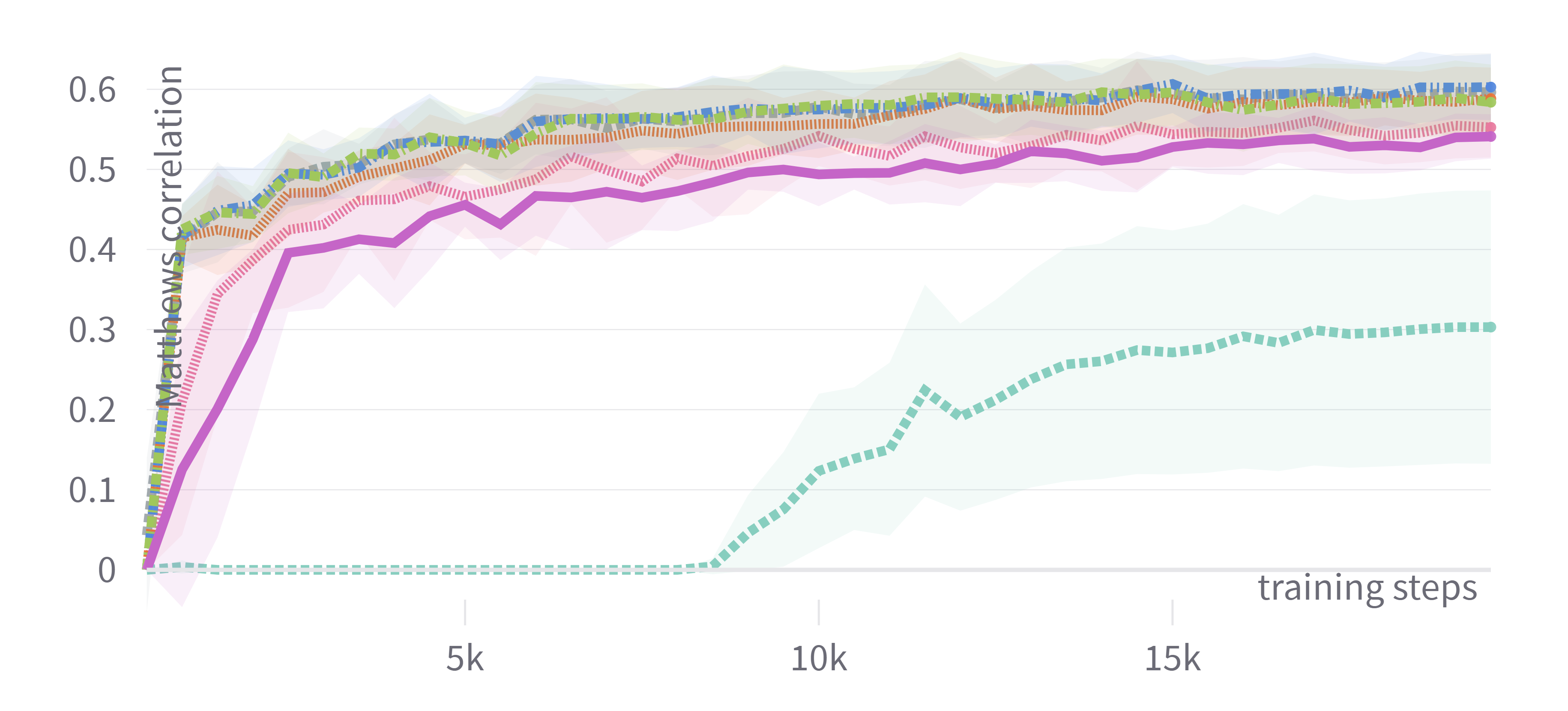}
        \caption{CoLA}
        \label{subfig:distilrobertacolamathtuned}
        \vspace{0.5mm}
    \end{subfigure}
    \hfill
    \begin{subfigure}[b]{0.49\textwidth}
        \centering
        \includegraphics[width=\textwidth]{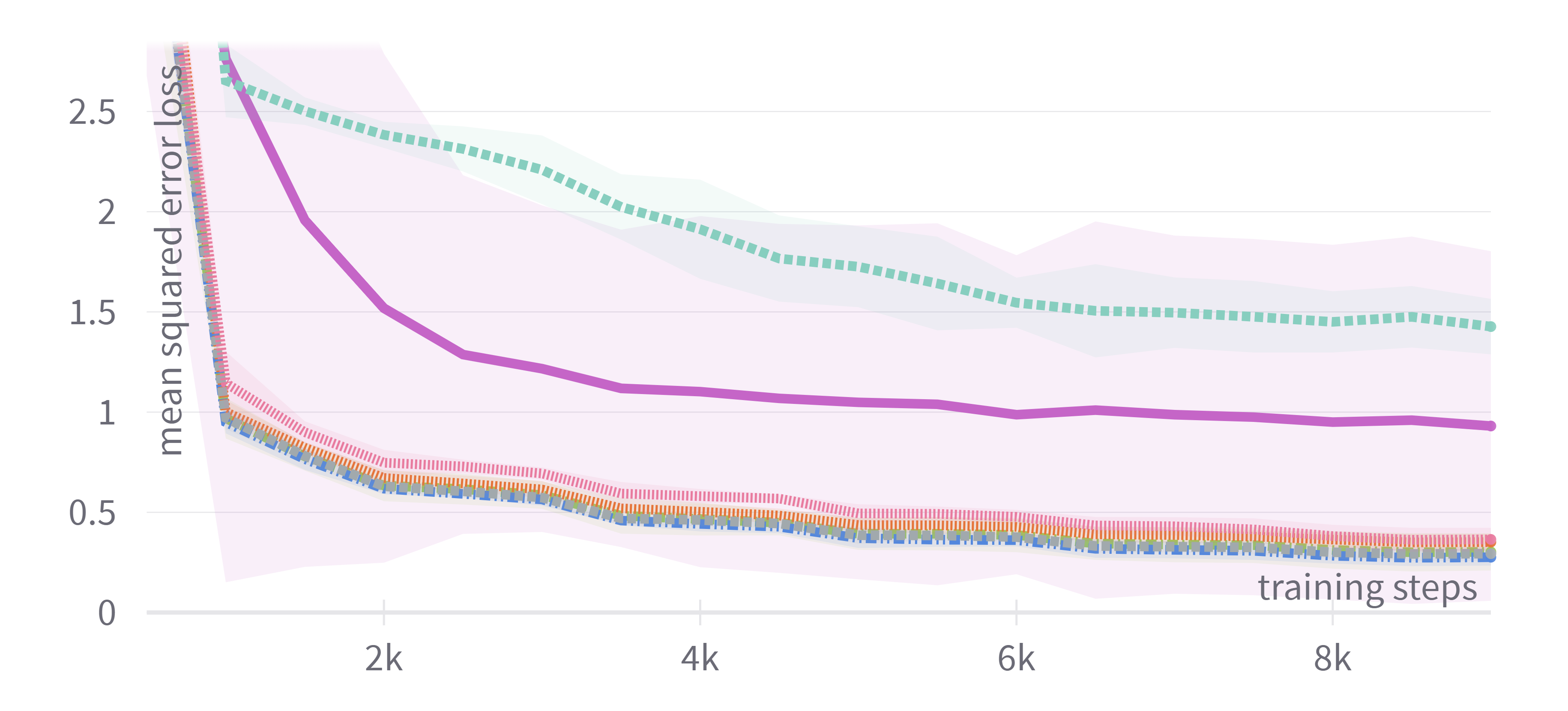}
        \caption{STS-B}
        \label{subfig:distilrobertastsblosstuned}
        \vspace{0.5mm}
    \end{subfigure}
    \hfill\addtocounter{subfigure}{-1}
    \begin{subfigure}[b]{0.49\textwidth}
        \centering
        \includegraphics[width=\textwidth]{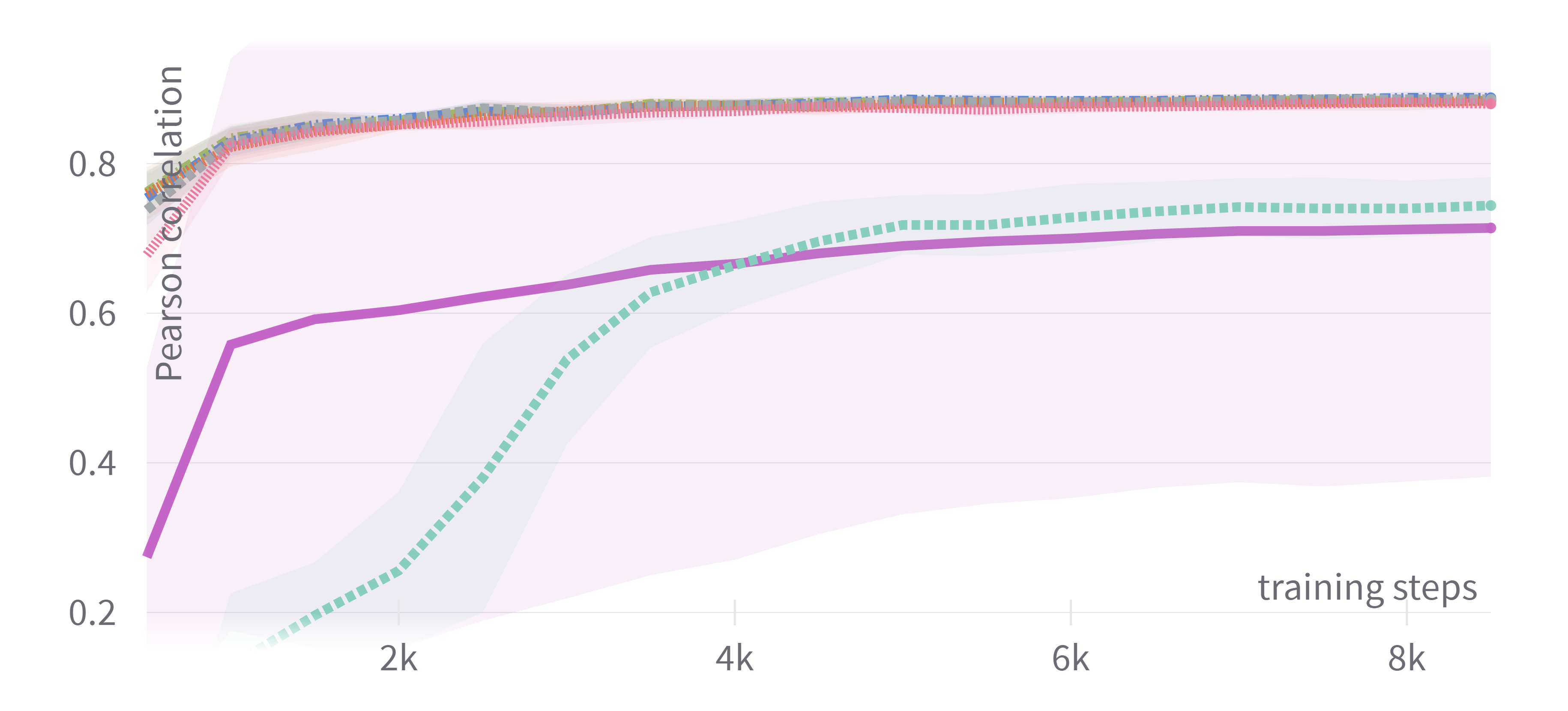}
        \caption{STS-B}
        \label{subfig:distilrobertastsbpeartuned}
        \vspace{0.5mm}
    \end{subfigure}
    \hfill
    \begin{subfigure}[b]{0.49\textwidth}
        \centering
        \includegraphics[width=\textwidth]{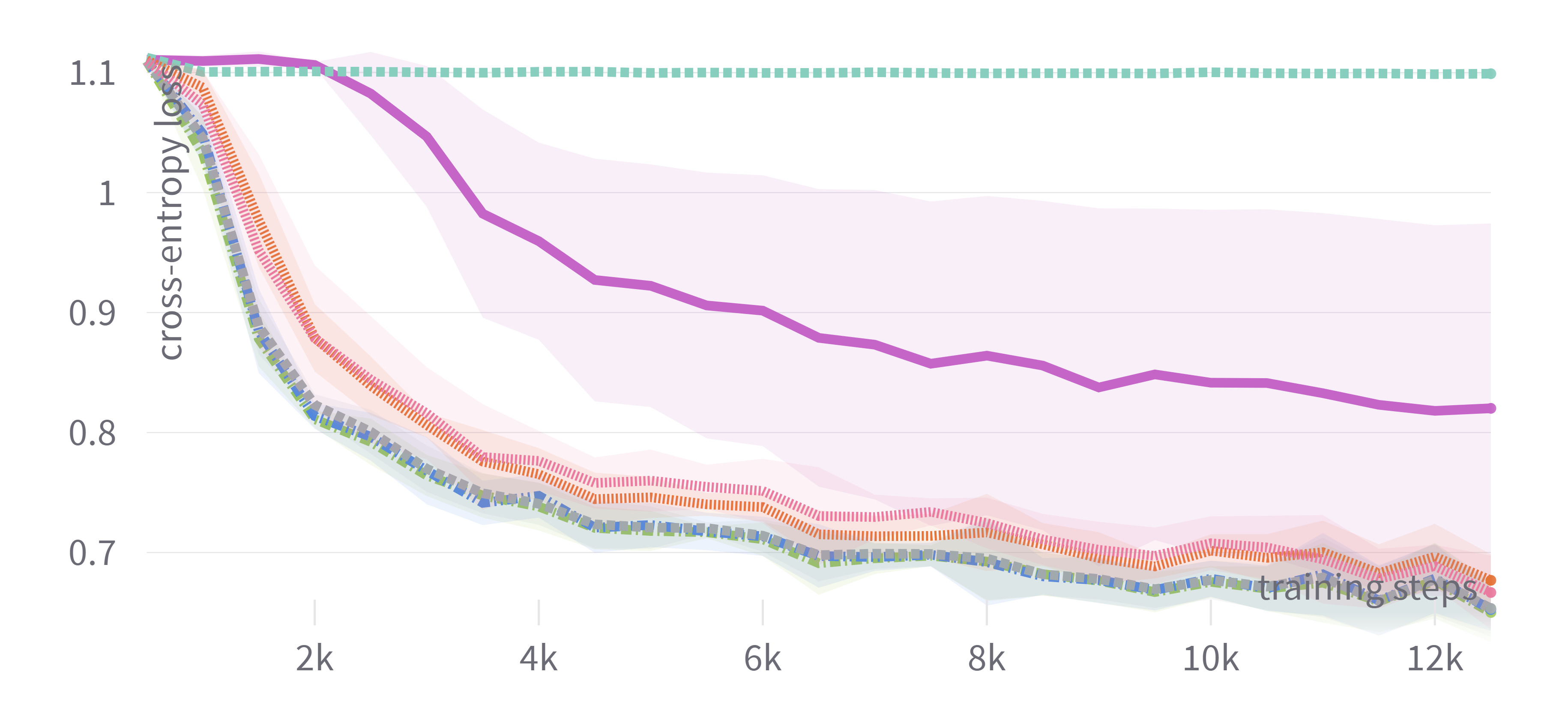}
        \caption{MNLI}
        \label{subfig:distilrobertamnlilosstuned}
    \end{subfigure}
    \hfill\addtocounter{subfigure}{-1}
    \begin{subfigure}[b]{0.49\textwidth}
        \centering
        \includegraphics[width=\textwidth]{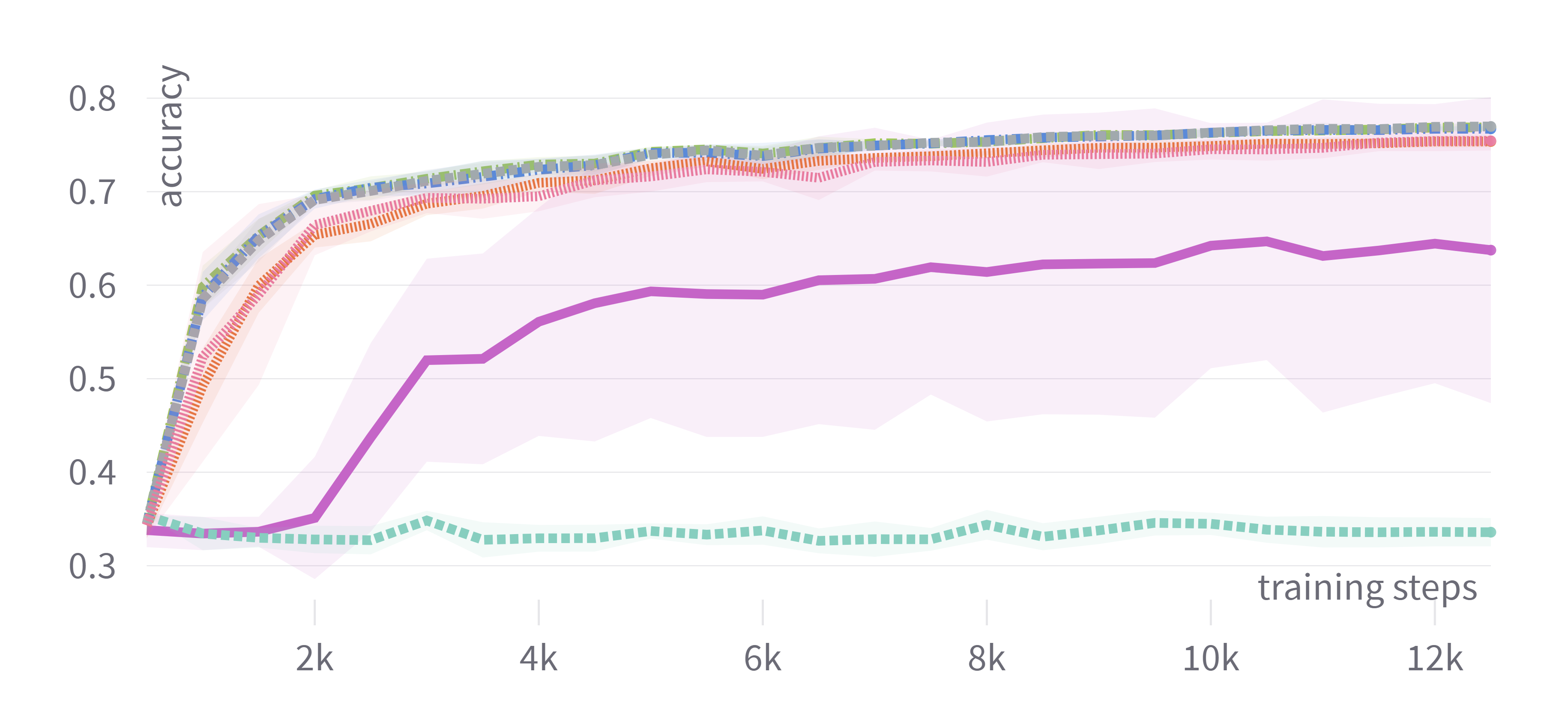}
        \caption{MNLI}
        \label{subfig:distilrobertamnliacctuned}
    \end{subfigure}

\vspace*{-2mm}
\caption{\textbf{Training loss} (left) and \textbf{evaluation score on development data} (right) with \textbf{all hyperparameters of the optimizers tuned}, using \underline{\textbf{DistilRoBERTa}}. For each dataset, we use \textbf{five random data splits}, and plot the \textbf{average} and \textbf{standard deviation} (shadow). The \textbf{results are similar} to those of the the experiments with DistilBERT (cf.\ Fig.~\ref{fig:tuned_curves}), except that SGDM now performs better in development score on CoLA (where it lagged behind the other adaptive optimizers in Fig.~\ref{fig:tuned_curves}) and it now performs poorly on STS-B and MNLI (where it was competent). Hence, these experiments confirm that \textbf{SGDM is overall clearly better than SGD}, but still \textbf{worse than the adaptive optimizers}, when all hyperparameters are tuned. Again, \textbf{the five adaptive optimizers perform very similarly}.
}
\label{fig:distilroberta_tuned_curves}

\end{figure*}

Figure~\ref{fig:distilroberta_tuned_curves} shows the training losses and development scores when \emph{all hyperparameters} are tuned, as in Fig.~\ref{fig:tuned_curves}, but now using DistilRoBERTa instead of DistilBERT. The results are similar to those of Fig.~\ref{fig:tuned_curves}, except that SGDM now performs better in terms of development scores on CoLA (where it lagged behind the other adaptive optimizers in Fig.~\ref{fig:tuned_curves}) and it now performs poorly on STS-B and MNLI (where it was competent). Hence, these experiments confirm that SGDM is overall clearly better than SGD, but still worse than the adaptive optimizers, when all hyperparameters are tuned. Again, the five adaptive optimizers have very similar development scores, despite occasional larger differences in the training losses they reach. 

\begin{figure*}
\centering

\includegraphics[width=0.7\textwidth]{optimizers.png}

    \begin{subfigure}[b]{0.49\textwidth}
        \centering
        \includegraphics[width=\textwidth]{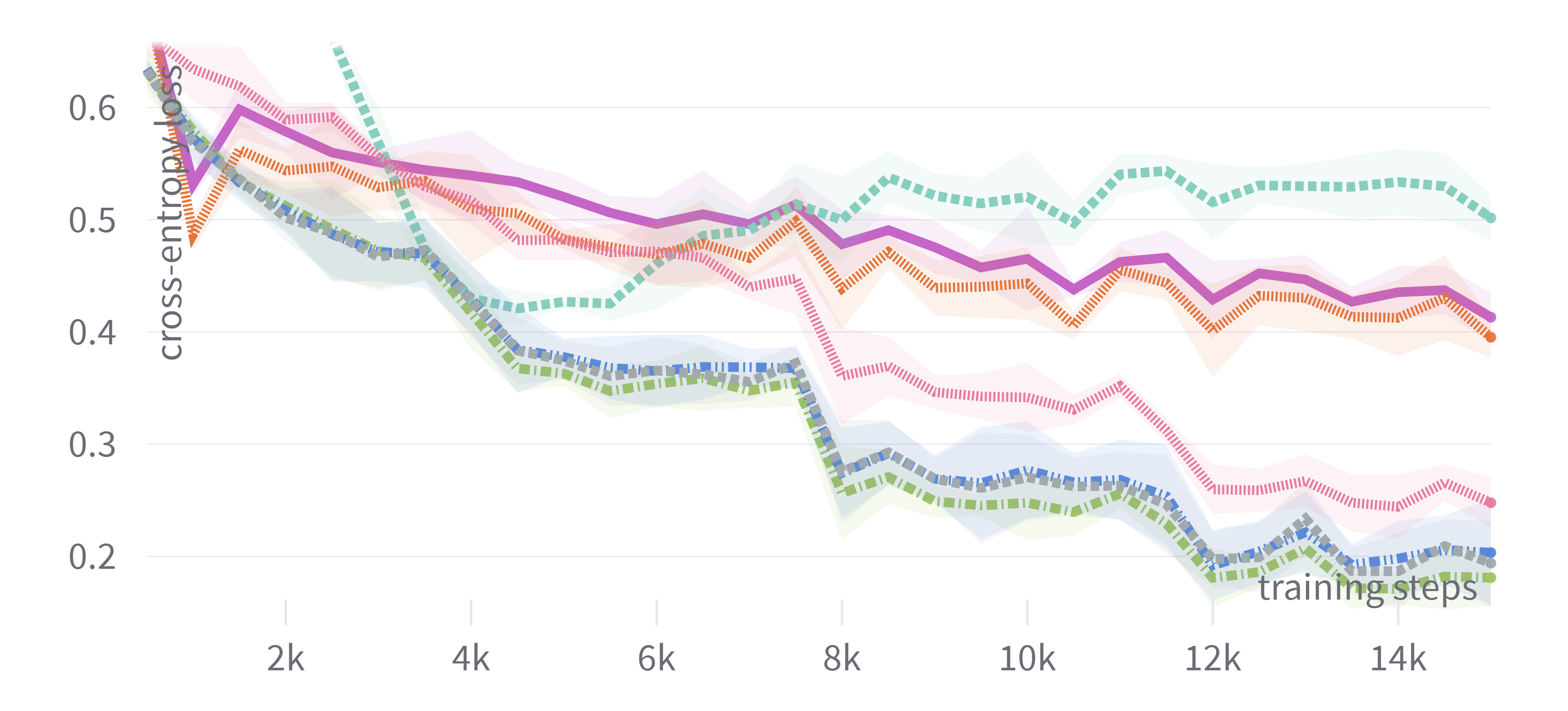}
        \caption{SST-2}
        \label{subfig:distilrobertasst2losslrtuned}
        \vspace{0.5mm}
    \end{subfigure}
    \hfill\addtocounter{subfigure}{-1}
    \begin{subfigure}[b]{0.49\textwidth}
        \centering
        \includegraphics[width=\textwidth]{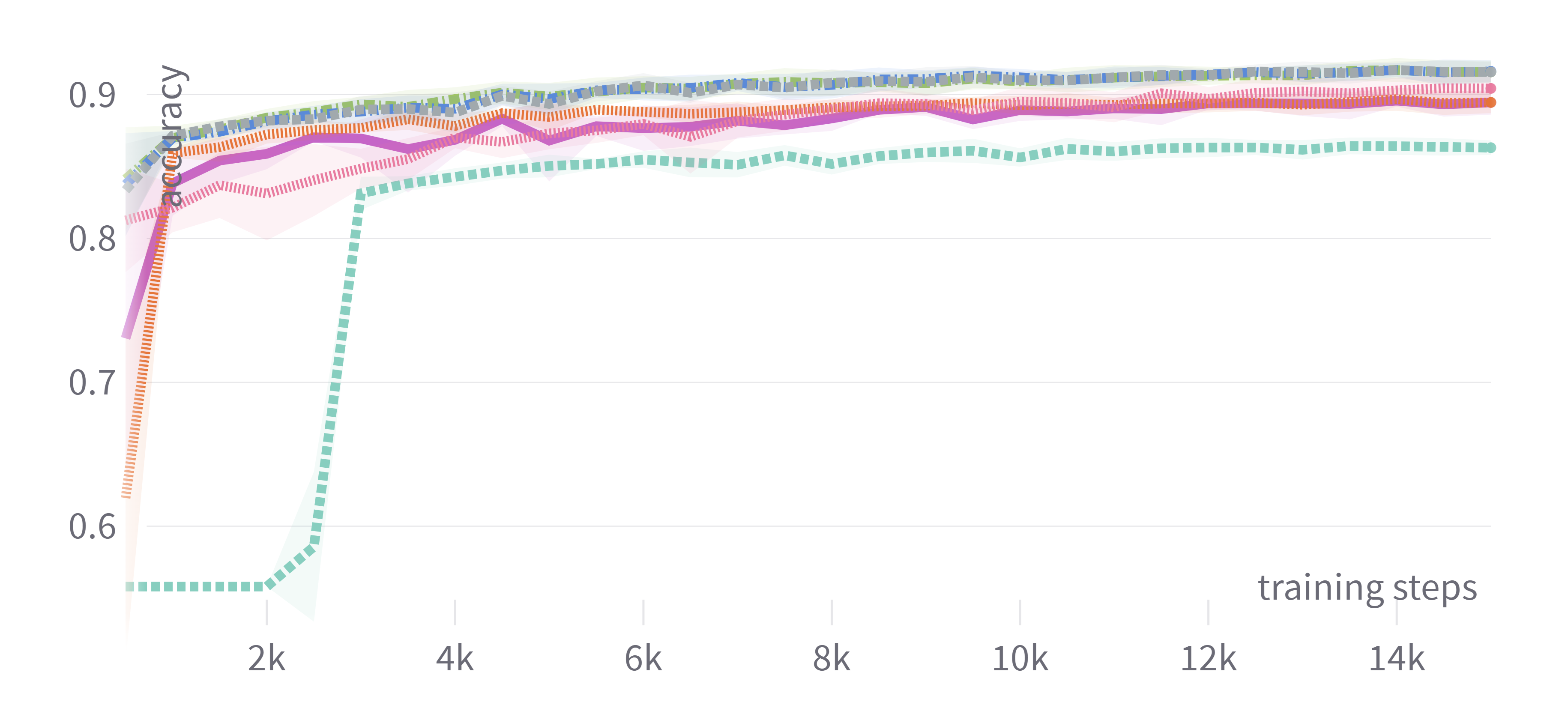}
        \caption{SST-2}
        \label{subfig:distilrobertasst2lacclrtuned}
        \vspace{0.5mm}
    \end{subfigure}
    \hfill
    \begin{subfigure}[b]{0.49\textwidth}
        \centering
        \includegraphics[width=\columnwidth]{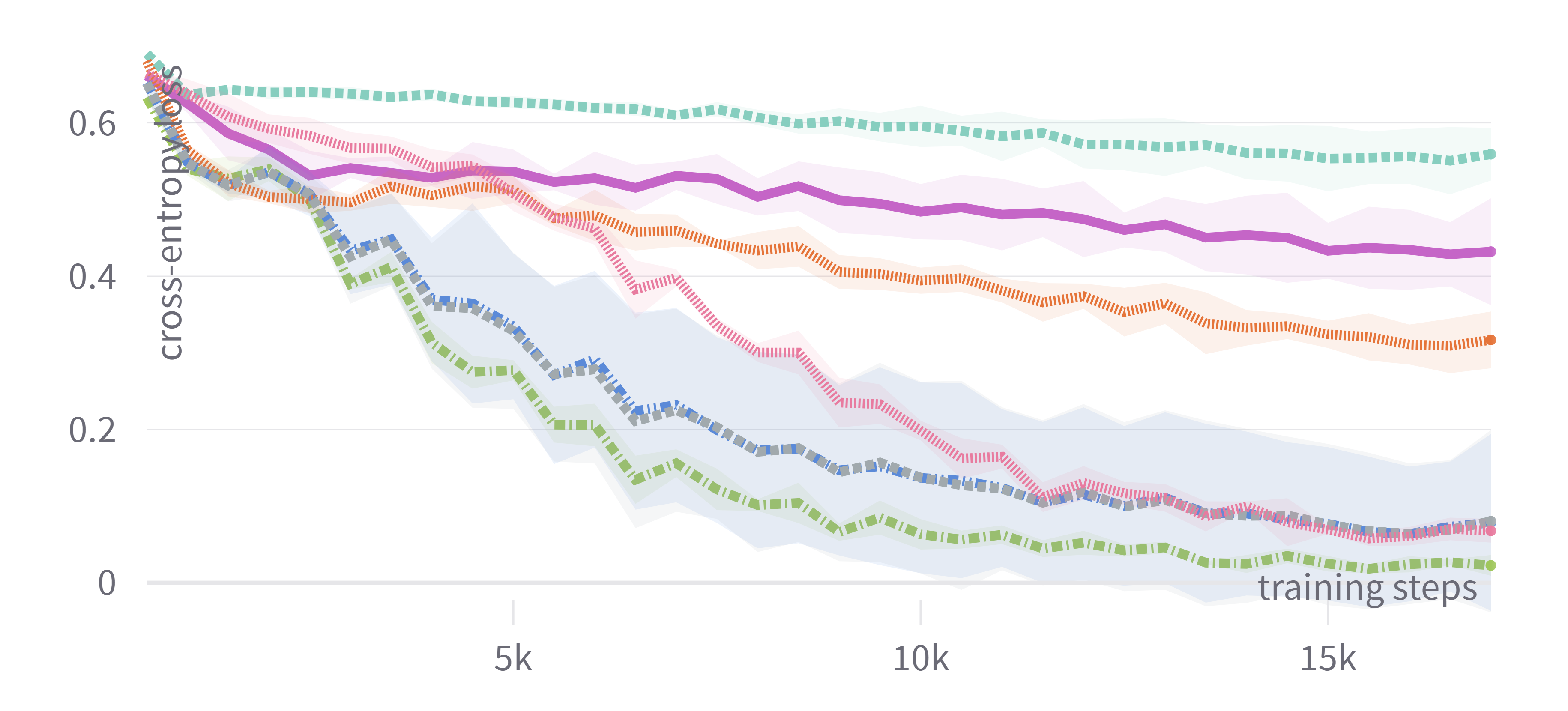}
        \caption{MRPC}
        \label{subfig:distilrobertamrpclosslrtuned}
        \vspace{0.5mm}
    \end{subfigure}
    \hfill\addtocounter{subfigure}{-1}
    \begin{subfigure}[b]{0.49\textwidth}
        \centering
        \includegraphics[width=\textwidth]{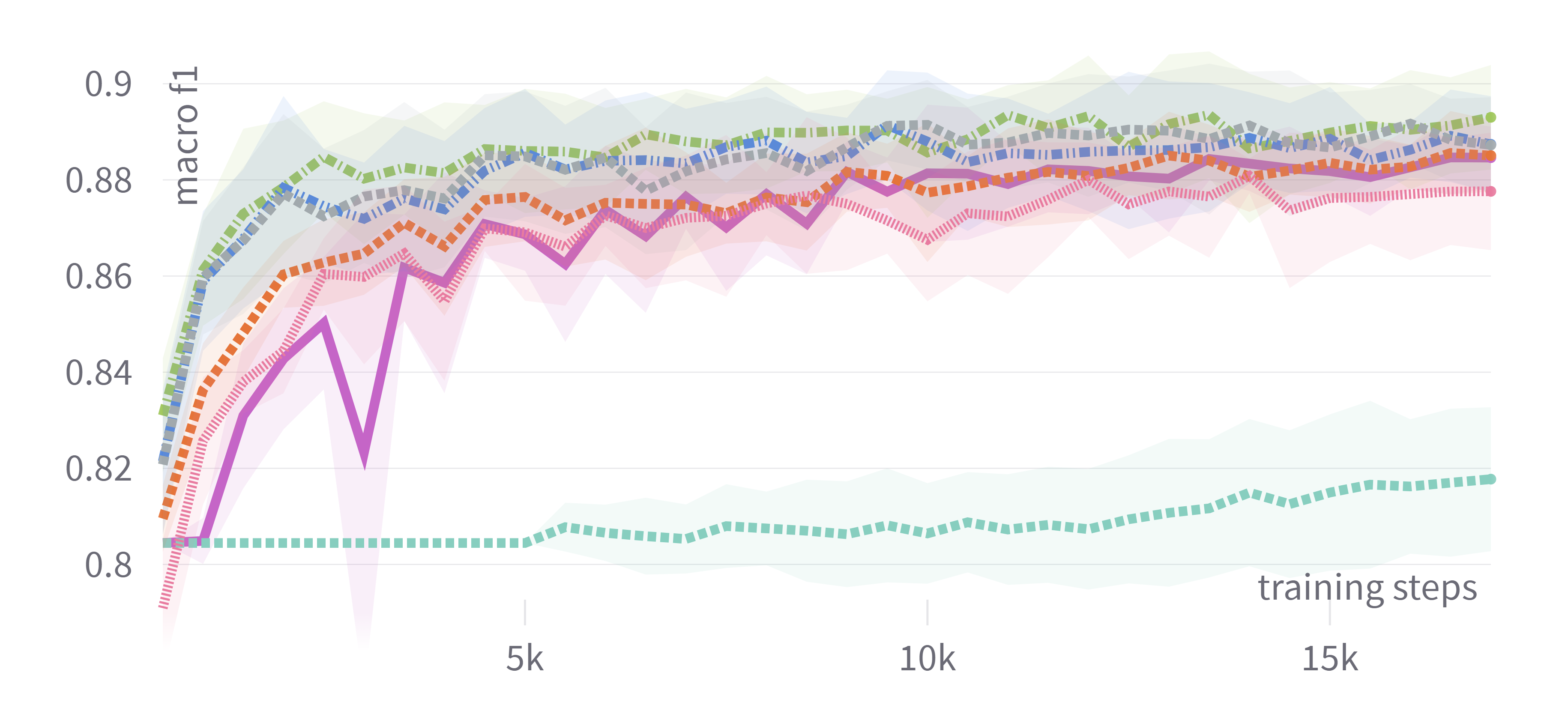}
        \caption{MRPC}
        \label{subfig:distilrobertamrpcf1lrtuned}
        \vspace{0.5mm}
    \end{subfigure}
    \hfill
    \begin{subfigure}[b]{0.49\textwidth}
        \centering
        \includegraphics[width=\textwidth]{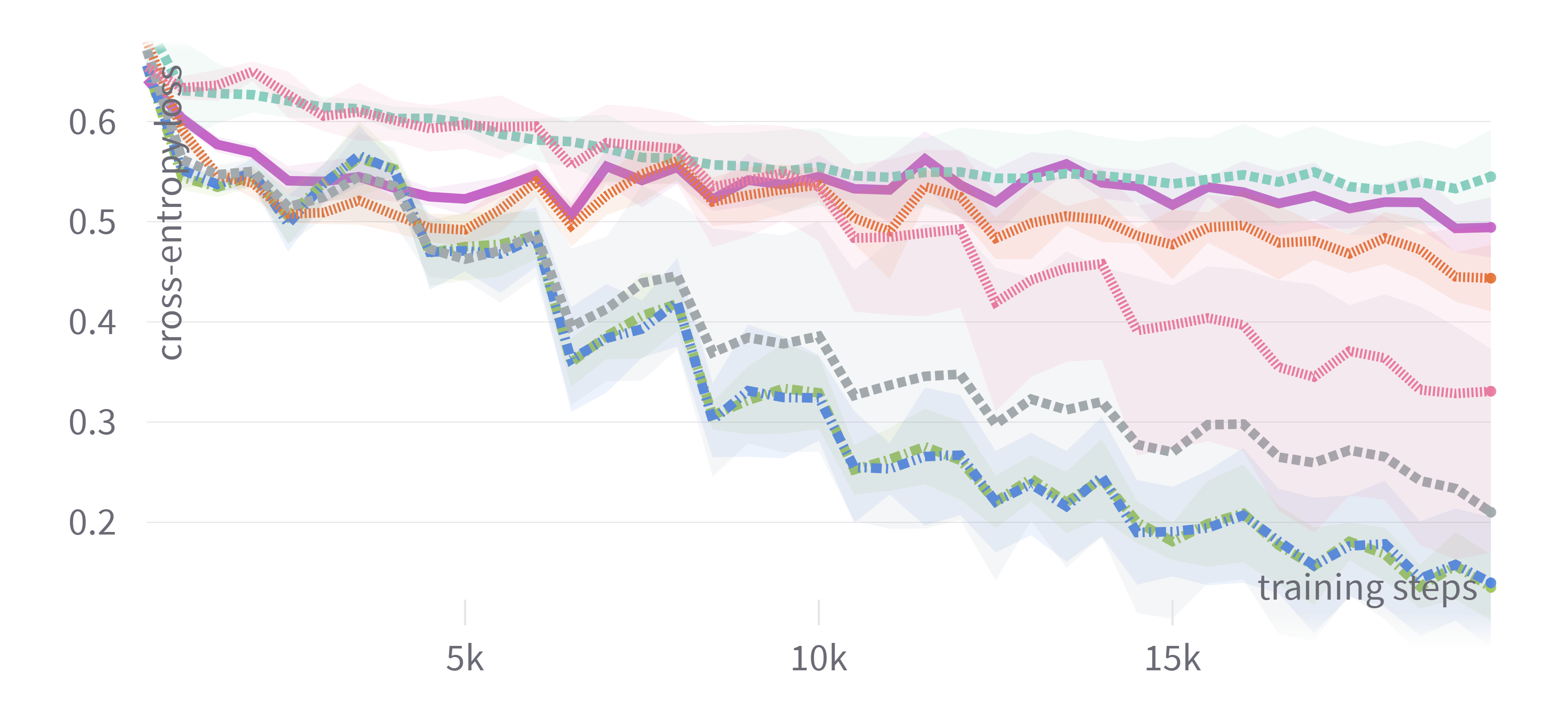}
        \caption{CoLA}
        \label{subfig:distilrobertacolalosslrtuned}
        \vspace{0.5mm}
    \end{subfigure}
    \hfill\addtocounter{subfigure}{-1}
    \begin{subfigure}[b]{0.49\textwidth}
        \centering
        \includegraphics[width=\textwidth]{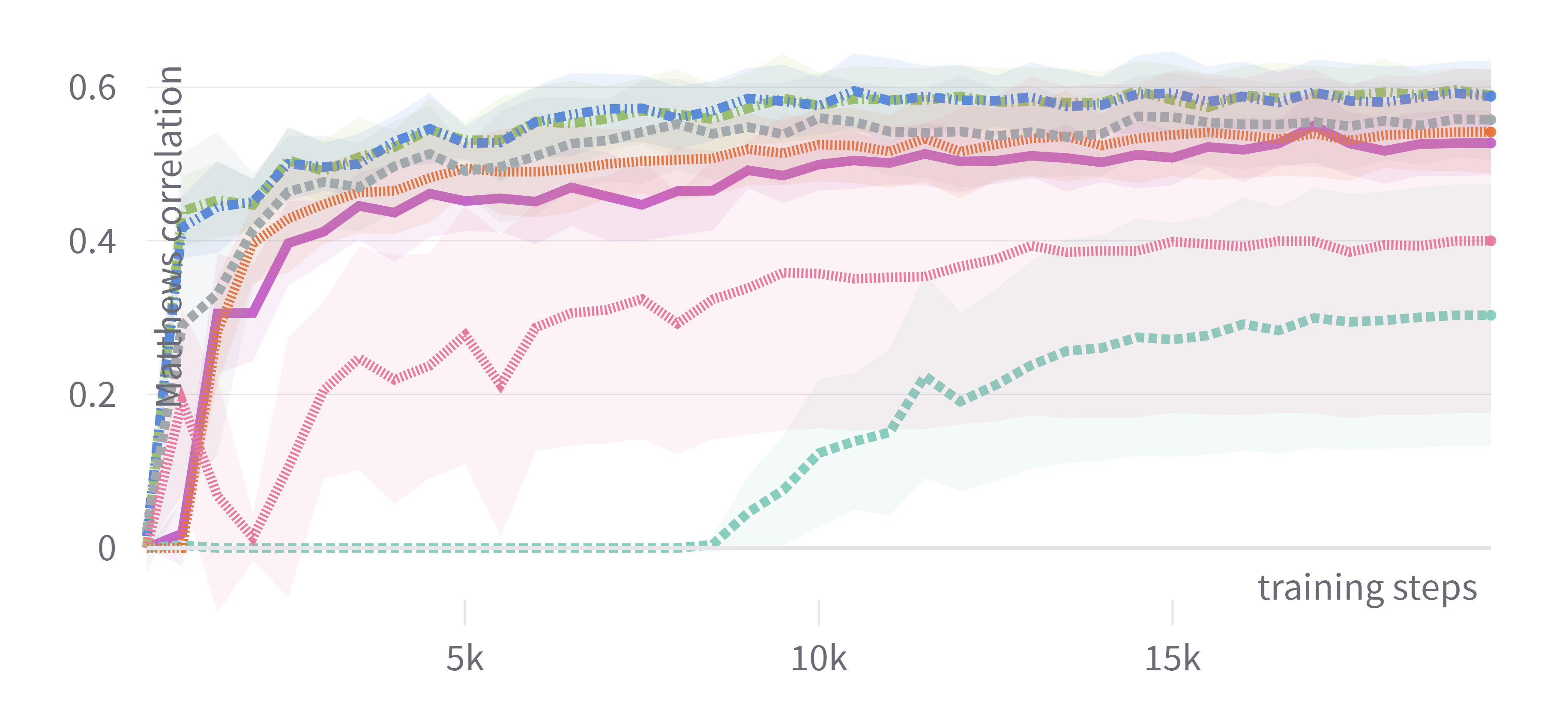}
        \caption{CoLA}
        \label{subfig:distilrobertacolamathlrtuned}
        \vspace{0.5mm}
    \end{subfigure}
    \hfill
    \begin{subfigure}[b]{0.49\textwidth}
        \centering
        \includegraphics[width=\textwidth]{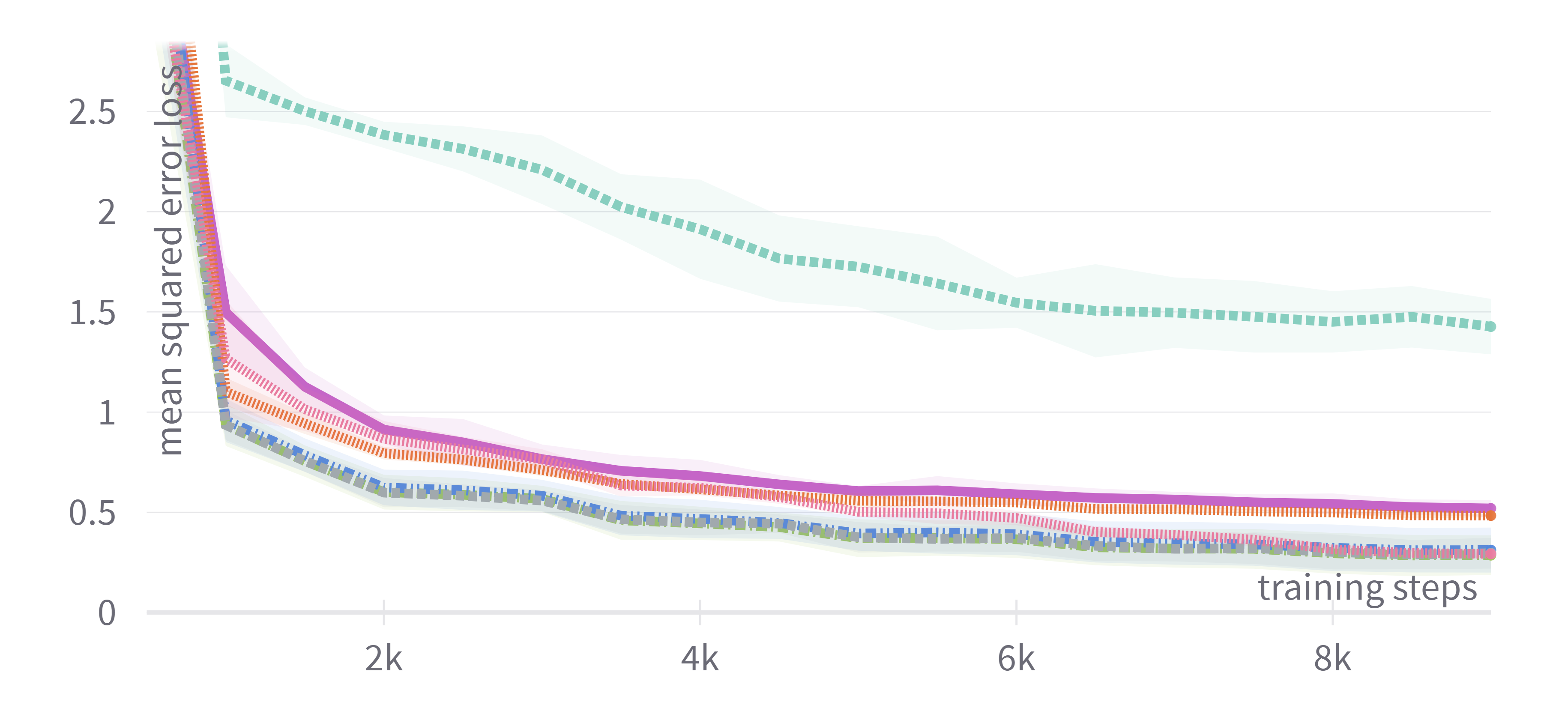}
        \caption{STS-B}
        \label{subfig:distilrobertastsblosslrtuned}
        \vspace{0.5mm}
    \end{subfigure}
    \hfill\addtocounter{subfigure}{-1}
    \begin{subfigure}[b]{0.49\textwidth}
        \centering
        \includegraphics[width=\textwidth]{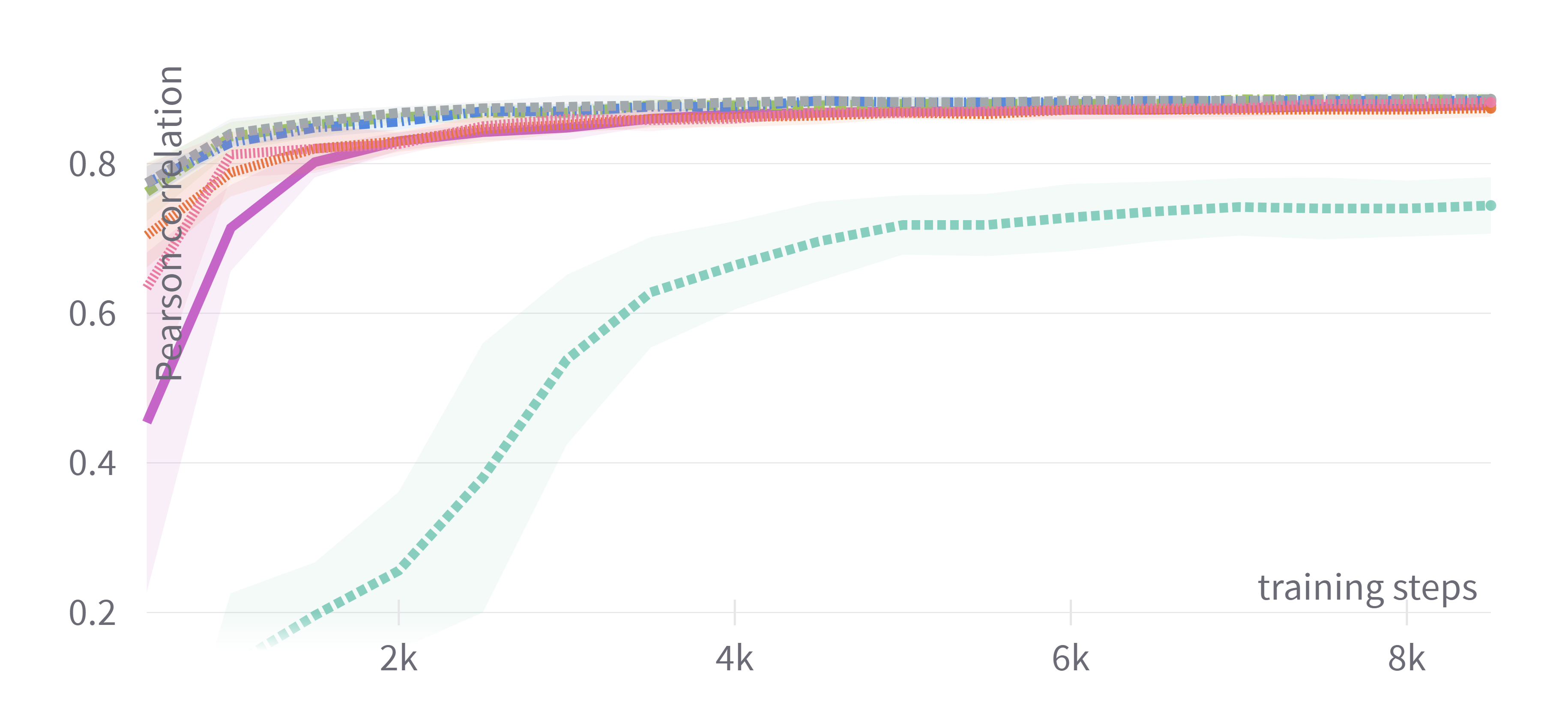}
        \caption{STS-B}
        \label{subfig:distilrobertastsbpearlrtuned}
        \vspace{0.5mm}
    \end{subfigure}
    \hfill
    \begin{subfigure}[b]{0.49\textwidth}
        \centering
        \includegraphics[width=\textwidth]{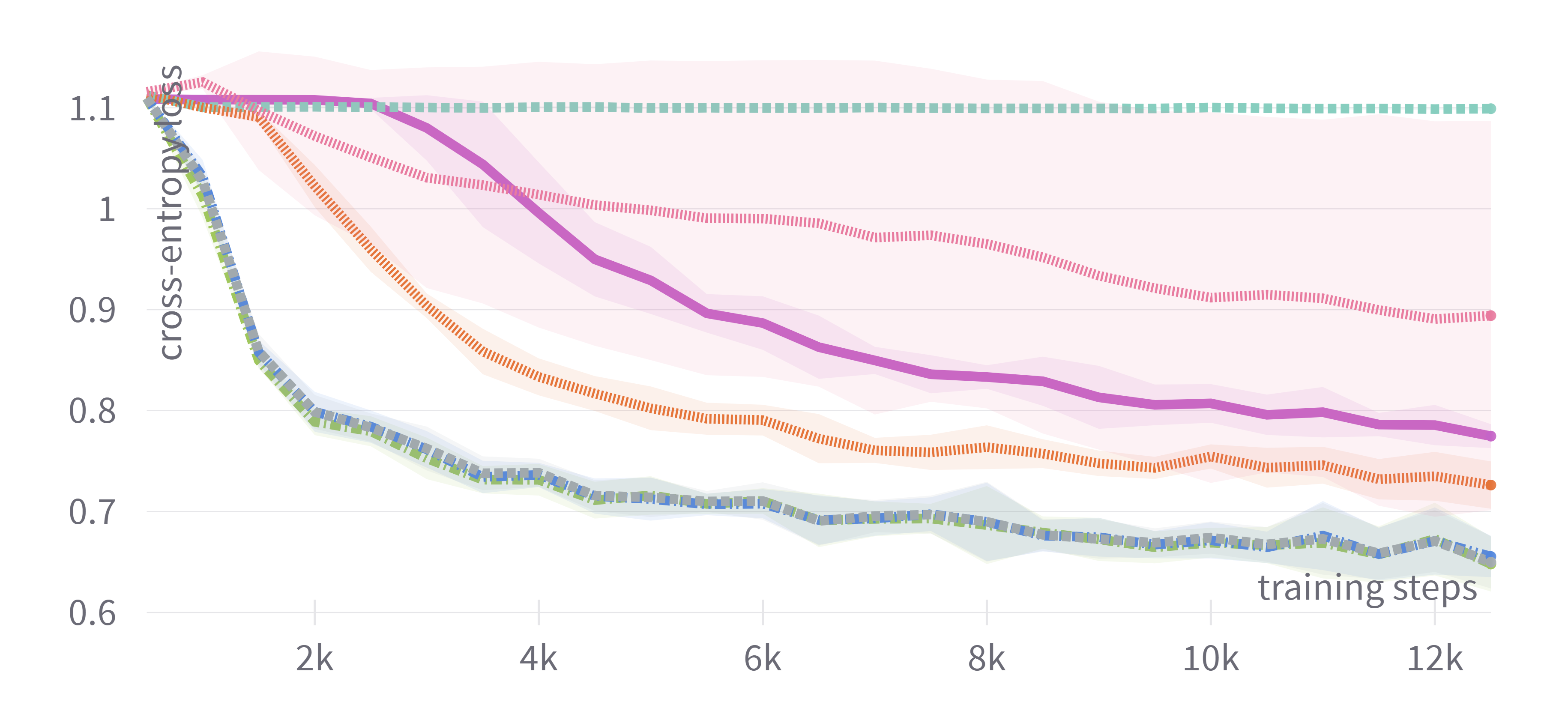}
        \caption{MNLI}
        \label{subfig:distilrobertamnlilosslrtuned}
    \end{subfigure}
    \hfill\addtocounter{subfigure}{-1}
    \begin{subfigure}[b]{0.49\textwidth}
        \centering
        \includegraphics[width=\textwidth]{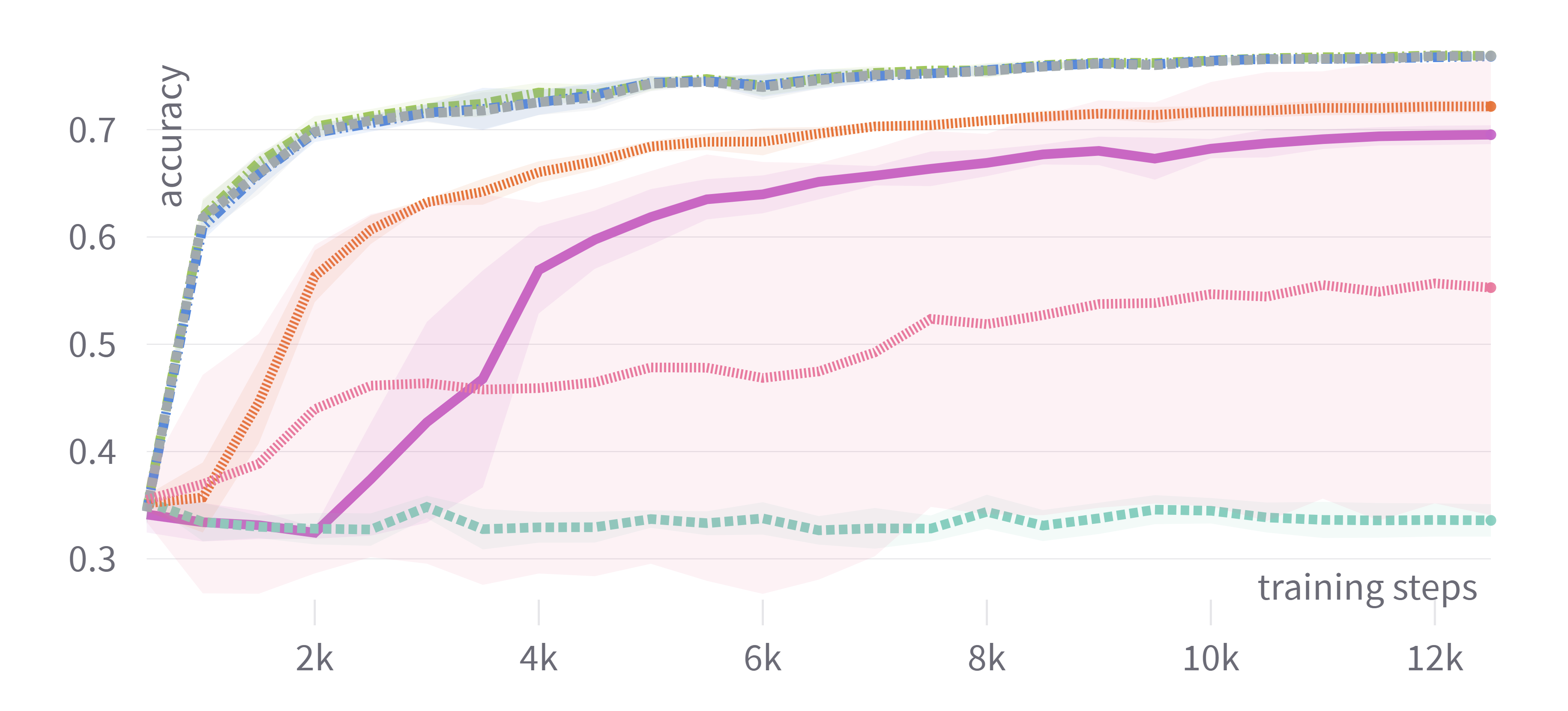}
        \caption{MNLI}
        \label{subfig:distilrobertamnliacclrtuned}
    \end{subfigure}

\vspace*{-2mm}
\caption{\textbf{Training loss} (left) and \textbf{evaluation score on development data} (right) having \textbf{tuned only the learning rate}, using \underline{\textbf{DistilRoBERTa}}. For each dataset, we use \textbf{five random data splits}, and plot the \textbf{average} and \textbf{standard deviation} (shadow). As in the corresponding DistilBERT experiments (Fig.~\ref{fig:tunedlr_curves}), \textbf{AdaMax} lags (now slightly) behind on CoLA and (more clearly) on MNLI in terms of development scores. The only important difference from Fig.~\ref{fig:tunedlr_curves} is that \textbf{AdaBound} is now also clearly worse than the other adaptive optimizers in development scores on CoLA and MNLI, where it is outperformed even by SGDM. Hence, these experiment confirm that \textbf{tuning only the learning rate} of adaptive optimizers is \textbf{in most cases (not always) as good as tuning all their hyperparameters}. 
}
\label{fig:distilroberta_tunedlr_curves}

\end{figure*}

Figure~\ref{fig:distilroberta_tunedlr_curves} shows the training losses and development scores when \emph{only the learning rate} is tuned, as in Fig.~\ref{fig:tunedlr_curves}, but now using DistilRoBERTa instead of DistilBERT. As in Fig.~\ref{fig:tunedlr_curves}, AdaMax lags behind (but now only slightly) on CoLA and (more clearly) on MNLI in terms of development scores. The only important difference compared to Fig.~\ref{fig:tunedlr_curves} is that AdaBound is now also clearly worse than the other adaptive optimizers in development scores on CoLA and MNLI, where it is outperformed even by SGDM. Hence, these experiments confirm that tuning only the learning rate of adaptive optimizers is in most cases (but not always, AdaMax and AdaBound being exceptions on CoLA and MNLI) as good as tuning all their hyperparameters. 

\begin{figure*}
\centering

\includegraphics[width=0.7\textwidth]{optimizers.png}

    \begin{subfigure}[b]{0.49\textwidth}
        \centering
        \includegraphics[width=\textwidth]{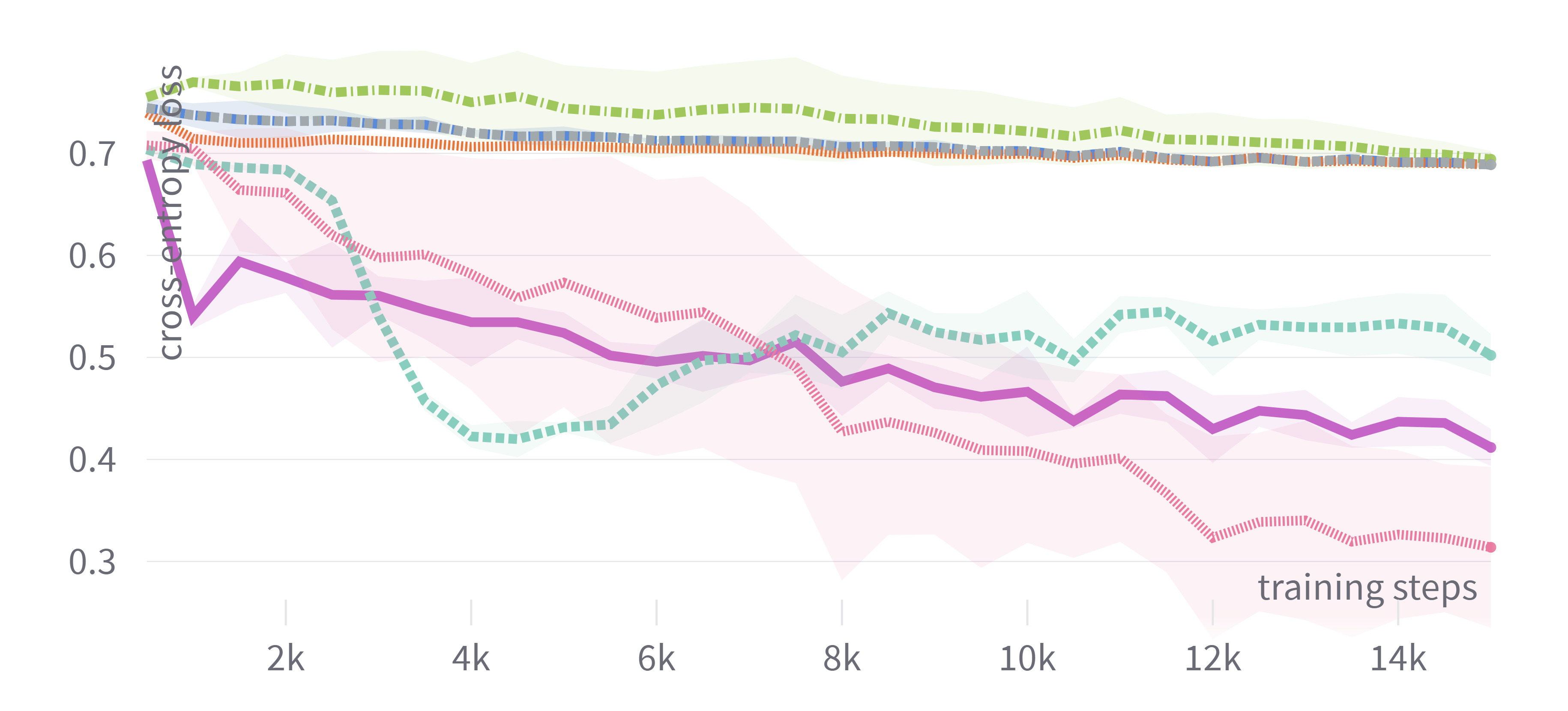}
        \caption{SST-2}
        \label{subfig:distilrobertasst2lossdef}
        \vspace{0.5mm}
    \end{subfigure}
    \hfill\addtocounter{subfigure}{-1}
    \begin{subfigure}[b]{0.49\textwidth}
        \centering
        \includegraphics[width=\textwidth]{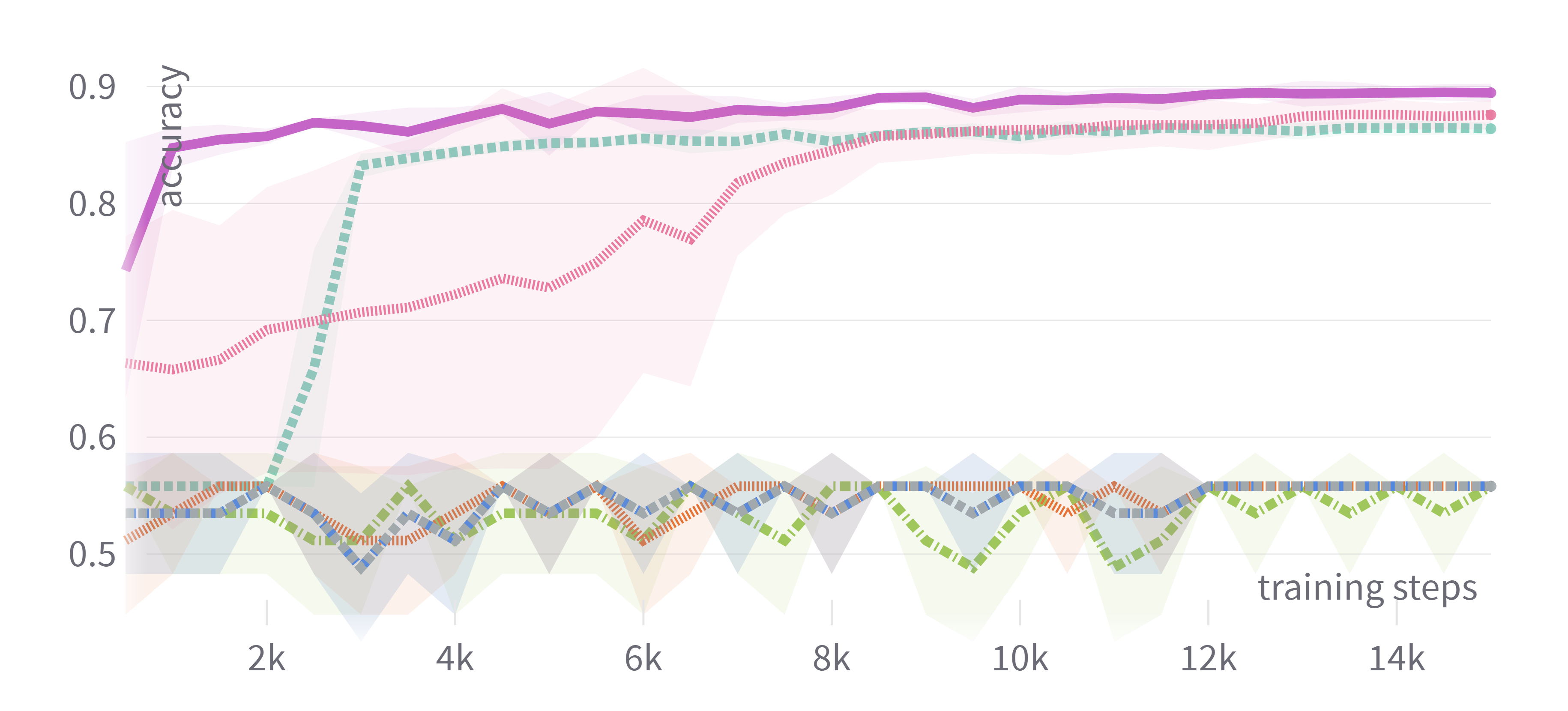}
        \caption{SST-2}
        \label{subfig:distilrobertasst2accdef}
        \vspace{0.5mm}
    \end{subfigure}
    \hfill
    \begin{subfigure}[b]{0.49\textwidth}
        \centering
        \includegraphics[width=\columnwidth]{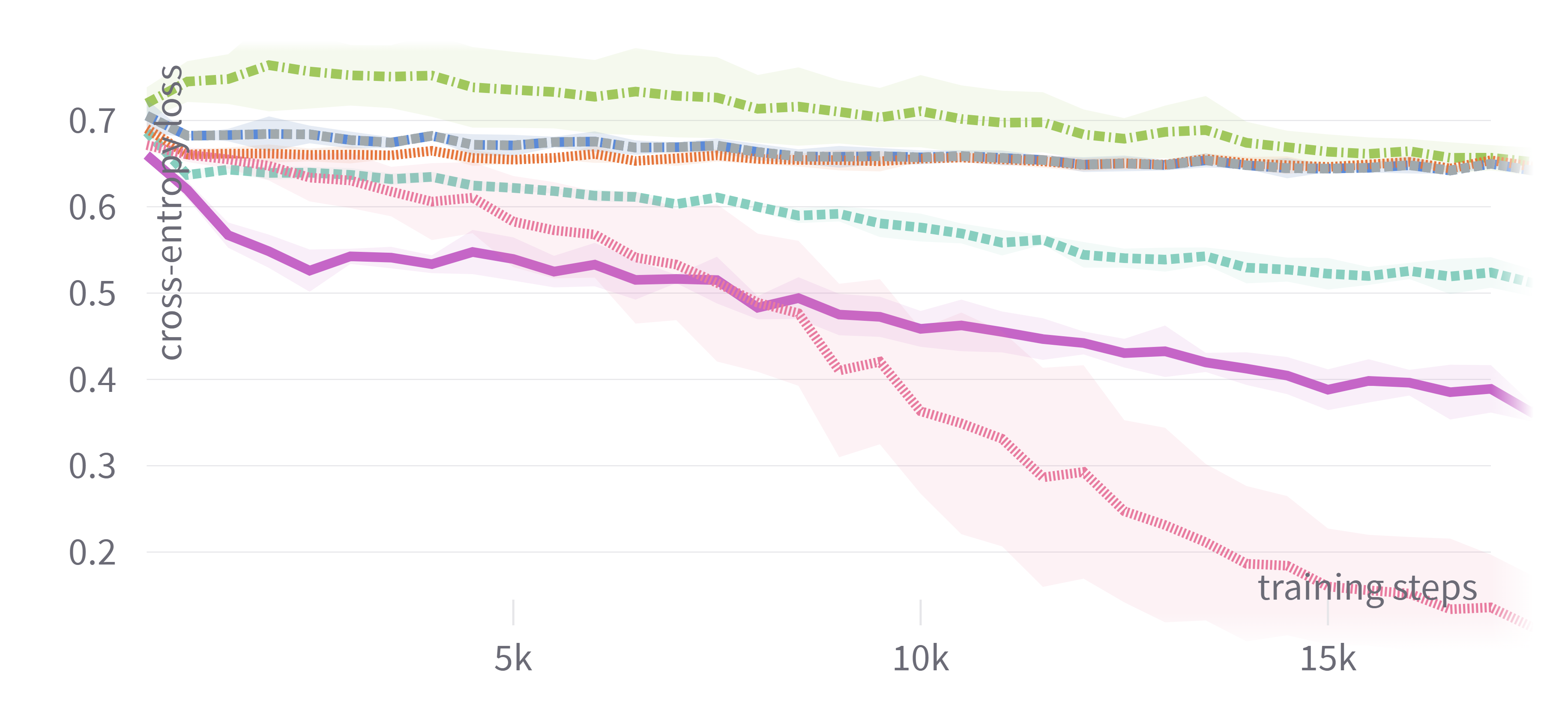}
        \caption{MRPC}
        \label{subfig:distilrobertamrpclossdef}
        \vspace{0.5mm}
    \end{subfigure}
    \hfill\addtocounter{subfigure}{-1}
    \begin{subfigure}[b]{0.49\textwidth}
        \centering
        \includegraphics[width=\textwidth]{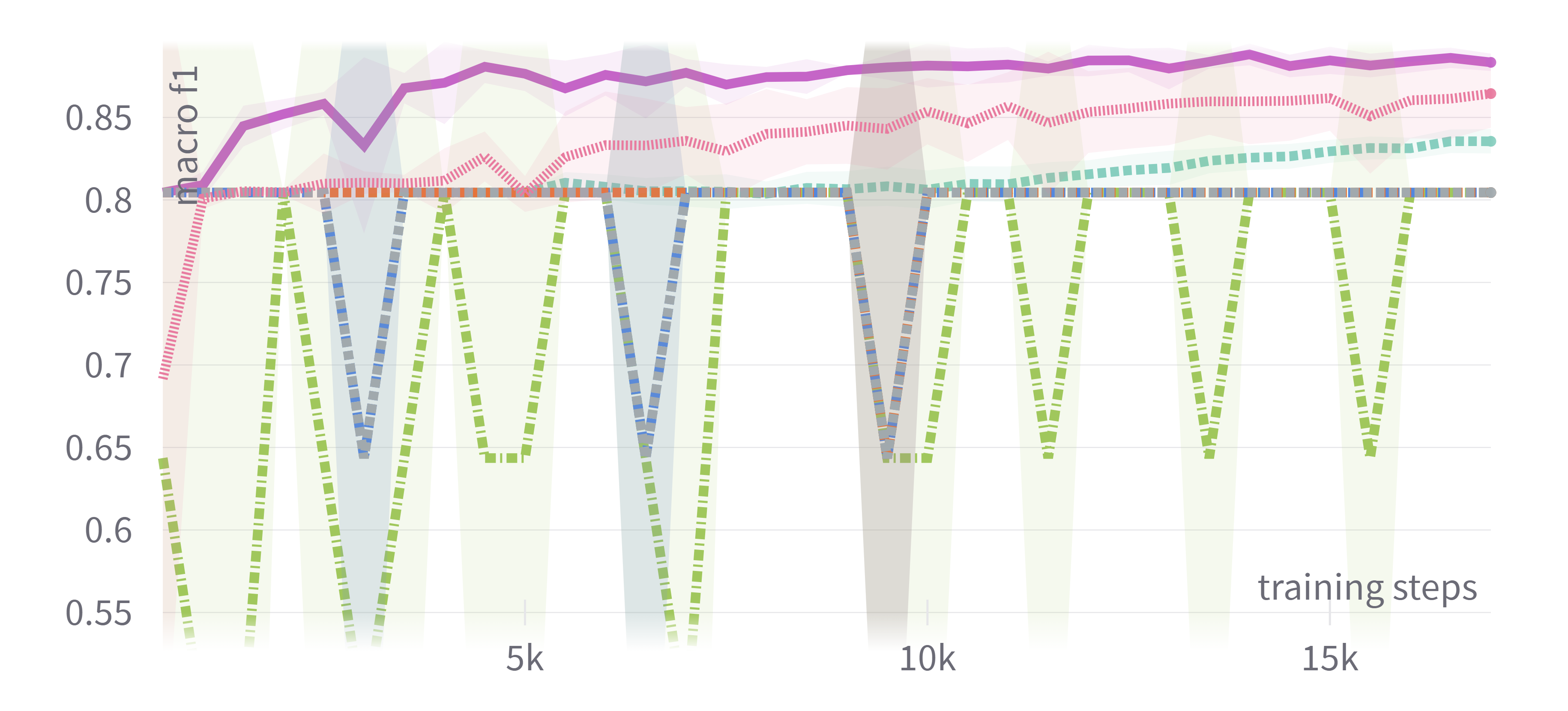}
        \caption{MRPC}
        \label{subfig:distilrobertamrpcf1def}
        \vspace{0.5mm}
    \end{subfigure}
    \hfill
    \begin{subfigure}[b]{0.49\textwidth}
        \centering
        \includegraphics[width=\textwidth]{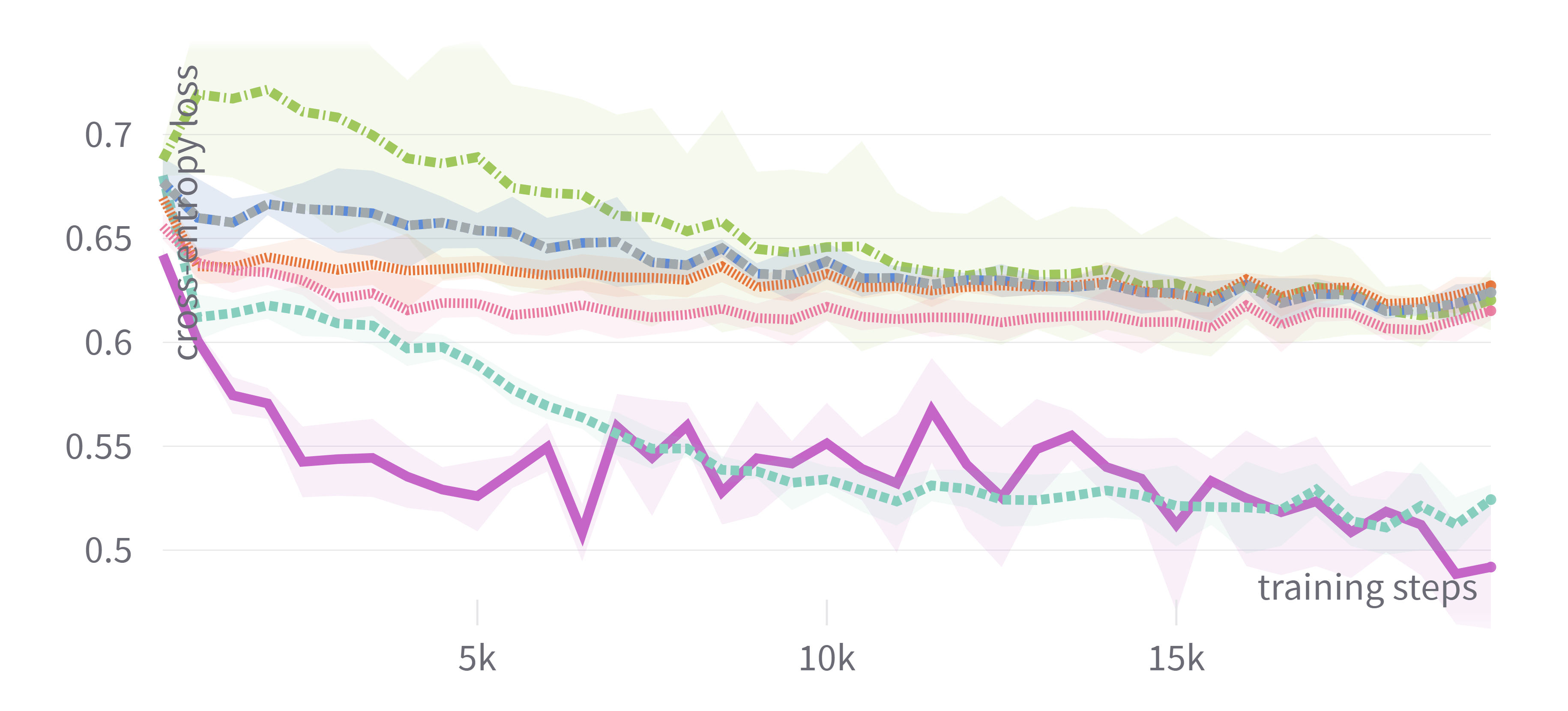}
        \caption{CoLA}
        \label{subfig:distilrobertacolalossdef}
        \vspace{0.5mm}
    \end{subfigure}
    \hfill\addtocounter{subfigure}{-1}
    \begin{subfigure}[b]{0.49\textwidth}
        \centering
        \includegraphics[width=\textwidth]{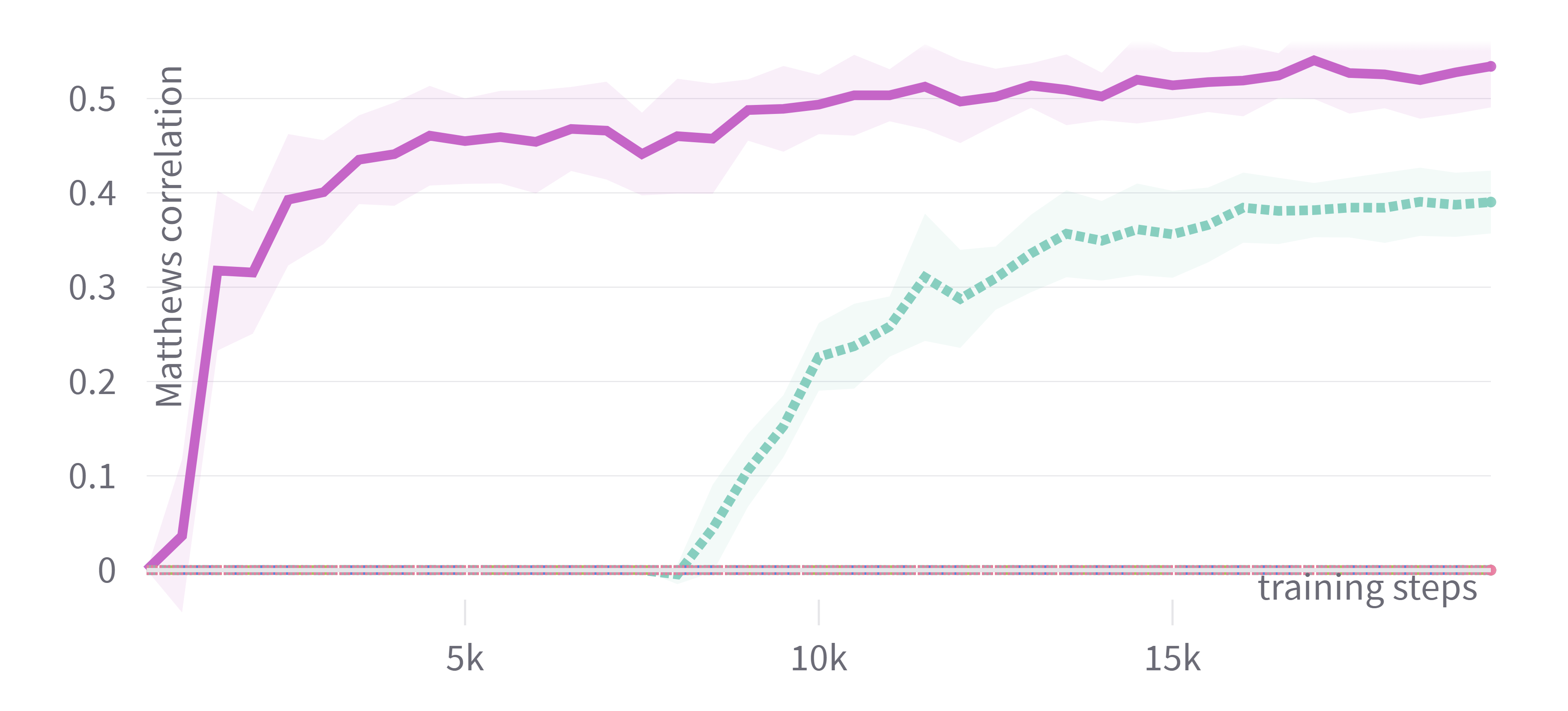}
        \caption{CoLA}
        \label{subfig:distilrobertacolamathdef}
        \vspace{0.5mm}
    \end{subfigure}
    \hfill
    \begin{subfigure}[b]{0.49\textwidth}
        \centering
        \includegraphics[width=\textwidth]{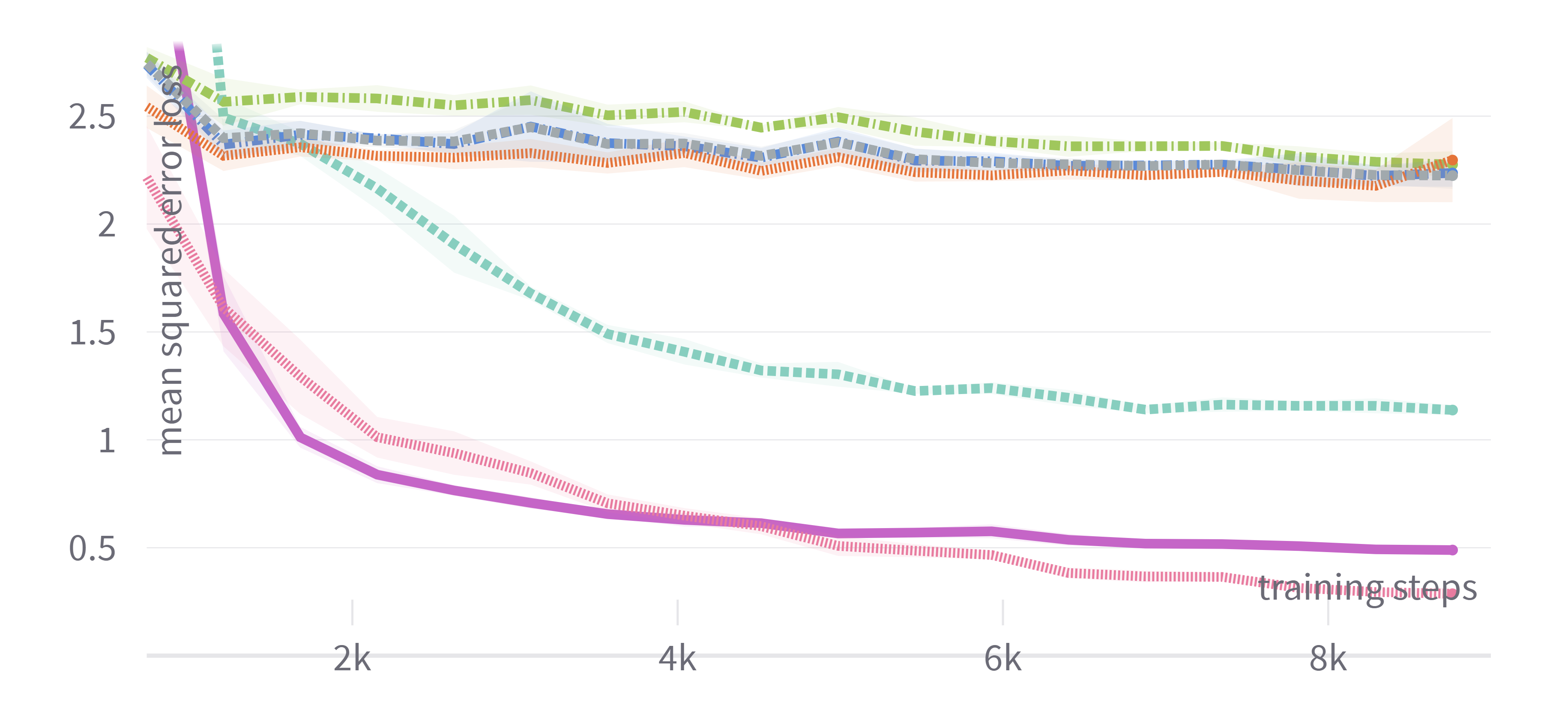}
        \caption{STS-B}
        \label{subfig:distilrobertastsblossdef}
        \vspace{0.5mm}
    \end{subfigure}
    \hfill\addtocounter{subfigure}{-1}
    \begin{subfigure}[b]{0.49\textwidth}
        \centering
        \includegraphics[width=\textwidth]{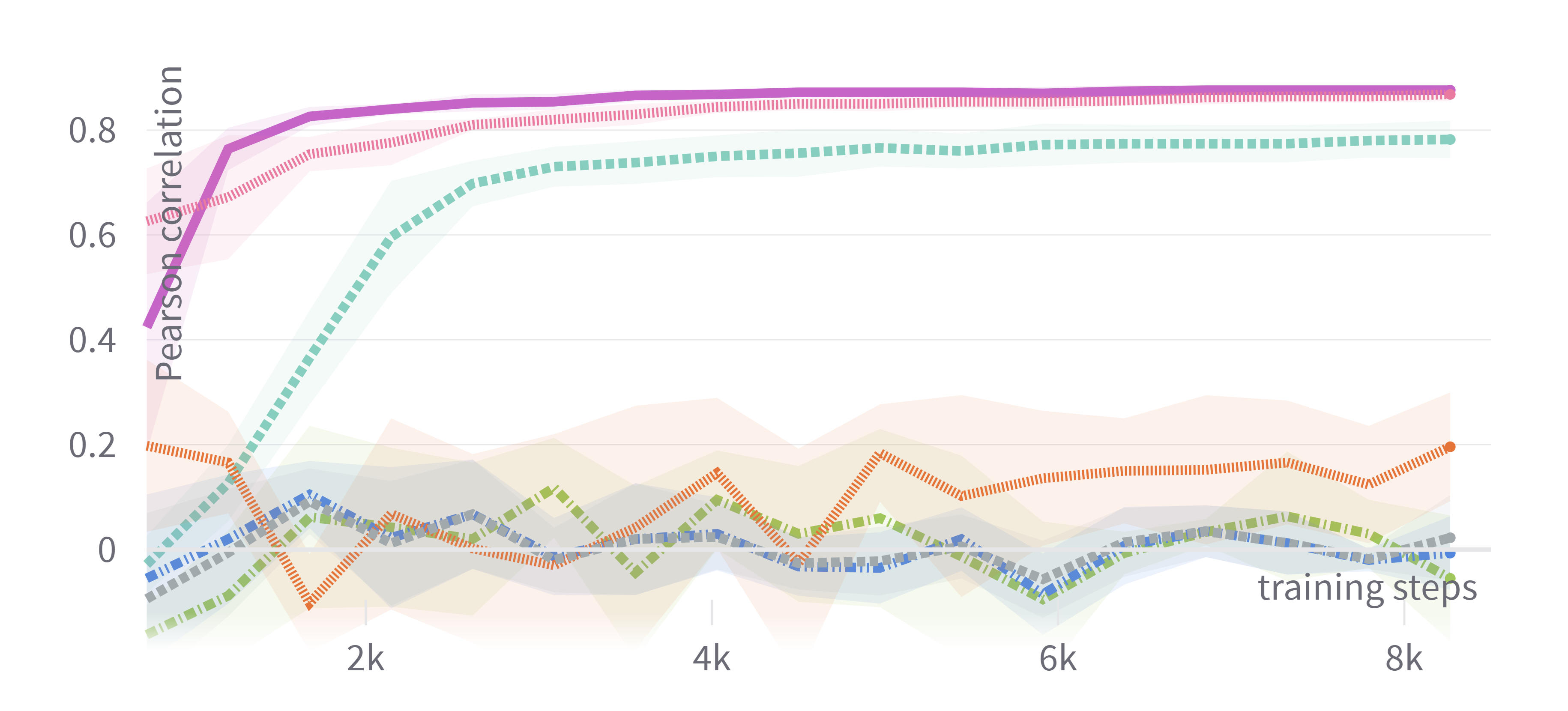}
        \caption{STS-B}
        \label{subfig:distilrobertastsbpearlossdef}
        \vspace{0.5mm}
    \end{subfigure}
    \hfill
    \begin{subfigure}[b]{0.49\textwidth}
        \centering
        \includegraphics[width=\textwidth]{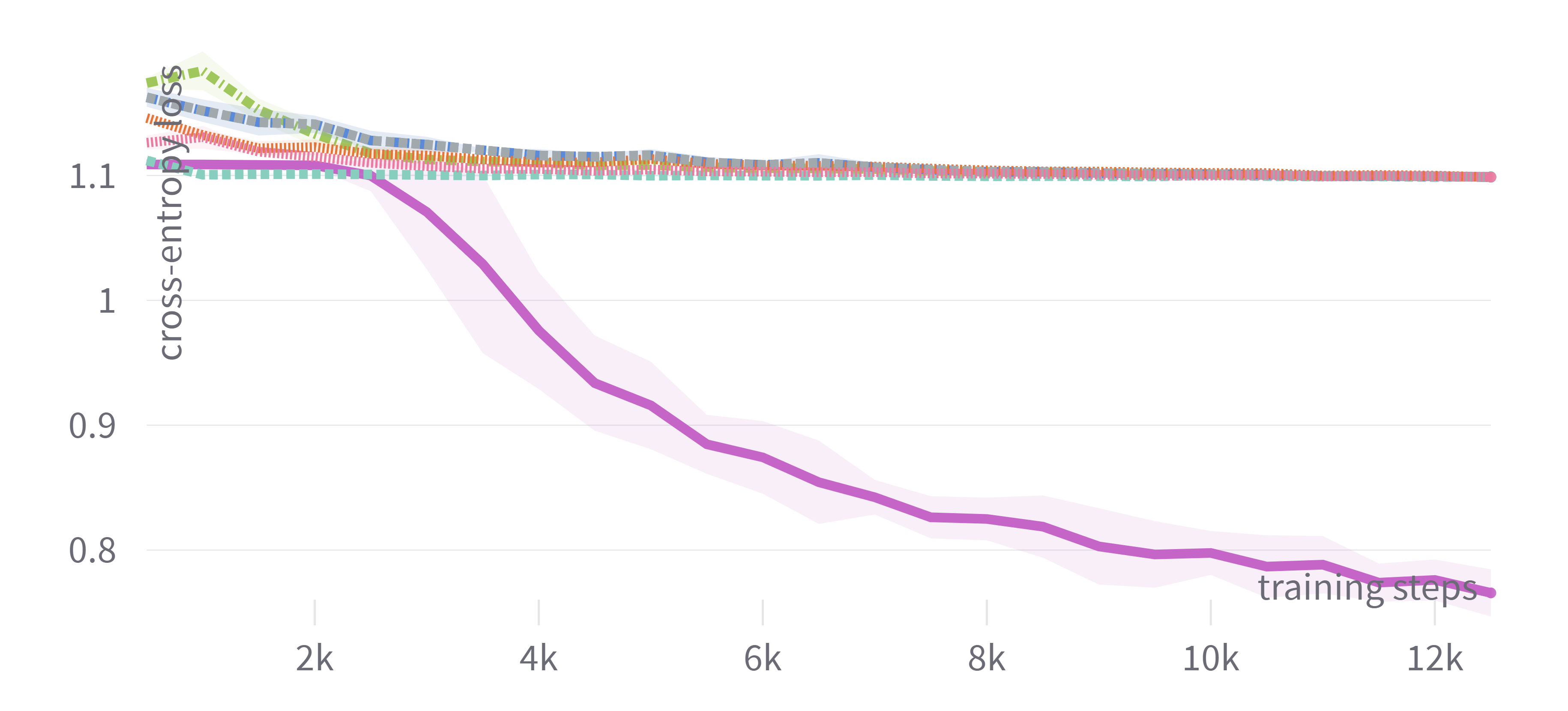}
        \caption{MNLI}
        \label{subfig:distilrobertamnlilossdef}
    \end{subfigure}
    \hfill\addtocounter{subfigure}{-1}
    \begin{subfigure}[b]{0.49\textwidth}
        \centering
        \includegraphics[width=\textwidth]{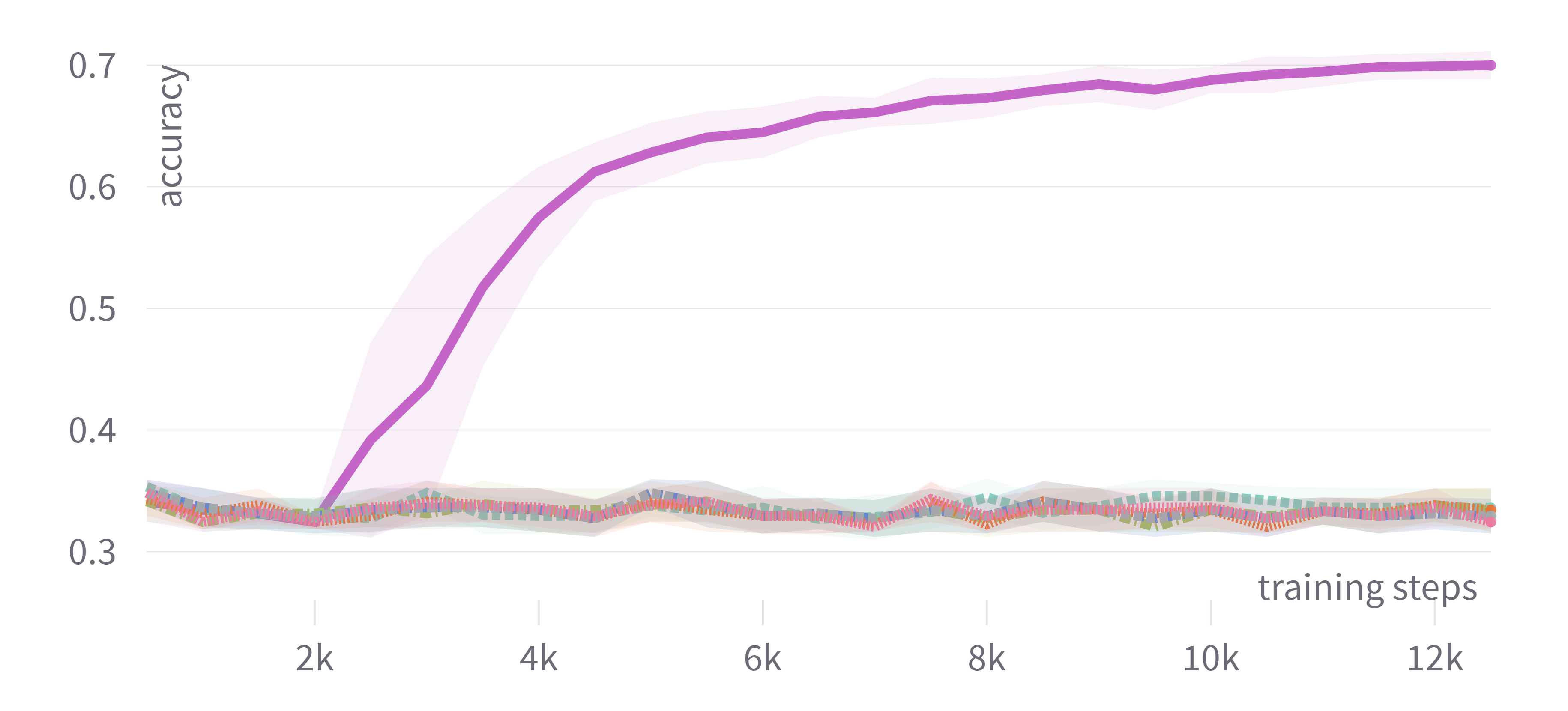}
        \caption{MNLI}
        \label{subfig:distilrobertamnliaccdef}
    \end{subfigure}

\vspace*{-2mm}
\caption{\textbf{Training loss} (left) and \textbf{evaluation score on development data} (right) with \textbf{all hyperparameters of the optimizers set to their defaults}, using \underline{\textbf{DistilRoBERTa}}. Again, we use \textbf{five random data splits}, and plot the \textbf{average} and \textbf{standard deviation} (shadow). As in the corresponding experiments with DistilBERT (Fig.~\ref{fig:default_curves}), \textbf{SGDM is now the best} in terms of development scores; again \textbf{AdaBound} eventually reaches almost the same performance as SGDM on SST-2 and STS-B, but not on the other datasets; the other adaptive optimizers perform overall poorly.
}
\label{fig:distilroberta_default_curves}

\end{figure*}

Figure~\ref{fig:distilroberta_default_curves} shows the training losses and development scores when using \emph{defaults}, as in Fig.~\ref{fig:default_curves}, but now using DistilRoBERTa instead of DistilBERT. As in Fig.~\ref{fig:default_curves}, SGDM is now the best in terms of development scores, and AdaBound is overall the second best. Again AdaBound eventually matches the development scores of SGDM on SST-2 and STSB, but not on the other datasets. The other adaptive optimizers perform overall poorly. 

Tables~\ref{table:distilrobertatunedtest}--\ref{table:distilrobertadefaulttest} show results on \emph{test data}, now using DistilRoBERTa. The best results are now slightly improved in most cases, as one might expect, compared to Tables~\ref{table:tunedtest}--\ref{table:defaulttest} where DistilBERT was used. Otherwise the conclusions are very similar to those of Tables~\ref{table:tunedtest}--\ref{table:defaulttest}, they are summarized in the captions of Tables~\ref{table:distilrobertatunedtest}--\ref{table:distilrobertadefaulttest}, and they are aligned with the findings of Fig.~\ref{fig:distilroberta_tuned_curves}--\ref{fig:distilroberta_default_curves}.

\begin{table*}[htbp]
\centering
\resizebox{\textwidth}{!}
{
\begin{tabular}{|l||c|c|c|c|c|c|c|}
 \hline
 & AdamW & AdaMax & Nadam & AdaBound &Adam& SGDM & SGD \\
 \hline
 $\epsilon$ & $[1\mathrm{e-}7, 1\mathrm{e-}5]$ & $[1\mathrm{e-}7, 1\mathrm{e-}5]$ & $[1\mathrm{e-}7, 1\mathrm{e-}5]$ & $[1\mathrm{e-}7, 1\mathrm{e-}5]$ & $[1\mathrm{e-}7, 1\mathrm{e-}5]$ & $[1\mathrm{e-}7, 1\mathrm{e-}3]$ & $[1\mathrm{e-}7, 1\mathrm{e-}3]$\\
 $\rho_1$ & $[0.8, 0.95]$ & $[0.8,0.95]$ & $[0.8,0.95]$ & $[0.8,0.95]$ & $[0.8,0.95]$ & -- & --\\
 $\rho_2$ & $[0.9,0.99999]$ & $[0.9,0.99999]$ & $[0.9,0.99999]$ & $[0.9, 0.99999]$ & $[0.9,0.99999]$ & -- & -- \\
 $\delta$ & $[1\mathrm{e-}9,1\mathrm{e-}7]$ & $[1\mathrm{e-}9,1\mathrm{e-}7]$ & $[1\mathrm{e-}9,1\mathrm{e-}7]$ & $[1\mathrm{e-}9,1\mathrm{e-}7]$ & $[1\mathrm{e-}9,1\mathrm{e-}7]$ & -- & -- \\
 $\alpha$ & -- & -- & $[1\mathrm{e-}4,1\mathrm{e-}2]$ & -- & -- & $[0.7, 0.9999]$ & --\\
 $\epsilon^*$ & -- & -- & -- & $[1\mathrm{e-}2,1\mathrm{e-}1]$ & -- & -- & --\\
 $\gamma$  & -- & -- & -- & $[1\mathrm{e-}4,2\mathrm{e-}3]$ & -- & -- & --\\
\hline
\end{tabular}
}
\caption{\textbf{Hyperparameter search space} for all the optimization algorithms. $\epsilon^*$ and $\gamma$ (not shown in Algorith~\ref{adamax_adabound}) are used by AdaBound's lower and upper bound functions ($\eta_{t}^{l}$, $\eta_{t}^{u}$); $\gamma$ controls the convergence speed of these functions, and $\epsilon^*$ is the learning rate used in the final training stages, where AdaBound transforms to SGD.}
\label{table:hypersearch}
\end{table*}

\begin{table*}[htbp]
\centering
{\small
\begin{tabular}{|l||c|c|c|c|c|c|c|}
 \hline
  & AdamW & AdaMax & Nadam & AdaBound &Adam& SGDM & SGD \\
 \hline
 $\epsilon$ & $1\mathrm{e-}3$ & $2\mathrm{e-}3$ & $2\mathrm{e-}3$ & $1\mathrm{e-}3$ & $1\mathrm{e-}3$ & $1\mathrm{e-}3$ & $1\mathrm{e-}3$\\
 $\rho_1$ & $0.9$ & $0.9$ & $0.9$ & $0.9$ & $0.9$ & -- & --\\
 $\rho_2$ & $0.999$ & $0.999$ & $0.999$ & $0.999$ & $0.999$ & -- & -- \\
 $\delta$ & $1\mathrm{e-}8$ & $1\mathrm{e-}8$ & $1\mathrm{e-}8$ & $1\mathrm{e-}8$ & $1\mathrm{e-}8$ & -- & -- \\
 $\alpha$ & -- & -- & $4\mathrm{e-}3$ & -- & -- & $0.9$ & --\\
 $\epsilon^*$ & -- & -- & -- & $0.1$ & -- & -- & --\\
 $\gamma$  & -- & -- & -- & $1\mathrm{e-}3$ & -- & -- & --\\
 \hline
\end{tabular}
} 
\caption{\textbf{Default hyperparameter values} per optimizer. The search space for the learning rate ($\epsilon$) does not include the defaults for adaptive optimizers, because it is standard practice when fine-tuning Transformers with adaptive optimizers to use much smaller learning rates.}
\label{table:hyperdef}
\end{table*}

\begin{table*}[htbp]
\centering
{\small
\begin{tabular}{|l||c|c|c|c|c|c|c|}
 \hline
  & SST-2 & MRPC & CoLA & MNLI & STSB & Default & Search Space\\
 \hline
AdaBound & $8.17\mathrm{e-}6$ & $4.51\mathrm{e-}6$ & $8.27\mathrm{e-}6$ & $1.26\mathrm{e-}6$ & $2.06\mathrm{e-}6$ & $1\mathrm{e-}3$ & $[1\mathrm{e-}7, 1\mathrm{e-}5]$ \\
AdamW    & $9.80\mathrm{e-}6$ & $6.97\mathrm{e-}6$ & $5.59\mathrm{e-}6$ & $9.50\mathrm{e-}6$ & $7.99\mathrm{e-}6$ & $1\mathrm{e-}3$ & $[1\mathrm{e-}7, 1\mathrm{e-}5]$ \\
AdaMax   & $8.78\mathrm{e-}6$ & $8.44\mathrm{e-}6$ & $6.59\mathrm{e-}6$ & $9.19\mathrm{e-}6$ & $7.58\mathrm{e-}6$ & $2\mathrm{e-}3$ & $[1\mathrm{e-}7, 1\mathrm{e-}5]$ \\
Nadam    & $9.74\mathrm{e-}6$ & $6.46\mathrm{e-}6$ & $6.22\mathrm{e-}6$ & $9.58\mathrm{e-}6$ & $6.89\mathrm{e-}6$ & $2\mathrm{e-}3$ & $[1\mathrm{e-}7, 1\mathrm{e-}5]$ \\
Adam     & $9.61\mathrm{e-}6$ & $6.54\mathrm{e-}6$ & $7.09\mathrm{e-}6$ & $9.47\mathrm{e-}6$ & $9.13\mathrm{e-}6$ & $1\mathrm{e-}3$ & $[1\mathrm{e-}7, 1\mathrm{e-}5]$ \\
SGDM     & $8.03\mathrm{e-}4$ & $3.81\mathrm{e-}4$ & $3.41\mathrm{e-}4$ & $3.54\mathrm{e-}4$ & $6.49\mathrm{e-}4$ & $1\mathrm{e-}3$ & $[1\mathrm{e-}7, 1\mathrm{e-}3]$ \\
SGD      & $9.66\mathrm{e-}4$ & $7.69\mathrm{e-}4$ & $1.68\mathrm{e-}4$ & $9.58\mathrm{e-}4$ & $9.77\mathrm{e-}4$ & $1\mathrm{e-}3$ & $[1\mathrm{e-}7, 1\mathrm{e-}3]$ \\
 \hline
\end{tabular}
} 
\caption{\textbf{Tuned learning rate} ($\epsilon$) per optimizer and dataset, \textbf{averaged over the five random splits}, when \textbf{tuning all hyperparameters}, using \textbf{DistilBERT}. Default $\epsilon$ and search space also shown. All the tuned values are far from the defaults. The \textbf{tuned learning rate of SGDM} (and SGD) is \textbf{closer to the default}, comparing to the other optimizers, which may explain why SGDM is the best optimizer overall when using the optimizers with defaults.}
\label{table:lr}
\end{table*}

\begin{table*}[htbp]
\centering
{\small 
\begin{tabular}{|l||c|c|c|c|c|c|c|}
 \hline
  & SST-2 & MRPC & CoLA & MNLI & STSB & Default & Search Space\\
 \hline
AdaBound & $0.87$ & $0.90$ & $0.86$ & $0.86$ & $0.88$ & $0.90$ & $[0.80, 0.95]$ \\
AdamW    & $0.92$ & $0.85$ & $0.88$ & $0.88$ & $0.88$ & $0.90$ & $[0.80, 0.95]$ \\
AdaMax   & $0.90$ & $0.88$ & $0.87$ & $0.86$ & $0.86$ & $0.90$ & $[0.80, 0.95]$ \\
Nadam    & $0.93$ & $0.85$ & $0.82$ & $0.86$ & $0.88$ & $0.90$ & $[0.80, 0.95]$ \\
Adam     & $0.89$ & $0.86$ & $0.88$ & $0.88$ & $0.87$ & $0.90$ & $[0.80, 0.95]$ \\
\hline
\end{tabular}
} 
\caption{\textbf{Tuned 1st momentum decay rate} ($\rho_1$) per optimizer (when applicable) and dataset, \textbf{averaged over the five random splits}, when \textbf{tuning all hyperparameters}, using \textbf{DistilBERT}. Default $\rho_1$ and search space also shown.
}
\label{table:beta1}
\end{table*}

\begin{table*}[htbp]
\centering
{\small
\begin{tabular}{|l||c|c|c|c|c|c|c|}
 \hline
  & SST-2 & MRPC & CoLA & MNLI & STSB & Default & Search Space\\
 \hline
AdaBound & $0.93$ & $0.94$ & $0.95$ & $0.93$  & $0.96$ & $0.999$ & $[0.9,0.99999]$ \\
AdamW    & $0.92$ & $0.96$ & $0.95$ & $0.98$  & $0.94$ & $0.999$ & $[0.9,0.99999]$ \\
AdaMax   & $0.92$ & $0.97$ & $0.94$ & $0.93$  & $0.91$ & $0.999$ & $[0.9,0.99999]$ \\
Nadam    & $0.93$ & $0.96$ & $0.94$ & $0.95$  & $0.96$ & $0.999$ & $[0.9,0.99999]$ \\
Adam     & $0.96$ & $0.95$ & $0.95$ & $0.95$  & $0.93$ & $0.999$ & $[0.9,0.99999]$ \\
 \hline
\end{tabular}
} 
\caption{\textbf{Tuned 2nd momentum decay rate} ($\rho_2$) per optimizer (when applicable) and dataset, \textbf{averaged over the five random splits}, when \textbf{tuning all hyperparameters}, using \textbf{DistilBERT}. Default $\rho_2$ and search space also shown. 
}
\label{table:beta2}
\end{table*}

\begin{table*}[htbp]
\centering
{\small 
\begin{tabular}{|l||c|c|c|c|c|c|c|}
 \hline
  & SST-2 & MRPC & CoLA & MNLI & STSB & Default & Search Space\\
 \hline
AdaBound & $3.31\mathrm{e-}08$ & $2.09\mathrm{e-}08$ & $2.88\mathrm{e-}08$ & $4.18\mathrm{e-}08$ & $2.36\mathrm{e-}08$ & $1\mathrm{e-}8$ & $[1\mathrm{e-}9,1\mathrm{e-}7]$ \\
AdamW    & $6.44\mathrm{e-}08$ & $3.37\mathrm{e-}08$ & $2.92\mathrm{e-}09$ & $3.53\mathrm{e-}08$ & $2.77\mathrm{e-}08$ & $1\mathrm{e-}8$ & $[1\mathrm{e-}9,1\mathrm{e-}7]$ \\
AdaMax   & $2.18\mathrm{e-}08$ & $4.71\mathrm{e-}08$ & $5.59\mathrm{e-}09$ & $2.95\mathrm{e-}08$ & $2.97\mathrm{e-}08$ & $1\mathrm{e-}8$ & $[1\mathrm{e-}9,1\mathrm{e-}7]$ \\
Nadam    & $1.22\mathrm{e-}08$ & $1.78\mathrm{e-}08$ & $6.24\mathrm{e-}08$ & $7.29\mathrm{e-}09$ & $3.46\mathrm{e-}08$ & $1\mathrm{e-}8$ & $[1\mathrm{e-}9,1\mathrm{e-}7]$ \\
Adam     & $2.49\mathrm{e-}08$ & $2.67\mathrm{e-}08$ & $1.45\mathrm{e-}08$ & $4.23\mathrm{e-}08$ & $2.19\mathrm{e-}08$ & $1\mathrm{e-}8$ & $[1\mathrm{e-}9,1\mathrm{e-}7]$ \\

 \hline
\end{tabular}
} 
\caption{\textbf{Tuned small constant} $\delta$ for each optimizer (when applicable) and dataset, \textbf{averaged over the five random splits}, when \textbf{tuning all  hyperparameters}, using \textbf{DistilBERT}. Default $\delta$ and search space also shown. 
}
\label{table:delta}
\end{table*}

\begin{table*}[htbp]
\centering
{\small 
\begin{tabular}{|l||c|c|c|c|c|c|c|}
 \hline
  & SST-2 & MRPC & CoLA & MNLI & STSB & Default & Search Space\\
 \hline
Nadam & $1.69\mathrm{e-}3$ & $3.21\mathrm{e-}3$ & $3.73\mathrm{e-}3$ & $1.01\mathrm{e-}4$ & $4.35\mathrm{e-}3$ & $4\mathrm{e-}3$ & $[1\mathrm{e-}4,1\mathrm{e-}2]$ \\
SGDM  & $0.93$             & $0.97$             & $0.86$             & $0.99$             & $0.98$             & $0.9$           & $[0.7,0.9999]$   \\

 \hline
\end{tabular}
} 
\caption{\textbf{Tuned momentum strength} ($\alpha$) per optimizer (when applicable) and dataset, \textbf{averaged over the five random splits}, when \textbf{tuning all  hyperparameters}, using \textbf{DistilBERT}. Default $\alpha$ and search space also shown.
}
\label{table:mom}
\end{table*}

\begin{table*}[htbp]
\centering
{\small
\begin{tabular}{|l||c|c|c|c|c|c|c|}
 \hline
 AdaBound & SST-2 & MRPC & CoLA & MNLI & STSB & Default & Search Space\\
 \hline
$\epsilon^*$ & $7e-2$     & $7.19\mathrm{e-}2$ & $6.35\mathrm{e-}2$ & $1.22\mathrm{e-}1$ & $1.1\mathrm{e-}1$  & $4\mathrm{e-}3$ & $[1\mathrm{e-}2,1\mathrm{e-}1]$ \\
$\gamma$     & $3.03e-4$  & $7.21\mathrm{e-}4$ & $2.04\mathrm{e-}4$ & $9.92\mathrm{e-}4$ & $8.46\mathrm{e-}4$ & $0.9$           & $[1\mathrm{e-}4,2\mathrm{e-}3]$   \\

 \hline
\end{tabular}
} 
\caption{\textbf{Tuned AdaBound-specific hyperparameters} ($\epsilon^*$ and $\gamma$) per dataset,  \textbf{averaged over the five random splits}, when \textbf{tuning all  hyperparameters}, using \textbf{DistilBERT}. $\epsilon^*$ and $\gamma$ are used by AdaBound's lower and upper bound functions ($\eta_{t}^{l}$, $\eta_{t}^{u}$), $\gamma$ controls the convergence speed of these functions, and $\epsilon^*$ is the learning rate used in the final training stages, where AdaBound transforms to SGD. Default values and search space are also shown.
}
\label{table:gamma}
\end{table*}

\begin{table*}[htbp]
\centering
{\small
\begin{tabular}{|l||c|c|c|c|c|c|c|}
 \hline
  & SST-2 & MRPC & CoLA & MNLI & STSB & Default & Search Space\\
 \hline
AdaBound & $2.43\mathrm{e-}6$ & $1.87\mathrm{e-}6$ & $1.58\mathrm{e-}6$ & $5.42\mathrm{e-}6$ & $1.72\mathrm{e-}7$ & $1\mathrm{e-}3$ & $[1\mathrm{e-}7, 1\mathrm{e-}5]$ \\
AdamW    & $9.20\mathrm{e-}6$ & $7.57\mathrm{e-}6$ & $7.63\mathrm{e-}6$ & $9.66\mathrm{e-}6$ & $8.68\mathrm{e-}6$ & $1\mathrm{e-}3$ & $[1\mathrm{e-}7, 1\mathrm{e-}5]$ \\
AdaMax   & $9.67\mathrm{e-}6$ & $8.46\mathrm{e-}6$ & $8.04\mathrm{e-}6$ & $9.74\mathrm{e-}6$ & $8.80\mathrm{e-}6$ & $2\mathrm{e-}3$ & $[1\mathrm{e-}7, 1\mathrm{e-}5]$ \\
Nadam    & $9.63\mathrm{e-}6$ & $8.16\mathrm{e-}6$ & $7.83\mathrm{e-}6$ & $9.68\mathrm{e-}6$ & $8.53\mathrm{e-}6$ & $2\mathrm{e-}3$ & $[1\mathrm{e-}7, 1\mathrm{e-}5]$ \\
Adam     & $9.11\mathrm{e-}6$ & $7.99\mathrm{e-}6$ & $5.49\mathrm{e-}6$ & $9.58\mathrm{e-}6$ & $7.42\mathrm{e-}6$ & $1\mathrm{e-}3$ & $[1\mathrm{e-}7, 1\mathrm{e-}5]$ \\
SGDM     & $9.56\mathrm{e-}4$ & $8.35\mathrm{e-}4$ & $7.43\mathrm{e-}4$ & $9.35\mathrm{e-}4$ & $9.42\mathrm{e-}4$ & $1\mathrm{e-}3$ & $[1\mathrm{e-}7, 1\mathrm{e-}3]$ \\
SGD      & $9.66\mathrm{e-}4$ & $7.69\mathrm{e-}4$ & $1.68\mathrm{e-}4$ & $9.58\mathrm{e-}4$ & $9.77\mathrm{e-}4$ & $1\mathrm{e-}3$ & $[1\mathrm{e-}7, 1\mathrm{e-}3]$ \\
 \hline
\end{tabular}
} 
\caption{\textbf{Tuned learning rate} ($\epsilon$) per optimizer and dataset, \textbf{averaged over the five random splits}, when \textbf{tuning only the learning rate}, using DistilBERT. Default $\epsilon$ and search space also shown. As in Table~\ref{table:lr}, the \textbf{tuned learning rate of SGDM} (and SGD) is \textbf{closer to the default}, comparing to the other optimizers, which may explain why SGDM is the best optimizer overall when using the optimizers with defaults.
}
\label{table:only_lr}
\end{table*}

\clearpage

\begin{table*}[htbp]
\centering
{\small
\begin{tabular}
{|l|c|c|c|c|l|}
 \hline
   & SST-2 & MRPC & CoLA & STS-B & MNLI\\
   Optimizer & Accuracy & Macro-F1 & Matthews & Pearson & Accuracy \\
    \hline 
  AdaBound  &$89.91  \; (0.47)$  &$81.47  \; (2.21)$ & $0.57  \; (0.03)$ &$\textbf{0.88}  \; (0.01)$ & $75.52  \; (1.08)$\\
 AdamW&$91.39  \; (0.70)$ &$81.88  \; (1.56)$  & $\textbf{0.58}  \; (0.05)$&$\textbf{0.88}  \; (0.01)$ & $76.74  \; (0.44)$ \\
 AdaMax &$\textbf{91.94}  \; (2.84)$  &$82.33  \; (1.82)$ & $0.56  \; (0.03)$   & $\textbf{0.88}  \; (0.01)$ & $75.40  \; (0.38)$\\
 Nadam   &$91.35  \; (0.87)$ &$82.27  \; (1.83)$  & $0.57  \; (0.02)$  & $\textbf{0.88}  \; (0.00)$  & $76.88  \; (0.49)$ \\
  Adam  &$91.17  \; (0.65)$  &$\textbf{83.08}  \; (1.34)$ &$\textbf{0.58}  \; (0.03)$ & $\textbf{0.88}  \; (0.00)$& $\textbf{76.89}  \; (0.47)$ \\
SGDM  &$89.45  \; (0.79)$ &$80.70  \; (1.51)$ &$0.52  \; (0.04)$ & $0.71  \; (0.34)$ &  $65.37  \; (13.46)$ \\
SGD  &$86.08  \; (0.32)$  &$65.30  \; (5.62)$ & $0.27  \; (0.17)$ & $0.74  \; (0.03)$ & $35.86  \; (0.52)$ \\
  \hline
\end{tabular}
} 
\centering
\vspace*{-2mm}
\caption{Evaluation scores on \textbf{test data} with \textbf{all hyperparameters} \textbf{tuned}, using \underline{\textbf{DistilRoBERTa}}. For each dataset, we use \textbf{five random splits} and report the \textbf{average} test score and the \textbf{standard deviation}. As one would expect, the best scores (bold) are now slightly \textbf{better than those of DistilBERT} (cf.\ Table~\ref{table:tunedtest}). Otherwise, the \textbf{findings are similar} to those of the experiments with DistilBERT (Table~\ref{table:tunedtest}), except that SGDM now performs better on CoLA (where it lagged behind the other adaptive optimizers in Table~\ref{table:tunedtest}) and it now performs poorly on STS-B and MNLI (where it was competent). Hence, these experiments confirm that \textbf{SGDM is overall clearly better than SGD}, but still \textbf{worse than the adaptive optimizers}, when all hyperparameters are tuned. Again, \textbf{the five adaptive optimizers perform very similarly}. These findings are also aligned with those of  Fig.~\ref{fig:distilroberta_tuned_curves}.
}
\label{table:distilrobertatunedtest}
\end{table*}

\begin{table*}[htbp]
\centering
{\small
\begin{tabular}
{|l|c|c|c|c|l|}
 \hline
   & SST-2 & MRPC & CoLA & STS-B & MNLI\\
   Optimizer & Accuracy & Macro-F1 & Matthews & Pearson & Accuracy \\
    \hline 
  AdaBound  &  $90.17  \; (0.93)$ & $80.84  \; (1.72)$ &$0.43  \; (0.15)$&$\textbf{0.88}  \; (0.00)$&$56.92  \; (19.86)$\\
 AdamW  &$91.08  \; (1.09)$ &$81.15  \; (1.76)$ & $\textbf{0.59}  \; (0.04)$ &$\textbf{0.88}  \; (0.00)$&$76.86  \; (0.48)$\\
 AdaMax  & $89.17  \; (0.26)$  &$80.96  \; (1.74)$ &$0.53  \; (0.03)$&$0.87  \; (0.01)$& $72.34  \; (0.39)$\\
 Nadam &  $91.08  \; (0.91)$  & $\textbf{81.81}  \; (1.85)$&$0.57  \; (0.03)$&$\textbf{0.88}  \; (0.00)$& $\textbf{76.97}  \; (0.37)$\\
  Adam & $\textbf{91.45}  \; (0.95)$& $81.47  \; (2.05)$ &$0.56  \; (0.05)$&$\textbf{0.88}  \; (0.00)$&$76.73  \; (0.59)$\\
SGDM  &  $89.11  \; (0.33)$ & $81.68  \; (1.39)$&$0.53  \; (0.03)$&$0.87  \; (0.01)$&$69.67  \; (0.88)$\\
SGD  &$86.08  \; (0.32)$& $65.30  \; (5.62)$& $0.27  \; (0.17)$&$0.74  \; (0.03)$ & $35.86  \; (0.52)$ \\
  \hline
\end{tabular}
} 
\centering
\vspace*{-2mm}
\caption{Evaluation scores on \textbf{test data}, having \textbf{tuned only the learning rate}, using \underline{\textbf{DistilRoBERTa}}. Again, we use five random splits and report the average and standard deviation. As in the corresponding DistilBERT experiments (Table~\ref{table:tunedlrtest}), \textbf{AdaMax} lags behind on CoLA and (more) on MNLI. The only important difference compared to Table~\ref{table:tunedlrtest} is that \textbf{AdaBound is now also clearly worse than the other adaptive optimizers} on CoLA and MNLI, where it is outperformed even by SGDM (but not plain SGD). Hence, these experiments confirm the conclusion that \textbf{tuning only the learning rate} of adaptive optimizers is \textbf{in most cases (but not always) as good as tuning all their hyperparameters}. These findings are aligned with those of Fig.~\ref{fig:distilroberta_tunedlr_curves}. Again, the best scores (bold) are better than those of DistilBERT (Table~\ref{table:tunedlrtest}), except for SST-2.
}
\label{table:distilrobertatunedlrtest}
\end{table*}

\begin{table*}[tb]
\centering
{\small 
\begin{tabular}
{|l|c|c|c|c|c|}
 \hline
   & SST-2 & MRPC & CoLA & STS-B & MNLI\\
   Optimizer & Accuracy & Macro-F1 & Matthews & Pearson & Accuracy \\
    \hline 
  AdaBound    &$88.36  \; (0.62)$&$76.48  \; (2.54)$& $0.00  \; (0.00)$& $0.86  \; (0.00)$& $35.34  \; (0.01)$\\
 AdamW    &$55.79  \; (0.01)$&$62.57  \; (0.00)$&$0.00  \; (0.00)$&$0.14  \; (0.11)$&$35.33  \; (0.01)$\\
 AdaMax     &$55.80  \; (0.00)$&$62.57  \; (0.00)$&$0.00  \; (0.00)$&$0.30  \; (0.10)$&$35.33  \; (0.01)$\\
 Nadam    &$55.80  \; (0.01)$&$62.57  \; (0.00)$&$0.00  \; (0.00)$&$0.17  \; (0.03)$& $35.34  \; (0.01)$\\
  Adam    &$55.80  \; (0.00)$& $62.57  \; (0.00)$&$0.00  \; (0.00)$&$0.14  \; (0.11)$& $35.34  \; (0.01)$\\
SGDM    & $\textbf{89.20}  \; (0.62)$& $\textbf{82.28}  \; (1.66)$&$\textbf{0.54}  \; (0.04)$&$\textbf{0.87}  \; (0.01)$&$\textbf{70.14}  \; (0.81)$\\
SGD    &$86.29  \; (0.34)$&$69.14  \; (3.30)$& $0.36  \; (0.03)$&$0.78  \; (0.02)$& $35.90  \; (0.49)$\\
  \hline
\end{tabular}
} 
\centering
\vspace*{-2mm}
\caption{Evaluation scores on \textbf{test data}, with \textbf{all hyperparameters} of the optimizers set to their \textbf{defaults}, using \underline{\textbf{DistilRoBERTa}}. Again, we use five random splits and report the average and standard deviation. As in the corresponding experiments with DistilBERT (Table~\ref{table:defaulttest}), \textbf{SGDM is now the best} optimizer; again \textbf{AdaBound} performs well on SST-2 and STSB, now also relatively well on MRPC, but not on the other two datasets; the other adaptive optimizers perform much worse.
These findings are alinged with those of Fig.~\ref{fig:distilroberta_default_curves}. Again, the best scores (bold) are better than those of DistilBERT (Table~\ref{table:tunedlrtest}), except for SST-2.
}
\label{table:distilrobertadefaulttest}
\vspace*{-1mm}
\end{table*}

\end{document}